\documentclass{article}

\usepackage[preprint]{neurips_2025}

\usepackage[utf8]{inputenc} %
\usepackage[T1]{fontenc}    %
\usepackage{hyperref}       %
\usepackage{url}            %
\usepackage{booktabs}       %
\usepackage{amsfonts}       %
\usepackage{tcolorbox}
\usepackage{nicefrac}       %
\usepackage{microtype}      %
\usepackage{xcolor}         %
\usepackage{hyperref}
\usepackage{url}
\usepackage[utf8]{inputenc} %
\usepackage[T1]{fontenc}    %
\usepackage{hyperref}       %
\usepackage{url}  
\usepackage{amsmath}
\usepackage[table]{xcolor}
\usepackage{wrapfig}
\usepackage{floatrow}
\usepackage{caption}

\usepackage{subcaption}
\usepackage{enumitem}
\usepackage{natbib}
\usepackage{graphicx}%
\usepackage{booktabs}       %
\usepackage{amsfonts}       %
\usepackage{nicefrac}       %
\usepackage{microtype}      %
\usepackage{xcolor}
\usepackage{enumitem}%
\usepackage{multirow}
\usepackage{graphicx}
\usepackage{tabularx}
\usepackage{longtable} %
\usepackage{marvosym}
\usepackage{fontawesome5}
\usepackage{marvosym}

\newcommand{\systemname}{\emph{extract--retrieve--predict} }
\newcommand{\sysstop}{\emph{extract--retrieve--predict}. }
\newcommand{\syscomma}{\emph{extract--retrieve--predict}, }

\newcommand{\mycolortext}[1]{\textcolor{black}{#1}}

\title{Where Do Images Come From? \\ Analyzing Captions to Geographically Profile Datasets}

\author{
Abhipsa Basu$^{1}$ \quad 
Yugam Bahl$^{2}$\thanks{Work done while at the Indian Institute of Science, Bangalore} \quad
Kirti Bhagat$^{1}$ \quad
Preethi Seshadri$^{3}$\\
\textbf{R. Venkatesh Babu$^{1}$} \quad
\textbf{Danish Pruthi$^{1}$} \\
$^{1}$Indian Institute of Science, Bangalore \\
$^{2}$ TNSQ AI \quad
$^{3}$University of California, Irvine
}

\begin{document}

\maketitle

\begin{abstract}
Recent studies show that text-to-image models often fail to generate geographically representative images, raising concerns about the representativeness of their training data and motivating the question: \textit{which parts of the world do these training examples come from?} We geographically profile large-scale multimodal datasets by mapping image–caption pairs to countries based on location information extracted from captions using LLMs. Studying English captions from three widely used datasets (Re-LAION, DataComp1B, and Conceptual Captions) across $20$ common entities (e.g., house, flag), we find that the United States, the United Kingdom, and Canada account for $48.0\%$ of samples, while South American and African countries are severely under-represented with only $1.8\%$ and $3.8\%$ of images, respectively. We observe a strong correlation between a country’s GDP and its representation in the data ($\rho = 0.82$). Examining non-English subsets for $4$ languages from the Re-LAION dataset, we find that representation skews heavily toward countries where these languages are predominantly spoken.
Additionally, 
we find that higher representation does not necessarily 
translate to greater visual or semantic diversity. Finally, analyzing country-specific images generated by Stable Diffusion v1.3 trained on Re-LAION, we show that while generations appear realistic, they are severely limited in their coverage compared to real-world images.

\begin{center}
    \renewcommand{\arraystretch}{1}
    \begin{tabular}{rl}
         \Mundus~\href{https://geoprofiling.github.io/}{\path{https://geoprofiling.github.io/}} \\
    \end{tabular}
\end{center}

\end{abstract}

\section{Introduction}

Vision-language models (VLMs)~\citep{radford2021learning, li2023blip} exhibit geographic biases, 
raising concerns about their deployment in real-world settings. Specifically, recent studies~\citep{Basu_2023_ICCV, hall2023dig, hall2024towards} demonstrate that text-to-image models fail to generate images that accurately reflect different geographical regions of the world, instead exhibiting strong US- and India-centric skews. 
Yet the origins of these geographic disparities and the extent to which they exist in the training data remain understudied.
This raises a central question: how are different geographical regions represented in large-scale vision-language datasets?

Answering this question is essential for understanding and addressing geographic bias in multimodal systems. 
Such an analysis enables data curators to measure and improve geographical representativeness, allows practitioners to make informed dataset choices, and helps auditors probe how training data composition relates to downstream model behavior~\citep{razeghi2022impact}.
However, characterizing the geographical distribution of image-caption data is challenging in practice. Given a multimodal dataset, one may attempt to infer the country of origin of an image-caption pair using image content, metadata, URLs, or captions. While several image geolocalization methods exist,  
their use is largely limited to street-view imagery~\citep{geoclip, astruc2024openstreetview, li2024georeasoner}, limiting their generalizability to the diverse image types found in vision-language datasets. Moreover, image metadata frequently lacks reliable location information, while URLs often provide noisy or incorrect signals. 

Captions, on the other hand, describe image content and may mention locations, making them a natural signal for studying geographical distributions. Inferring location from captions is still inherently difficult: determining whether a word or phase corresponds to a location can be highly contextual (e.g., ``Buffalo'' may refer to either a place or an animal). Even when a location is identified, assigning it to the correct country is often non-trivial (e.g., ``Cambridge'' refers to cities in both the United States and the United Kingdom).

In this paper, we study the geographical distribution of large-scale vision-language datasets by analyzing captions associated with common visual entities (e.g., houses, roads). For each entity, we identify image–caption pairs wherein the caption mentions a location and infer the corresponding country using an LLM. The LLM extracts location mentions from captions and then maps them to countries using a prominent geo-database~\cite{geonames}. Crucially, since not all captions mentioning an entity correspond to images actually depicting that entity, we filter pairs to retain only those where the entity is visually present. This approach allows us to examine country-level distributions of images and to compare visual diversity across countries with different levels of representation.

\begin{figure*}[t!]
\centering
\includegraphics[trim=1cm 3cm 0cm 2cm, clip, width =\textwidth]{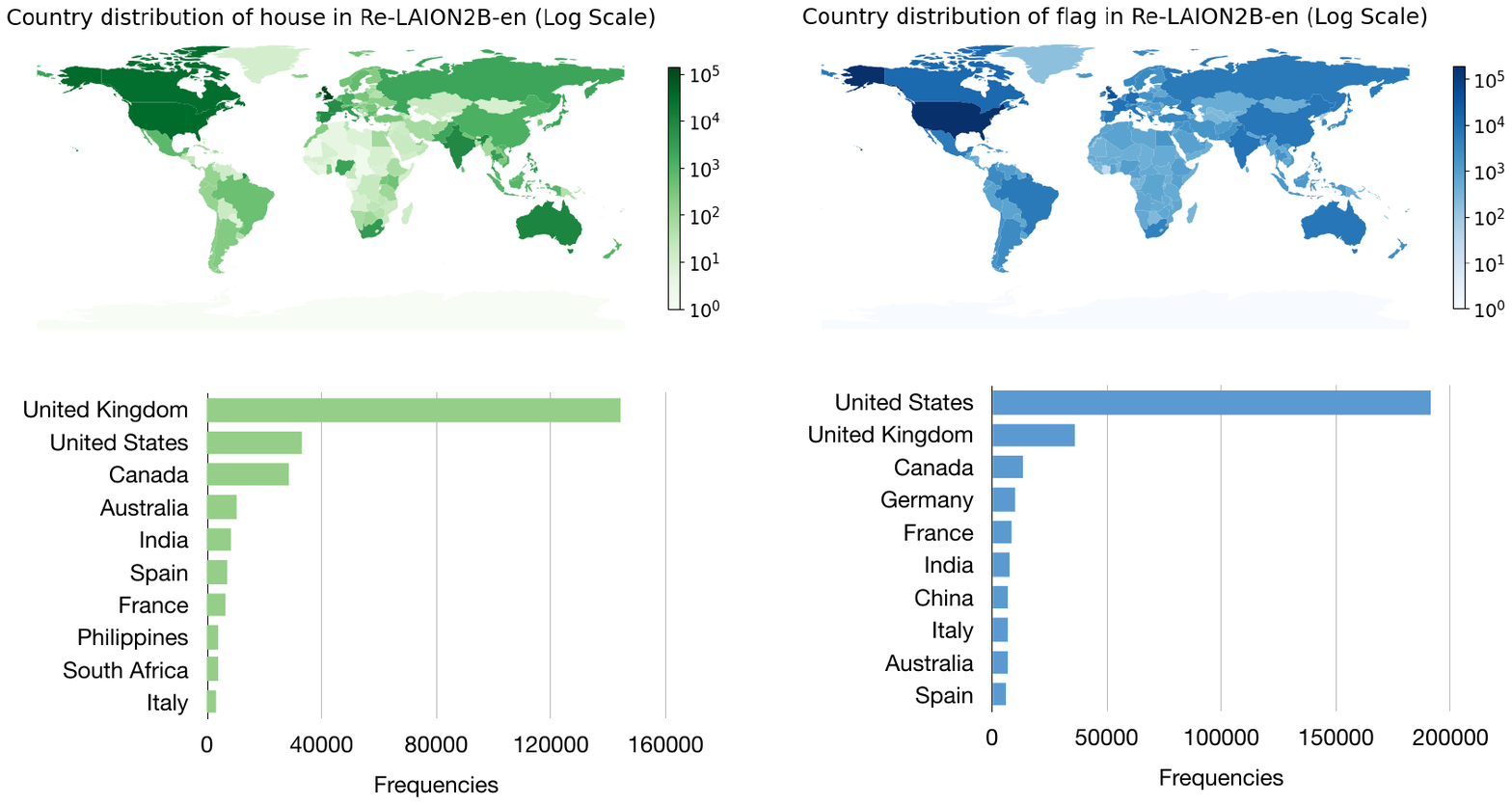}
\caption{\textbf{Distribution of countries for \textit{house} and \textit{flag} in Re-LAION2B-en.} We show the uneven distribution of countries worldwide for house- and flag-related image-caption pairs in the Re-LAION2B-en dataset, along with the top $10$ most frequent countries for each entity.
}
\label{fig:teaser}
\end{figure*}

We geographically profile (geo-profile) three widely used vision–language datasets--Re-LAION2B-en (the English subset of the popular LAION dataset~\cite{schuhmann2022laion}~\footnote{\url{https://huggingface.co/datasets/laion/relaion2B-en-research-safe}}, DataComp1B~\cite{gadre2023datacomp}, and CC12M~\cite{sharma2018conceptual}, all of which are commonly used to train multimodal models such as Stable Diffusion~\cite{ramesh2022hierarchical} and Qwen-VL~\cite{bai2023qwen}. Analyzing
$20$ common entities across these datasets, we uncover substantial geographic disparities
 (e.g., Figure~\ref{fig:teaser}). For instance, in Re-LAION2B-en, $77.2\%$ of location-specified captions originate from just $15$ countries, and moreover, $12$ of the top $15$ countries are common across all three datasets. Across all datasets, the frequency of image-caption pairs per country strongly correlates with nominal GDP ($\rho = 0.82$), with South American and African countries consistently underrepresented, appearing in only $1.8\%$ and $3.8\%$ of captions, respectively. Comparing geographical distributions to real-world occurrence for a subset of entities, we find that, on average, $33.8\%$ of countries are underrepresented relative to their true distributions. We further analyze multilingual captions (Spanish, Hindi, Greek, Japanese) in Re-LAION, and find that countries where these languages are predominantly spoken are the most frequently represented. For example, South America accounts for $26.4\%$ of the countries mentioned in Spanish captions, compared to only $1.8\%$  in English captions. 
 We also study image and caption diversity within countries, finding that representation is only moderately correlated with diversity. Finally, analyzing country-specific images generated by Stable Diffusion v1.3 trained on LAION, we show that although generations appear realistic, they are limited in their coverage.

While geographic biases in model outputs have been documented, the underlying data distributions remain largely unexamined. We provide the first large-scale, data-centric analysis of geographical representation in vision-language datasets, revealing substantial disparities across regions. Our work provides a foundation for systematic quantification of geographical representation and informs the development of more geographically-inclusive multimodal datasets and models.

\section{Related Work}
\label{sec:related}

\noindent \textbf{Evaluation of Geographical Representation in Models.}
Assessing the geographical representation of large-scale models is important for several applications, including text-to-image generation~\citep{Basu_2023_ICCV,hall2023dig, 10.1145/3600211.3604711, hall2024towards}, language generation~\citep{erasure, zhou2022richer, li-etal-2022-herb, godey2024scaling} and image search~\citep{mandal2021dataset}. Some studies highlight economic and geographical disparities in model performance~\citep{gustafson2023exploring, de2019does, bhagat2025richer, bhagat2025tales}. A recent study finds that popular text-to-image generative models exhibit lower diversity and realism when generating images from African and Western Asian countries as compared to images from European nations \citep{hall2023dig}. Moreover, another work~\citep{hall2024towards} discovers that the perception of geographical representation varies region to region over the globe, making evaluation of such models more difficult. 
Some works also highlight how countries with large English-speaking populations are underpredicted by language models~\citep{erasure}. Not only do these biases affect generative models, but also the state-of-the-art object recognition models. For example, Gustafon et al.~\citep{gustafson2023exploring} show that performances of popular models like CLIP~\citep{radford2021learning} degrade with decreasing income level. Similar performance drops are seen in other works as well~\citep{ramaswamy2024geode}. A potential factor underlying several of these findings is the composition of training data, which motivates this analysis. 

\noindent \textbf{Geoparsing Algorithms.} Extracting locations from text and mapping them to countries is well studied~\citep{middleton2018location, martinez2020knowledge, hu2022gazpne2, spacy, kordopatis2017geotagging, luo2011geotagging}. Tools like Geoparsepy~\citep{middleton2018location}, a multilingual geoparsing system using the OpenStreetMap (OSM)\citep{osm} gazetteer, help recognize diverse place names. %
GazPNE2 \citep{hu2022gazpne2} uses deep learning with gazetteers like OSM and GeoNames~\citep{geonames} for location extraction. While precise, these systems often suffer from low recall~\citep{hu2023location}, rely heavily on gazetteers, and perform best on formal text, limiting their coverage. Our approach leverages modern LLMs to extract and disambiguate location mentions into country names to improve coverage. Several works attempt to geolocalize images directly~\cite{geoclip, haas2024pigeon, li2024georeasoner}, using advanced vision–language models (VLMs) which have demonstrated strong capabilities~\cite{geobench2024, waheed2025vlm, jay2025evaluating}, particularly using large proprietary models such as Gemini~\cite{gemini} and GPT~\cite{achiam2023gpt}. However, utilizing evaluating the efficacy of such methods is challenging because web-scale datasets lack ground-truth data (about the geographical location of images), and manually annotating them is prohibitively expensive and prone to errors.

\noindent \textbf{Geographical Profiling of Existing Datasets}. Previous studies~\citep{de2019does, shankar2017no, naggita, wang2022revise, faisal-etal-2022-dataset} show that open-source visual datasets like ImageNet~\citep{deng2009imagenet}, OpenImages~\citep{krasin2017openimages}, and MS-COCO~\citep{lin2014microsoft} overrepresent North America and Europe~\citep{de2019does, shankar2017no}. Additionally, web-scraped images from African countries often reflect Western perspectives rather than local ones~\citep{naggita}. REVISE~\citep{wang2022revise} measures biases in image datasets with respect to objects, people and also geographies.
Most studies obtain geographical annotations via the Flickr API or external services. Another analysis~\citep{faisal-etal-2022-dataset} finds that datasets like SQuAD~\citep{rajpurkar2016squad} and MLQA~\citep{lewis2019mlqa} overrepresent English-speaking and wealthier countries. While their work focuses on language datasets, our study examines vision-language datasets. 

\noindent \textcolor{black}{\textbf{Existing Geo-diverse Datasets}. Existing datasets overrepresent Western, English-speaking countries~\citep{de2019does, shankar2017no}. 
Thus, several geo-diverse datasets have been recently proposed.
For example, \textit{DollarStreet}~\citep{rojas2022the} and \textit{GeoDE}~\citep{ramaswamy2024geode} are geo-diverse image datasets collected through manual and crowdsourcing efforts. 
Other multi-cultural geo-diverse image and language datasets include GeoYFCC~\citep{dubey2021adaptive}, MaRVL~\citep{liu2021visually}, GD-VCR~\citep{yin-etal-2021-broaden}, CultureAtlas~\citep{fung2024massively}, \textit{GeoNet}~\citep{Kalluri_2023_CVPR}, among others. }

\section{Methodology}
\label{sec:geoprofiling}
In this section, we describe our approach of detecting countries from captions.
We first discuss the necessary notations, and then provide details about the various components of our system. 

\noindent \textbf{Preliminaries}. Let $\mathcal{D}=\{(x_i, y_i)\}_{i=1}^N$ be a vision-language dataset where $y_i$ is the caption accompanying image $x_i$.
To analyze geographical distributions across different entities we analyse $k$ entities, wherein each entity $e_1 \dots e_k$ is among the most frequent entities in multimodal datasets such as Re-LAION~\cite{schuhmann2022laion}, and is widely recognizable and globally relevant (e.g., `house').
We then randomly sample $\mathcal{D}_e \in \mathcal{D}$ for each entity $e$
such that for each $(x_i, y_i) \in \mathcal{D}_e$, $e$ is a word in $y_i$. 
We explore several approaches to geolocalize each $(x_i, y_i)$ to $c_i \in \mathcal{C}$, which is the the set of all countries including an additional ``no country'' tag.
For the scope of this study, we choose countries as our denomination of analysis, following previous works~\cite{Basu_2023_ICCV, hall2023dig}. 
In the following subsections, we describe the method we use for geo-profiling $\mathcal{D}_e$ (\S\ref{subsec:geoprofile}).
To ensure that we only consider images that contain the entity $e$, we train and use an entity-presence classifier (\S\ref{subsec: entity-presence}).

\subsection{Geolocalizing Captions}
\label{subsec:geoprofile}

Inferring countries from image captions is a two-step process: (1) identifying location mentions within captions (e.g., \emph{Cambridge}), and (2) mapping these locations to their corresponding countries (e.g., the United Kingdom). 

\paragraph{Our Approach.}

We explore the option of identifying countries from captions via naive string matching against geo-databases such as GeoNames~\citep{geonames}, and then with NER taggers~\citep{spacy} to identify location mentions. However, these approaches suffer from several limitations. Naive string matching ignores contextual information and can incorrectly tag common words that coincide with place names. Although NER-based methods reduce some false positives by explicitly labeling locations, they often miss valid location mentions, leading to false negatives. In addition, geo-databases do not encode the likelihood that a place name belongs to a particular country, making it difficult to resolve ambiguous locations that appear in multiple countries or accurately detect locations that are also common nouns (e.g., Buffalo).  Other geoparsing tools, such as Geoparsepy~\cite{middleton2018location}, inherit similar limitations. Furthermore, given that web-scraped captions often exhibit substantial linguistic variability, including informal phrasing, incomplete sentences, grammatical errors, and complex compositional structures, the above methods are inadequate for reliable country inference.

These shortcomings motivate us to use LLM-based approaches that can incorporate contextual reasoning and external geographic knowledge in a unified framework. To predict countries from captions, we investigate three approaches, including a zero-shot LLM baseline and two methods that improve on it. First, we study the zero-shot performance of LLMs in geolocalizing captions. Second, we propose an \emph{extract--retrieve--predict} protocol. In this approach, the LLM first \emph{extracts} location mentions from a caption, if present. These mentions are then used to \emph{retrieve} the top-$k$ matching locations from the GeoNames database~\cite{geonames}, a widely used gazetteer containing location names and their corresponding countries. Finally, the model \emph{predicts} the country, with the retrieved locations and their associated country names provided as additional context in the prompt. This design provides the model with structured geographic candidates while still allowing it to leverage its internal world knowledge for final country prediction. 

In addition, we study a third approach based on \emph{in-context learning}, where a few country-annotated captions are included in the prompt as exemplars. This provides representative supervision without explicit model fine-tuning.

\paragraph{Evaluation Datasets.}

To evaluate different approaches across diverse geographical contexts, we curate three annotated caption datasets. The first dataset consists of $5{,}000$ captions randomly sampled from the web-crawled LAION dataset, with country names manually annotated by the authors. This dataset, denoted as $\mathcal{D}_{\text{self}}$, evaluates a model's ability to geo-profile real-world web captions of varying linguistic complexity.

To ensure broad global coverage, the second evaluation set is constructed using all GeoNames locations with populations of at least $10{,}000$ inhabitants, resulting in approximately $57\text{K}$ locations. For each location, a location-specified caption is selected from $\mathcal{D}_{\text{self}}$, and the original location mention is replaced with the sampled GeoNames location, manually ensuring that there are no other unrelated location mentions within the caption. 
Country annotations for this set are obtained directly from the GeoNames database. This dataset, denoted as $\mathcal{D}_{\text{geo}}$, evaluates the ability to geolocalize captions referring to locations worldwide and spanning a wide range of population densities.

Finally, to assess performance on underrepresented regions, we curate a third dataset, $\mathcal{D}_{\text{marginalized}}$, comprising $1{,}665$ sentences extracted from Wikipedia corresponding to ten countries with low GDP, spread across different continents--Botswana, Algeria, Guyana, Tuvalu, Kiribati, Malta, S\~{a}o Tom\'{e} and Pr\'{\i}ncipe, Tajikistan, Yemen, and Uruguay. Each sentence contains references to local cities. Further details on dataset construction are provided in Appendix~\ref{subsec: geoprofile-2}.

\paragraph{Models and Baselines.}

We evaluate multiple LLM backbones for country inference, including LLaMA-3.1-8B Instruct~\citep{dubey2024llama}, Qwen-2.5-7B Instruct~\citep{bai2023qwen}, and Gemini-2.5-Flash~\cite{gemini}, alongside non-LLM baselines such as substring matching, NER-assisted matching, and existing geoparsing tools. The commercial LLMs like Gemini-2.5-Flash~\cite{gemini} and GPT~-4~\cite{achiam2023gpt} are found to outperform the open-source ones; consequently, both the in-context learning and extract--retrieve--predict approaches are implemented using Gemini-2.5~-Flash. Although in-context learning (using $1\%$ of the combined country-annotated evaluation sets as examples) yields strong performance, it is impractical for large-scale analysis due to its substantial context-length requirements. We therefore adopt the extract--retrieve--predict approach to geo-profile captions in large-scale vision-language datasets. Quantitative results for all methods are reported in Table~\ref{tab:geotagging}, with prompt details and baseline descriptions provided in Appendix~\ref{subsec: geoprofile-2}. 
We further extend the extract--retrieve--predict approach to multilingual captions by translating them into English using the NLLB-200-3.3B model~\cite{costa2022no}.

\textbf{Note on Utilizing Images for Geolocalizing Data Points}. Recent advances in vision–language models (VLMs) have demonstrated strong capabilities in directly geolocalizing images~\cite{geoclip, haas2024pigeon, geobench2024, waheed2025vlm, jay2025evaluating}, particularly for large proprietary models such as Gemini and GPT. In principle, such models could combine visual cues with accompanying captions to infer geographic information. However, there is currently no reliable way to validate image-based geolocalization performance on existing web-scale vision–language datasets, as these datasets lack ground-truth annotations. Moreover, externally annotating such datasets with country-level labels at scale, and subsequently querying commercial models, is technically challenging, prohibitively costly, and susceptible to annotation noise. Consequently, we restrict our analysis to caption-based signals and leave image-based geolocalization, along with its large-scale validation and integration, to future work.

\begin{table}[]
\caption{\textbf{Performance of different geo-profiling methods}. LLM-based approaches outperform pre-LLM baselines across all three test sets, with Gemini-2.5-Flash achieving the best overall performance. We further improve its zero-shot results by using the model to first \textit{extract} explicit location mentions from each caption, \textit{retrieve} the most similar entries from a geo-database, and then \textit{predict} the country by augmenting the prompt with the retrieved locations and their associated country names.
}
\small
\label{tab:geotagging}
\begin{tabular}{@{}l|l|ll|ll|ll@{}}
\toprule
\multirow{2}{*}{Method Type} & \multirow{2}{*}{\begin{tabular}[c]{@{}l@{}}Method/Model \\ Name \end{tabular}} & \multicolumn{2}{l|}{\begin{tabular}[c]{@{}l@{}}GeoNames \\ Dataset $\mathcal{D}_{\text{geo}}$ \\ (57K samples)\end{tabular}} & \multicolumn{2}{l|}{\begin{tabular}[c]{@{}l@{}}Marginalized \\ Dataset $\mathcal{D}_{\text{marginalized}}$\\ (1.6K samples)\end{tabular}} & \multicolumn{2}{l}{\begin{tabular}[c]{@{}l@{}}Self Annotated\\ Dataset $\mathcal{D}_{\text{self}}$\\ (5K samples)\end{tabular}} \\ \cmidrule(l){3-8} 
 &  & \multicolumn{1}{l|}{Precision} & Recall & \multicolumn{1}{l|}{Precision} & Recall & \multicolumn{1}{l|}{Precision} & Recall \\ \midrule
Pre-LLM & String Matching & \multicolumn{1}{l|}{$0.24$} & $0.80$ & \multicolumn{1}{l|}{$0.13$} & $0.81$ & \multicolumn{1}{l|}{$0.9$} & $0.28$ \\
 & NER Tagger & \multicolumn{1}{l|}{$0.49$} & $0.41$ & \multicolumn{1}{l|}{$0.49$} & $0.21$ & \multicolumn{1}{l|}{$0.76$} & $0.57$ \\
 & Geoparsepy & \multicolumn{1}{l|}{$0.55$} & $0.49$ & \multicolumn{1}{l|}{$0.67$} & $0.65$ & \multicolumn{1}{l|}{$0.80$} & $0.67$ \\
 & Geograpy3 & \multicolumn{1}{l|}{$0.17$} & $0.21$ & \multicolumn{1}{l|}{$0.38$} & $0.61$ & \multicolumn{1}{l|}{$0.48$} & $0.66$ \\
 \midrule
\multirow{6}{*}{\begin{tabular}[c]{@{}l@{}}Zero-Shot \\ LLM\end{tabular}} & Gemma2-9B-it & \multicolumn{1}{l|}{$0.73$} & $0.73$ & \multicolumn{1}{l|}{$0.79$} & $0.78$ & \multicolumn{1}{l|}{$0.91$} & $0.91$ \\
 & Qwen-2.5-7B Instruct & \multicolumn{1}{l|}{$0.81$} & $0.61$ & \multicolumn{1}{l|}{$0.81$} & $0.71$ & \multicolumn{1}{l|}{$0.92$} & $0.91$ \\
 & Llama-3.1-8B & \multicolumn{1}{l|}{$0.73$} & $0.73$ & \multicolumn{1}{l|}{$0.85$} & $0.84$ & \multicolumn{1}{l|}{$0.89$} & $0.88$ \\
 & GPT 4 & \multicolumn{1}{l|}{$0.83$} & $0.84$ & \multicolumn{1}{l|}{$0.92$} & $0.92$ & \multicolumn{1}{l|}{0.93} & $0.93$ \\
 & Gemini-2.5-flash & \multicolumn{1}{l|}{$0.93$} & $0.83$ & \multicolumn{1}{l|}{$0.95$} & $0.91$ & \multicolumn{1}{l|}{$0.88$} & $0.92$ \\
 \midrule
LLM + ICL & Gemini-2.5-flash & \multicolumn{1}{l|}{$0.98$} & $0.95$ & \multicolumn{1}{l|}{$0.96$} & $0.92$ & \multicolumn{1}{l|}{$0.88$} & $0.92$ \\
\midrule
\begin{tabular}[c]{@{}l@{}}\cellcolor{blue!10} LLM + \\\cellcolor{blue!10} extract-retrieve\\ \cellcolor{blue!10} -predict\end{tabular} & \cellcolor{blue!10} Gemini-2.5-flash & \multicolumn{1}{l|}{\cellcolor{blue!10} $0.97$} & \cellcolor{blue!10} $0.91$ & \multicolumn{1}{l|}{\cellcolor{blue!10} $0.95$} & \cellcolor{blue!10} $0.91$ & \multicolumn{1}{l|}{\cellcolor{blue!10} $0.91$} & \cellcolor{blue!10} $0.93$ \\ \bottomrule
\end{tabular}
\end{table}

\begin{figure*}[t!]
    \centering
    \includegraphics[width = 0.90\linewidth]{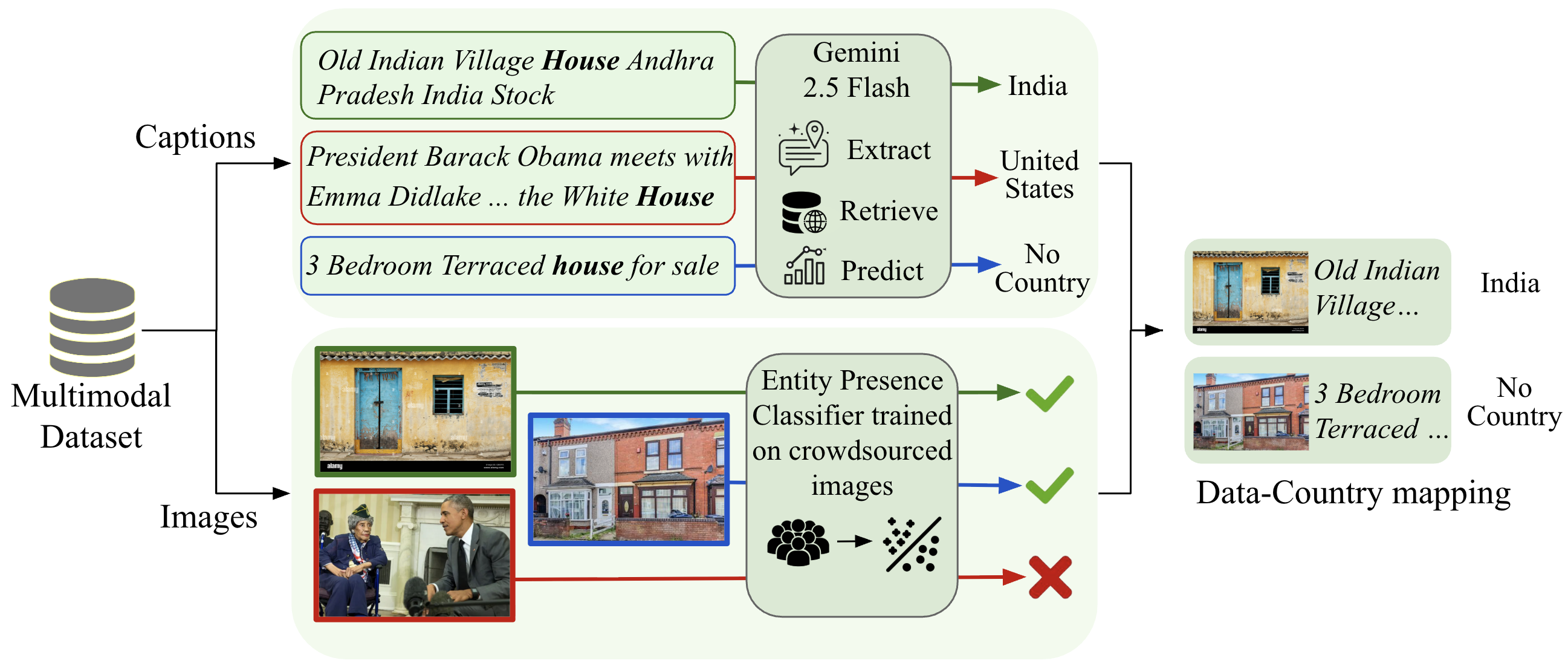}
    \caption{\textbf{The proposed approach.} Given an image-caption pair of an entity, we first map captions to countries (using Gemini), then filter out images that do not contain the entity (using the entity presence classifier), and finally return the countries for resulting image-caption pairs.
    } 
    \label{fig:method}
\end{figure*}

\subsection{Entity Presence Filtering}
\label{subsec: entity-presence}

While the dataset $\mathcal{D}_e$ consists of captions containing the entity $e$ and corresponding images, 
images in $\mathcal{D}_e$ may not always depict the entity $e$
. For example, a caption mentioning the entity ``house'' may depict the people in the house in the image rather than the entity itself.
This calls for a filtering step to remove these irrelevant images. %
A straightforward approach to this step might be to use zero-shot foundation models like CLIP~\citep{radford2021learning} and Visual Question Answering (VQA) models like BLIP~\cite{li2023blip} to identify images depicting the entity. However, we find such models to be inadequate. 
Therefore, we train, and subsequently use, an entity presence classifier, which is a binary classifier that predicts whether a given entity $e$ is present in an image. 

To train this classifier, we sample approximately $600$ images from $\mathcal{D}_e$ for each entity $e$ (Appendix Table~\ref{tab:annotation-image}). To ensure geographical diversity, we leverage the geolocalized caption predictions to sample images across a combination of $17$ economic regions~\citep{region} and $4$ income groups~\citep{income} (see Appendix~\ref{subsec:dataset creation}). Each image is annotated by $3$ crowd workers from the Prolific~\cite{prolific} platform, who are asked to indicate the presence or absence of the specified entity in the image, and the final label is determined by majority vote. Averaged across entities, all pairs of crowdworkers agree $83.6\%$ of times (entity-wise agreement scores provided in Appendix~\ref{subsec:inter}). From the annotated set, $100$ images are reserved for testing, while the remainder are used to train an SVM classifier on CLIP image features, yielding an F1-score of $0.88$. In comparison, a zero-shot CLIP classifier achieves a significantly lower F1-score of $0.69$, while the BLIP VQA model reaches $0.77$ (see Appendix~\ref{subsec:benchmark-svm} for details). We use the trained entity-presence classifier to filter out irrelevant image-caption pairs from $\mathcal{D}_e$. Additional implementation details are provided in Appendix~\ref{subsec:npf-app}, and an overview of our approach is presented in Figure~\ref{fig:method}. Qualitative examples of relevant and irrelevant images detected by our classifiers are shown in Appendix Figure~\ref{tab:images-qualitative}.

\section{Case Studies}
\label{sec: analyses}

Geo-profiling datasets can offer answers to several important questions 
about their contents and models trained using them. 
We first describe the chosen entities
and datasets to be analyzed.
We then study the geographical distributions of these datasets with respect to the chosen entities in subsection~\ref{subsec: distribution1} and compare the obtained distributions with their ground truth distributions of a subset of entities in subsection~\ref{subsec: gr}. We also study the country-wise diversity of the images and captions in subsection~\ref{subsec: diversity}. Qualitative examples of the images for different entities and countries are shown in Appendix~\ref{subsec: q}.

\noindent \textbf{Entity Selection.} 
We select $k=20$ entities frequently observed in large-scale vision-language datasets and representative of diverse visual contexts: \textit{house, apartment, bedroom, kitchen, toilet, living room, office, hotel, shop} (built environment and indoor spaces); \textit{mountain, hill, lake, beach, island, valley} (natural landscapes); \textit{car, road} (transportation and infrastructure); \textit{jewelry, wedding dress} (human-worn objects); and \textit{flag} (national symbol)\footnote{We avoid parent categories, such as food  or festivals, which have different instances across cultures.}. These entities are commonly encountered in datasets such as LAION, as verified using the WIMBD tool~\cite{elazar2024whats}, and are universal in nature.

\noindent \textbf{Datasets}. We analyze the English subsets of the Re-LAION-5B, the safe version of the original LAION-5B dataset (denoted by Re-LAION2B-en), the DataComp1B dataset and the Conceptual Captions 12M dataset (CC12M). To examine the effect of language on geographic distribution, we also study Spanish, Greek, Hindi and Japanese captions from the Re-LAION2B-multi dataset. While the Re-LAION and DataComp originate from Common Crawl~\citep{commoncrawl}, CC12M is collected using Google's undisclosed crawlers. For entity 
$e$, we sample up to 1M image-caption pairs from the English datasets using the WIMBD tool~\citep{elazar2024whats}, ensuring that each caption contains $e$. If fewer than $1$M captions exist, we include all available instances. For multilingual captions, we select up to $100\text{K}$ samples per entity, excluding those with fewer than $1$K captions in a given language.

\subsection{Distribution-Based Analysis}
\label{subsec: distribution1}

Based on the country predictions obtained via the \systemname approach, we study the geographical distribution of $\mathcal{D}_e$ for each entity $e$.

\begin{figure*}[t!]
    \centering
    \includegraphics[trim=1.5cm 5cm 2.5cm 1.5cm, clip, width=\linewidth]{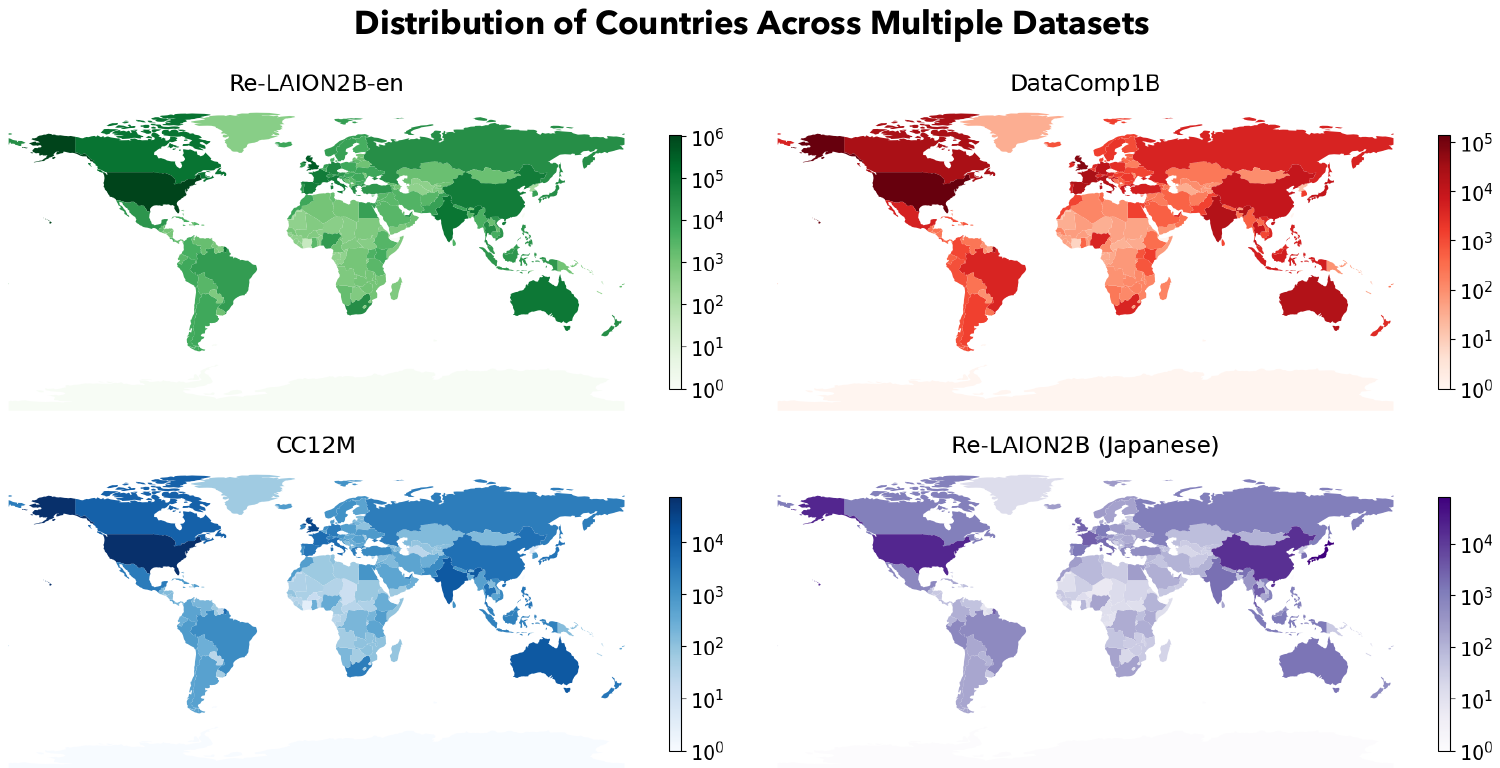} %
    \caption{\textbf{Distribution of the countries across different datasets}, averaged across entities. All datasets with English captions overrepresent countries like the US, UK, Canada, and under represent Afrian and South American nations. For the Japanese captions, Japan is the most represented country.
    }
    \label{fig:overall_distributions}
\end{figure*}

\begin{figure*}[t!]
    \centering
    \includegraphics[trim=0cm 23cm 0cm 0cm, clip, width=\linewidth]{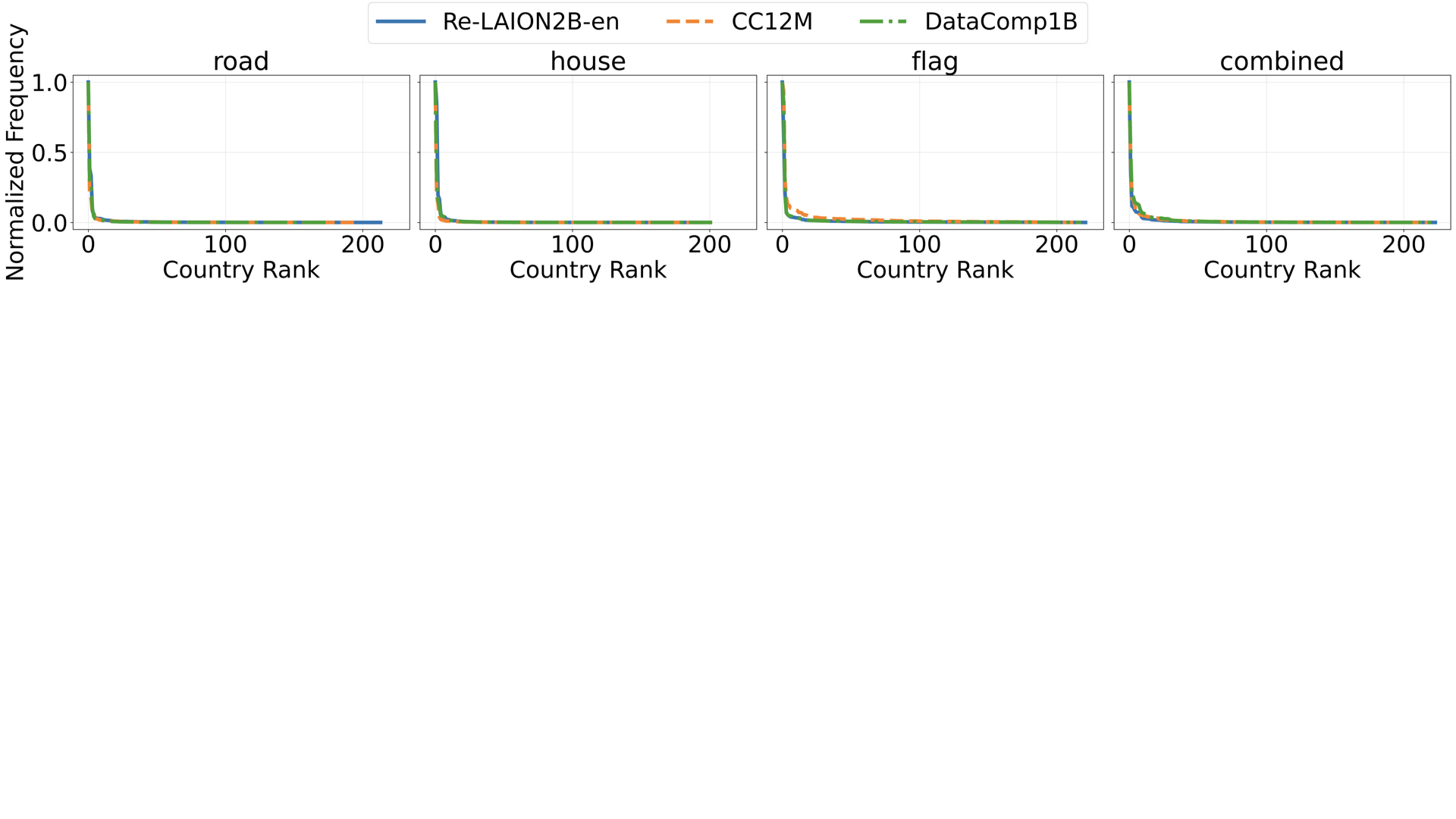} %
    \caption{\textbf{Country distributions exhibit long-tailed behavior across the different datasets}. This trend is visible for individual entities like road, house and flag, as well as for all entities combined.
    }
    \label{fig:long-tailed}
\end{figure*}

\noindent \textbf{Underspecified Image Captions}. We find $55.1\%$ of LAION2B-en, $57.6\%$ of DataComp1B, $58.0\%$ of CC-12M and $60.3\%$ of the multilingual captions as underspecified. Certain entities, like living room, kitchen and toilet, have over $80\%$ underspecification, likely due to the lack of location details in their descriptions. In contrast, apartment, hotel, flag, valley and island have the lowest rates ($<30\%$ on average), as their captions often include location cues. Table~\ref{tab:underspecifications_main} reports the percentage of underspecified captions for selected entities in the English-caption datasets. Detailed results for all entities and datasets are provided in Appendix Table~\ref{tab:underspecifications}. The high percentages of underspecified captions highlights a fundamental challenge in geo-profiling datasets. 

\begin{table}[]
\caption{\textbf{Percentage of location-underspecified captions across six entities}. While underspecification rates are found to be more for entities like car, kitchen, living room, they are significantly lower for others like flag, hotel and house. We find $55.1\%$ of Re-LAION2B-en, $57.6\%$ of DataComp1B and $58.0\%$ of CC12M captions are found to be location-underspecified. }
\label{tab:underspecifications_main}
\small
\begin{tabular}{@{}l|ccc@{}}
\toprule
Entity        & Re-LAION2B-en & DataComp1B & CC12M  \\ \midrule
flag          & $19.8$        & $24.5$     & $16.1$             \\
hotel         & $16.8$        & $30.8$     & $27.6$           \\
house         & $37.1$        & $49.3$     & $61.3$           \\
car           & $66.8$        & $73.5$     & $74.4$     \\
kitchen       & $83.0$        & $88.3$     & $82.2$                 \\
living room   & $84.8$        & $87.5$     & $83.6$             \\ \bottomrule
\end{tabular}
\end{table}

\noindent \textbf{Location-Specified Captions}. 
A key question we seek to find answer to in the paper is: \textit{how are different countries represented in large-scale vision–language datasets?}
We find that country-wise distributions in location-specified captions are highly skewed. The United States and the United Kingdom are the most frequently represented countries in Re-LAION2B-en, DataComp1B, and CC-12M, while in multilingual captions, the most represented countries largely correspond to regions where the respective languages are predominantly spoken\footnote{While the accuracy of multilingual outputs is constrained by translation quality, our findings suggest that increasing multilingual training data could improve cultural awareness in future models.}. Notably, $77.2\%$ of the location-specified captions in Re-LAION2B-en originate from just $15$ countries, with similar concentration observed in DataComp1B ($74.0\%$) and CC-12M ($70.8\%$), with entity- and dataset-wise details provided in Appendix~\ref{subsubsec:top15}. Correspondingly, Figure~\ref{fig:long-tailed} illustrates that country frequencies tend to exhibit a long-tailed pattern, both at the level of individual entities (e.g., \emph{road}, \emph{house}, and \emph{flag}) and when aggregated across entities. \textbf{Moreover, $12$ of the top $15$ countries are common across Re-LAION2B-en, DataComp1B, and CC-12M}. 
This suggests that distinct English caption datasets maintain nearly identical patterns of country representation. We show the top $5$ countries per dataset in Figure~\ref{fig:top10} (see Appendix Figure~\ref{fig:top15appn} for the top $15$ most frequent countries across the English and multilingial caption datasets). At the continent level, North America ($42.8\%$) and Europe ($30.8\%$) dominate Re-LAION2B-en, followed by Asia ($16.7\%$), aligning with prior findings of North American and European overrepresentation~\citep{de2019does, shankar2017no}. In contrast, Oceania, Africa, and South America together account for only $9.7\%$ of English captions, while South America is substantially better represented in Spanish ones ($26.4\%$), second only to Europe ($40.9\%$), reflecting the continent’s large Spanish-speaking population; dataset-specific continent-level statistics are provided in Appendix~\ref{subsec: continent}.

\noindent \textbf{Correlation with GDP (nominal).} 
We find a very strong correlation between country frequency in Re-LAION2B-en and nominal GDP ($\rho = 0.84$), with $13$ of $20$ entities showing $\rho \geq 0.8$. House is the only weakly correlated entity ($\rho = 0.28$). Frequency also shows weak but positive correlations with population, GDP per capita ($\rho = 0.26$ and $0.22$ respectively) and number of internet users ($\rho = 0.34$). For DataComp1B and CC-12M, frequency and nominal GDP correlate at $\rho = 0.81$ and $0.82$ respectively (see Appendix~\ref{subsec: datacomp-socio} for other socio-economic factors). These results show that wealthier countries tend to appear more frequently across entities.

\begin{figure*}[t!]
    \centering
    \includegraphics[trim=0cm 11.7cm 0cm 0cm, clip, width=\linewidth]{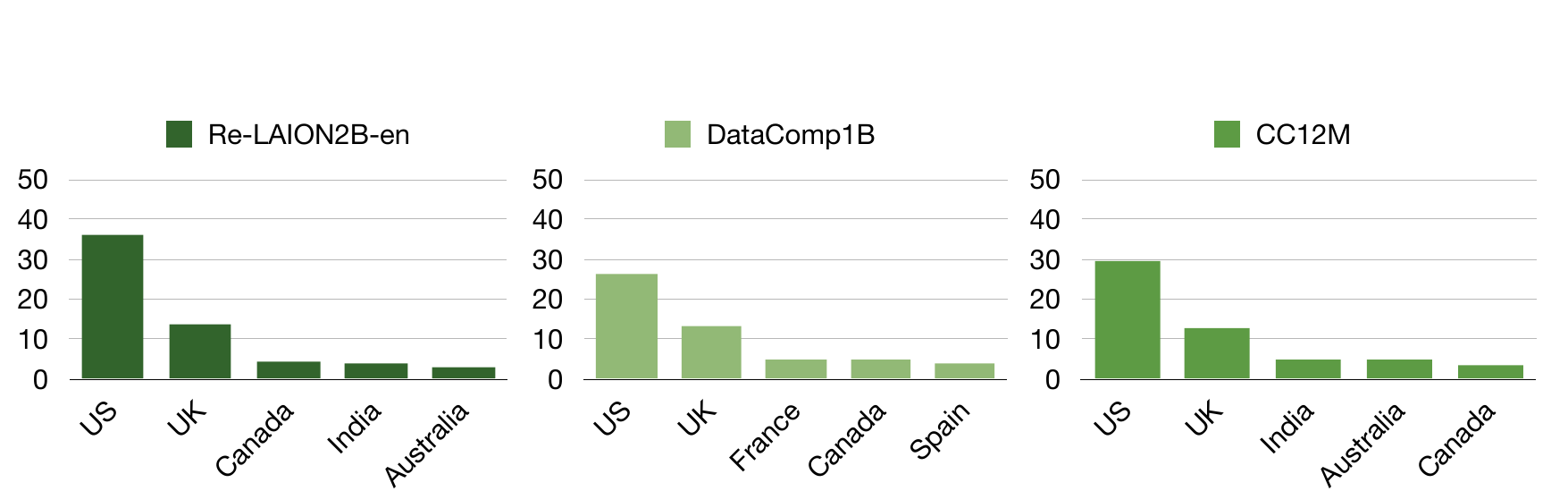} %
    \caption{\textbf{Distribution of the five most frequent countries} in Re-LAION2B-en along with DataComp1B and CC12M, averaged across entities. 
    }
    \label{fig:top10}
\end{figure*}

\subsection{Comparison with Real-world Distributions}
\label{subsec: gr}

In subsection~\ref{subsec: distribution1}, we observe considerable skewness in the distribution of the countries across multiple datasets. This prompts an important question: \textit{do the observed country distributions align with real-world distributions for different entities?}
For example, the entity \emph{beach} would be expected to appear more frequently in countries with coastlines than in landlocked countries.
While exact proportionality is unlikely, real-world prevalence provides a natural guideline for assessing geographic over- and under-representation.
We therefore capture the  geographical misalignment of a country for a given entity as the extent to which the country’s frequency in a dataset deviates from its real-world distribution. To obtain ground truth distributions, we rely on credible external sources such as Wikipedia and United Nations statistics (see Appendix~\ref{subsec: gr2}). We compute ground truth distributions for the entities \emph{house, hotel, car, jewelry, beach, road, mountain} and \emph{island}. For other entities, reliable real-world statistics are unavailable, and we therefore exclude them from this analysis.

Our notion of misalignment is inspired by prior work on \emph{geographical erasure}. In particular, Schwöbel et al.~\cite{erasure} study the underprediction of countries by large language models, defining an \emph{erasure set} as the set of countries that are generated at least $r$ times less frequently than expected when compared to a reference distribution (e.g., English-speaking population) ($r$ is a hyperparameter).
Following a similar spirit, we compare the empirical country distribution for selected entities in the studied datasets with its corresponding real-world distribution. A country is considered \emph{overrepresented} if it appears at least $r$ times more frequently in the dataset than in the real world for that entity, and \emph{underrepresented} if it appears less than $(\frac{1}{r})$ times.
While Schwöbel et al.~\cite{erasure} use $r=3$ to identify cases of extreme erasure, we adopt a more conservative threshold of $r=1.5$ in order to capture subtler deviations in geographical representation across countries.

\begin{table*}
\centering
\small
\caption{\textbf{Comparison of the geographical distributions for specific entities in the Re-LAION2B-en dataset with respect to the true distributions}. We show the percentage of countries that are underrepresented (Under) and overrepresentated (Over) for different entities. On average, $33.8\%$ countries are underrepresented, whereas $32.4\%$ are overrepresented across the studied entities.
}
\label{tab:representation}
\begin{tabular}{c|c|c|c|c|c|c|c|c|c}
\hline
 & House & Car & Jewelry & Beach & Road & Hotel & Mountain & Island & Avg \\ \hline
Under & $47.5$ & $47.1$ & $14.1$ & $39.1$ & $23.9$ & $37.3$ & $\mathbf{56.2}$  &  $4.9$ & $33.8$ \\
Over & $14.5$ & $21.1$ & $\mathbf{70.8}$ & $36.9$ & $38.0$ & $21.0$ & $14.1$  & $42.9$ & $32.4$ \\ \hline
\end{tabular}
\end{table*}

\noindent \textbf{Findings}. We report the percentage of under- and over-represented countries below.
We observe that none of the studied entities closely follow their corresponding ground-truth country distributions. In the Re-LAION2B-en dataset, for $5$ of the $8$ entities, more than $35\%$ of countries are underrepresented, as shown in Table~\ref{tab:representation} (corresponding results for DataComp1B are reported in Appendix Table~\ref{tab:representation-datacomp}). On average, $33.8\%$ of countries are underrepresented, while $32.4\%$ are overrepresented. Although the specific over- and underrepresented countries vary across entities, a consistent pattern emerges in which the United States is systematically overrepresented, whereas countries such as Cambodia, Zimbabwe, and Haiti are frequently underrepresented. These results not only reveal substantial and systematic geographic imbalances in large-scale image–caption datasets, but also underscore the value of such analyses for informing dataset curation efforts aimed at better aligning country-level representations with underlying real-world distributions.

\subsection{Diversity Analysis}
\label{subsec: diversity}
Several studies~\cite{hall2023dig, hall2024towards} highlight the lack of diversity in images of marginalized regions such as Africa generated by text-to-image models like Stable Diffusion~\cite{rombach2022high}. While our analysis reveals substantial underrepresentation of African countries in large-scale image-caption datasets, this observation alone does not characterize the variability of the content associated with different countries. This motivates the following question: \emph{Is the diversity of a country’s training images and captions related to how frequently that country appears in the dataset?}

To study this, we require a diversity metric that captures variability within a collection while accounting for semantic similarity between samples. Given a set of elements \(S=\{x_1, x_2, \cdots, x_N\}\) of size \(N\), a feature encoder \(f\), and a similarity matrix \(K\) with entries \(K_{ij} = k(f(x_i), f(x_j))\), where \(k\) denotes a similarity function and \(f(x_i)\) denotes the L2-normalized feature representation~\citep{liu2024rethinking} of \(x_i\), we measure diversity using the Vendi score~\cite{friedman2022vendi}, defined as
\[
div(S) = \exp\left(-\sum_{i=1}^N \lambda_i \log \lambda_i\right),
\]
where  \(\{\lambda_1, \cdots, \lambda_N\}\) are the eigenvalues of the normalized similarity matrix \(K/N\). Intuitively, \(div(S)\) reflects the effective number of distinct samples in the semantic space: it attains its maximum value \(|S|\) when all elements are mutually dissimilar and its minimum value is \(1\) when all elements are identical. The Vendi score has been widely adopted for measuring data diversity across modalities~\cite{askari2024improving, rezaei2025vendi, ba2024does}, and is well suited for comparing collections that vary substantially in size and internal redundancy. Particularly, Askari et al.~\cite{askari2024improving} utilize the Vendi-score to enhance the \textit{geographical diversity} of images generated by Text-to-Image models for different countries.

In practice, we compute the diversity separately for data points associated with a specific country and entity, considering only country-entity pairs with more than \(100\) images. We define the overall diversity of an entity as its average diversity across countries. For image diversity, we use CLIP ViT-B/32~\citep{radford2021learning} as the feature encoder \(f\), following prior work~\cite{askari2024improving}.

\noindent\textbf{Entity-wise Diversity of Images.}
We first compute overall image diversity for each studied entity. In Re-LAION2B-en, we observe that \(14\) out of \(20\) entities exhibit comparable diversity scores, with \emph{shop} achieving the highest score (\(16.4\)), while \emph{kitchen} and \emph{bedroom} are the least diverse (\(4.6\) each). The average diversity score across all entities is \(9.2\). Similar trends are observed in DataComp1B and CC12M, with an average entity-level diversity of \(10.0\). Detailed results for all entities and datasets are reported in Appendix Table~\ref{tab:div_entity_merged}.  

\noindent\textbf{Diversity vs.\ Frequency.}
We next examine whether countries that appear more frequently in the dataset are also associated with more diverse content. For each entity, we compute the Spearman rank correlation coefficient \(\rho_{fi}\) between country-wise image frequencies and image diversity scores. In Re-LAION2B-en, we observe a moderate positive correlation, with an average \(\rho_{fi} = 0.54\). indicating that a higher number of images from a country does not always imply greater visual diversity. 
Qualitative examples illustrating the differences between images from high- and low-diversity countries--Mexico and Norway, respectively--are shown in Appendix Figures~\ref{fig:mexico} and~\ref{fig:norway}. Although Norway appears more frequently in the dataset, it exhibits a lower diversity score compared to Mexico. To better understand this discrepancy, we analyse the associated captions from both countries, and find that images from Norway predominantly depict mountainous landscapes, scenic views, and bridges, whereas images from Mexico reflect a wider range of themes, including the police, state-related contexts, and migrants, among others (see Appendix Figure~\ref{fig:word_cloud}). This qualitative analysis highlights significant differences in how countries are represented in web-scraped datasets, which may in turn contribute to and reinforce harmful biases and stereotypes. The correlation between country-frequency and diversity is slightly stronger in DataComp1B, with an average \(\rho_{fi} = 0.67\) (Appendix Table~\ref{tab:div_corr_datacomp}).

Finally, for each entity \(e\), we compute \textbf{caption diversity} using the same Vendi score formulation and analyze its relationship with country frequency. Across both Re-LAION2B-en and DataComp1B, caption diversity exhibits a moderate positive correlation with frequency (average Spearman’s \(\rho = 0.49\) and \(0.51\), respectively), with entity-wise details shown in Appendix Tables~\ref{tab:div_corr_text_appendix_laion} and ~\ref{tab:div_corr_text_appendix_datacomp}. These trends are consistent with our observations for image diversity.

\begin{table}[]
\caption{\textbf{Frequency and diversity rank correlations for Re-LAION2B-en.} For each entity, we compute the correlation between the frequency of countries and diversity scores of the images ($\rho_{fi}$). We notice that the majority of entities exhibit moderate correlation (average Spearman's $\rho=0.54$). All reported differences are statistically significant (p-value \(< 0.01\)).}
\label{tab:div_corr}
\small
\begin{tabular}{c|cccccccccc}
\toprule
 & House & Flag & Car & Kitchen & Beach & Road & Hotel & \begin{tabular}[c]{@{}c@{}}Bed-\\ room\end{tabular} & Toilet & \begin{tabular}[c]{@{}c@{}}Apart-\\ ment\end{tabular} \\ \midrule
$\rho_{\text{fi}}$ & $0.28$ & $0.53$ & $0.80$ & $0.36$ & $0.55$ & $0.60$ & $0.44$ & $0.53$ & $0.36$ & $0.25$ \\ \midrule
 & \begin{tabular}[c]{@{}c@{}}Moun-\\ tain\end{tabular} & Island & Shop & Valley & \begin{tabular}[c]{@{}c@{}}Living\\ Room\end{tabular} & \begin{tabular}[c]{@{}c@{}}Wedding\\ Dress\end{tabular} & Hill & Lake & Office & Jewelry \\ \hline
$\rho_{\text{fi}}$ & $0.68$ & $0.45$ & $0.61$ & $0.93$ & $0.51$ & $0.46$ & $0.66$ & $0.50$ & $0.28$ & $0.85$ \\ \bottomrule
\end{tabular}
\end{table}

\section{Country-wise Model Generations}

Given the geographical biases observed in the studied vision-language datasets, we study their effect in models trained on such datasets. Specifically, we choose the Stable Diffusion v1.3, which has been exclusively trained on the LAION2B-en dataset. Basu et al.~\cite{Basu_2023_ICCV} find that images of different entities, generated by underspecified prompts (e.g., `\texttt{photo of a \{entity\}}') overrepresent countries like the US, UK, Canada and India, thus showing that models trained on such datasets are biased as well. 

\begin{figure*}[t!]
    \centering
    \includegraphics[trim=0cm 15cm 0cm 0cm, clip, width=\linewidth]{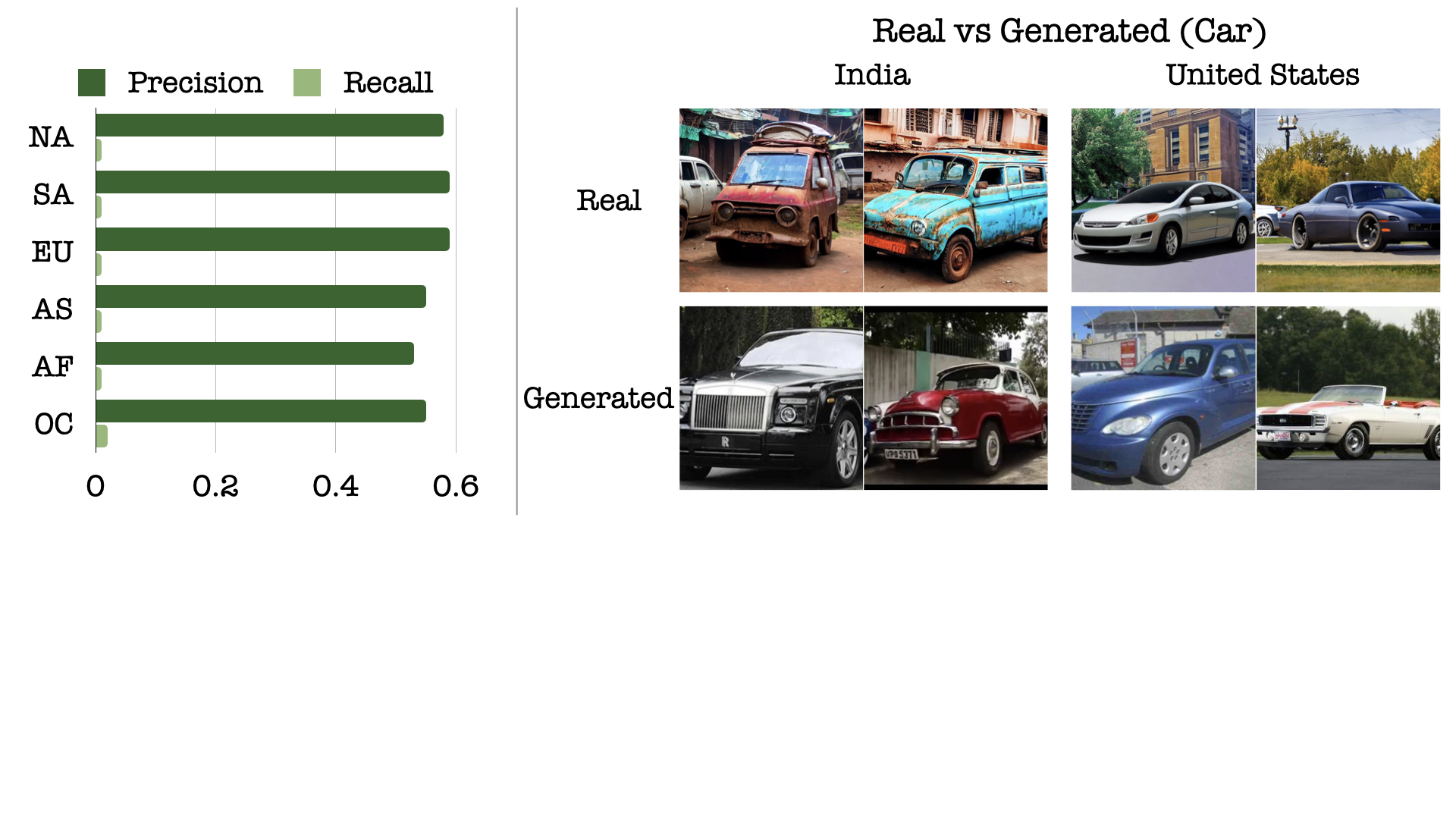} %
    \caption{\textbf{Analysis of images generated by Stable Diffusion v1.3 across continents} shows that, while the images are generally realistic, they do not fully reflect the diversity present in real-world data, as measured by the precision (realism) and recall (coverage) metrics of Sajjadi et al.~\cite{sajjadi2018assessing}. Qualitative inspection further reveals alarming differences across regions: for example, images generated for Indian cars tend to depict older and dilapidated vehicles (a pattern also noted in a recent work~\cite{2026geodiv}), whereas real images exhibit substantially greater visual variety--an issue that is notably less pronounced for the United States.
    }
    \label{fig:sd-prec-recall}
\end{figure*}

In this section, we study the images generated by the model with country-specified prompts. Precisely, for each entity-country pair, we generate $250$ images, and the prompts we use to generate the images are of the template `\texttt{photo of a \{entity\} in \{country\}}'. We recognize the differences between the used prompt template with the actual captions from the training data, which are often much more detailed. Hence, we only consider images from the dataset whose captions have $\leq 10$ tokens. We first find that the diversity of the synthetic images per entity (measured using the Vendi-Score method described in subsection~\ref{subsec: diversity}) is substantially lower than those of the real-images: the ratio of diversity of real vs synthetic images, averaged across entities and countries is $2.7$. This is perhaps because real images, obtained by scraping the internet, depict the entities in diverse contexts, whereas the synthetic ones are often closeups of the entity specified in the prompt. 

To further assess how well synthetic images reflect real-world visual distributions across geographic regions, we adopt the precision and recall metrics of Sajjadi et al.~\cite{sajjadi2018assessing} between the generated and real images of the entities per country. Precision measures the fraction of generated images whose embeddings lie within the local k-NN neighborhood of real images, capturing whether generated samples remain on the manifold of real data. Recall, in contrast, measures the fraction of real images that fall within the manifold induced by generated samples, quantifying how much of the real distribution is covered by the generator. Importantly, precision and recall disentangle realism from coverage: a model may produce visually convincing images (high precision) while failing to capture the semantic diversity of real data (low recall). This distinction is particularly critical for geographical analysis, where visual diversity within regions is high and long-tail semantics are common. Across countries and continents, we observe moderate-to-high precision but near-zero recall (see Figure~\ref{fig:sd-prec-recall}), indicating that synthetic images are moderately realistic and capture some country-specific cues, as also verified in the work of Basu et al.~\cite{Basu_2023_ICCV}. However, recall remains substantially lower than precision, revealing a stark gap between realism and coverage. This suggests that although the generated images appear realistic, the model repeatedly produces similar-looking images and fails to reflect the full range of visual appearances present in the real data.~\footnote{Note that the generation prompts specify only the \texttt{entity} and country, whereas captions in the real dataset may contain richer details. This difference in prompt specificity may partially contribute to the observed gap in recall, although the problem might be deeper than just having a more detailed prompt. We leave further investigation into the observed discrepancies in real vs synthetic images to future work.}

Our findings thus indicate an alarming precedent. Text-to-image models not only lack geographical representativeness in the case of underspecified prompts~\cite{Basu_2023_ICCV}, but also lack diversity in images across countries. A illustrative example can be seen in Figure~\ref{fig:sd-prec-recall}, where generated images of Indian cars appear dilapidated and old, suggesting stereotyping, even though the real images cover a diverse range of cars, an issue not seen in case of the US. This observation corroborates with the findings of Basu et al.~\cite{2026geodiv}, which highlights biased portrayals of countries by generative models, not only in terms of visual appearance of different entities and backgrounds but also from socio-economic angles. We show further examples of generated images of kitchens from Morocco and United States in the Appendix Figures~\ref{fig:morocco} and ~\ref{fig:us} respectively, where similar patterns are observed.

\section{Limitations} %
\label{sec:discussion}

In this section, we discuss the key limitations of our study.
We apply the \systemname framework to image–text pairs from Re-LAION2B-en~\citep{schuhmann2022laion}, DataComp~\citep{gadre2023datacomp}, and Conceptual Captions~\citep{sharma2018conceptual}—datasets that are commonly used to train large-scale text-to-image models. Our analysis primarily focuses on images paired with \textit{English} captions. While we additionally consider Spanish, Japanese, Greek, and Hindi captions from the Re-LAION2B-multi dataset, the relatively small number of languages studied remains a limitation; extending the analysis to a broader set of languages is an important direction for future work.
For non-English captions, we translate them into English using the NLLB-200-3.3B model~\citep{costa2022no}, followed by location extraction and country-prediction. Consequently, our analysis is sensitive to translation quality, and errors introduced during translation may propagate to downstream predictions. More broadly, although we observe consistent trends across datasets, our study is limited to three vision–language datasets. Furthermore, our estimates depend on the accuracy of predictions by the \systemname pipeline paired with the entity classifier.

In addition, several prominent text-to-image models, such as DALL·E~\citep{ramesh2022hierarchical} and Imagen~\citep{saharia2022photorealistic}, are trained on closed-source datasets. Although we analyze multiple large-scale datasets and can reasonably expect some overlap in data curation practices and training distributions, the extent to which our findings generalize to such closed datasets remains unclear. To facilitate such analysis on future datasets, we open-source our code.

Finally, \systemname produces geographical predictions only for captions that contain explicit or implicit geographical cues. Estimating locations from underspecified image–caption pairs remains a significant challenge. 
While modern vision–language models exhibit promising image-based geolocalization capabilities, there is currently no reliable way to validate their performance on existing web-scale vision–language datasets, which lack ground-truth annotations about their origins. Addressing this limitation likely requires changes in data curation practices: we encourage future dataset builders to document image provenance more systematically, whenever feasible, to enable supervised approaches to geolocalization. Lastly, it is important to note that although geolocalization raises important privacy concerns, our analysis is coarse-grained and does not involve identifying individuals or recovering precise GPS coordinates.

\section{Conclusion}
\label{sec: conclusion}

In this work, we analyzed the geographical distribution of captions in the Re-LAION2B-en, DataComp1B, and CC-12M datasets across $20$ entities. We found that the United States, the United Kingdom, and Canada appeared most frequently in location-specified captions, jointly accounting for $48\%$ of all such captions, while countries from South America and Africa were the least represented, appearing in only $1.8\%$ and $3.8\%$ of captions, respectively. Overall, more than $70\%$ of location-specified captions were predicted to originate from just $15$ countries, with $12$ of these top $15$ countries shared across all three datasets.
We further geolocalized captions in four non-English languages—Spanish, Greek, Hindi, and Japanese from the multilingual version of Re-LAION, and observed that the most frequently occurring countries for each language largely corresponded to regions where those languages are predominantly spoken. In addition, our analysis of country-wise image and caption diversity indicated a moderate correlation between a country’s frequency in the dataset and the diversity of its associated images, suggesting that higher representation did not necessarily translate to greater visual or semantic diversity. Finally, we observed that country-specific images generated by Stable Diffusion v1.3, trained on LAION, appeared visually realistic but exhibited limited and often stereotypical coverage when compared to the diversity present in real-world images.
These findings call for increased efforts in data collection and curation to ensure more diverse, global representation in vision-language training datasets.

\section*{Acknowledgements}

We are thankful to Soumya Dutta from the LEAP Lab, and Harsh Rangwani from the Vision and AI Lab, Indian Institute of Science Bangalore for their valuable feedback.
This work was supported in part by the AI2050 program at
Schmidt Sciences (Grant G-24-66186) and a grant from Google.

\bibliographystyle{unsrt}
\bibliography{neurips_2025}

\appendix
\section{Appendix}
\label{sec:appendix}

In this appendix, we begin by discussing the compute resources we use for geographically profiling the studied datasets (subsection~\ref{subsec:resources}). We next discuss the finer details about the entity-presence filtering step of the proposed system, including the details on the dataset selection for the crowdsourcing task, the survey details with inter-annotator agreements, and comparison with other methods (subsection~\ref{subsec:npf-app}). We next discuss the different approaches we explored for geolocalizing the image-caption pairs, their comparisons with our approach and the prompts used in each stage (subsection~\ref{subsec: geoprofile-2}). Subsection~\ref{subsec: datacomp-en} discusses details on the top $15$ most frequently occurring countries across datasets, percentage of underspecified captions per entity and dataset, as well as other related detail on distributional analysis. We present the distribution of continents in the studied datasets~\citep{schuhmann2022laion} for each entity (subsection~\ref{subsec: continent}) and some qualitative examples (subsection~\ref{subsec: q}). Finally, we end our discussion with some additional details on the comparison of the geographical compositions of the studied datasets with the real-world distributions with respect to selected entities (subsection~\ref{subsec: gr2}) and the diversity of images and their captions belonging to different countries (subsection~\ref{subsec:div_appendix}).

\begin{figure*}[h!]
    \centering
    \includegraphics[width = 0.85\linewidth]{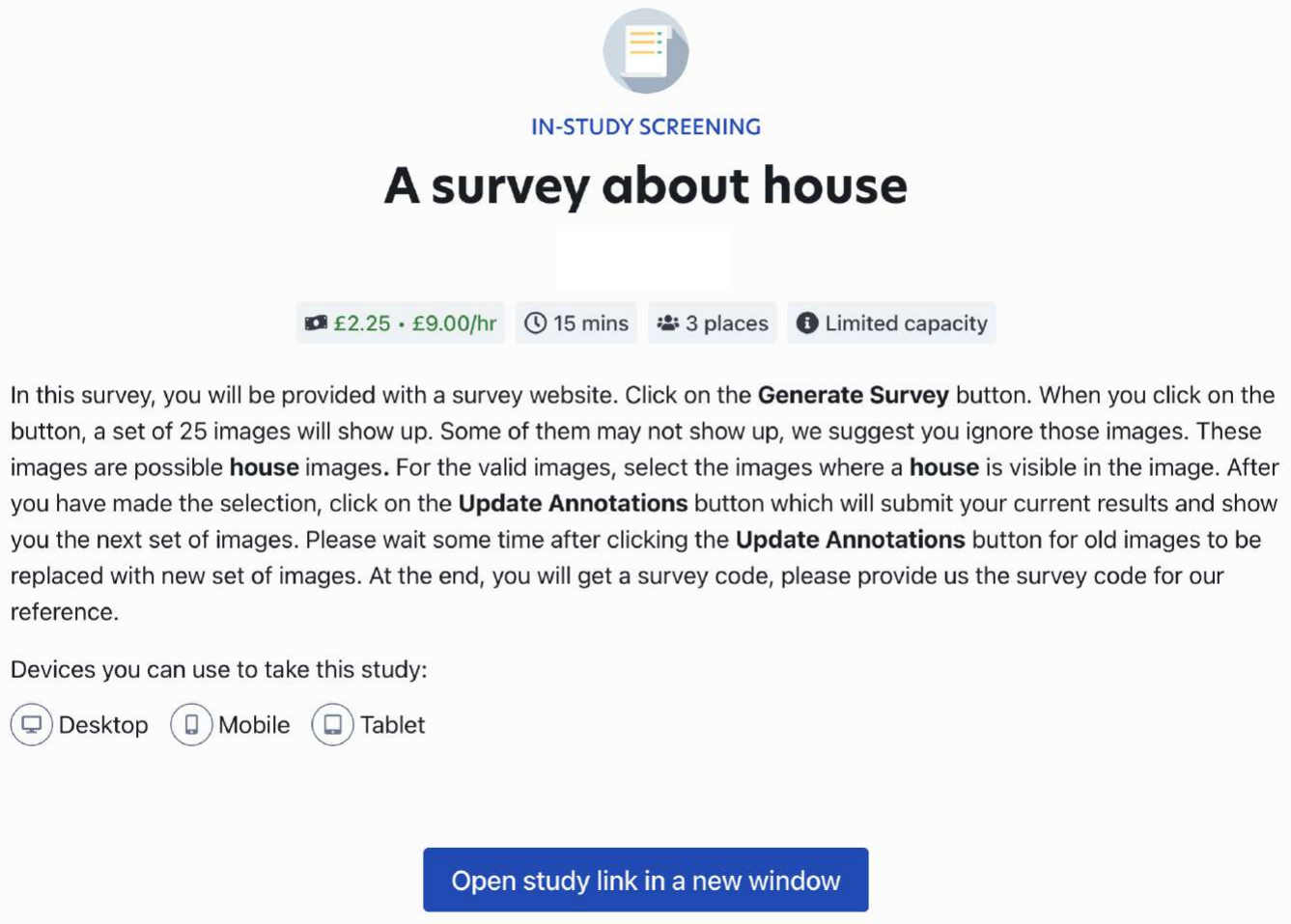}
    \caption{\textbf{Survey Instructions}. Given an entity $e$, we show an instruction sheet for the same, where we ask annotators to mark the shown images if they contain the same entity.}
    \label{fig:survey-1}
\end{figure*}

\subsection{Resources used by \systemname}
\label{subsec:resources}
We use a NVIDIA RTX A6000 GPU card to run the NLLB-200-3.3B model for translating multilingual captions to English, and other benchmarking tasks like using the BLIPv2 VQA model~\citep{li2023blip} and the CLIP model~\citep{radford2021learning} to evaluate on the crowd-annotated dataset. 
For the SVM classification performed during the Entity-Presence Classification stage, we use the default hyperparameters as provided in the python library \footnote{\url{https://scikit-learn.org/stable/modules/svm.html}}.

\noindent
\begin{figure*}[h!]
    \centering
    \includegraphics[trim=0cm 3cm 0cm 0cm, clip, width = 0.85\linewidth]{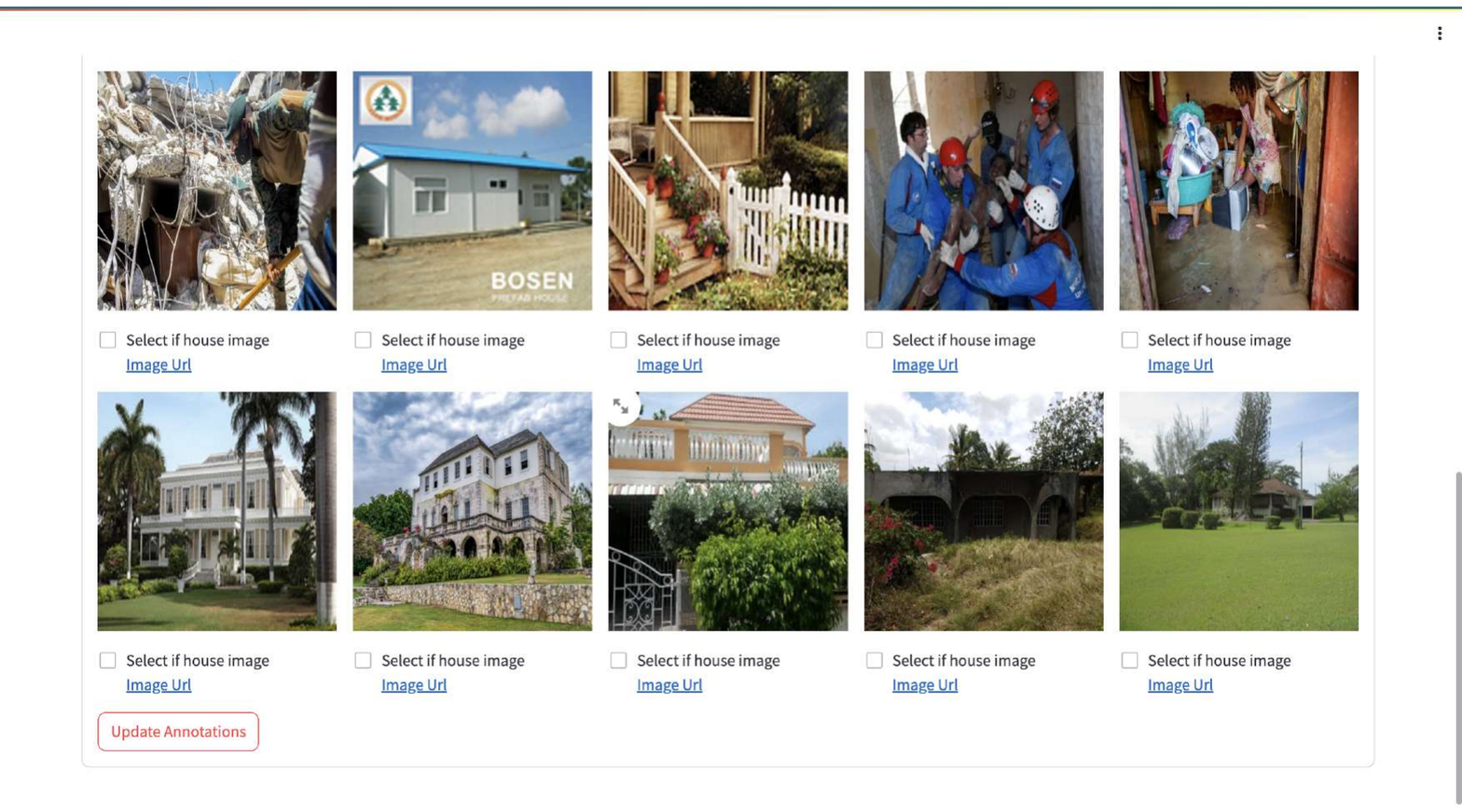}
    \caption{\textbf{Survey Main Page.} We show a set of $25$ images per page and prompt the user to press the Update Annotations button after the selections to show next set of images}
    \label{fig:survey-2}
\end{figure*}

\subsection{Entity-Presence Filtering - Further Details}
\label{subsec:npf-app}
The entity-presence filtering removes images that are irrelevant to the entity in question. Here, we present the different steps employed in this process: a) Creating a small dataset for annotation, b) Conducting a survey on crowdworkers with the annotated images. We next provide the overall inter-annotator agreements for each entity. Overall, we hire $3$ crowdworkers from the Prolific platform~\cite{prolific}, and pay each of them $\$2.45$ for the 15-minute survey.
\subsubsection{Annotation Dataset Creation}
\label{subsec:dataset creation}
We first select a small subset of images for each entity, and then hire crowdworkers to annotate them.
To ensure that the selected images are geodiverse, we divide the globe into $17$ regions as described by United Nations, and $4$ income groups as described by World Bank (see subsection~\ref{subsec: entity-presence}). The $17$ regions are as follows:
\emph{[`Northern Africa', `South America', `Sub-Saharan Africa', `Western Europe', `Australia and New Zealand', `Southern Europe', `Western Asia', `Eastern Europe', `Caribbean', `Central America', `Eastern Asia', `Oceania (excluding Australia and New Zealand)', `Northern America', `Southern Asia', `Central Asia', `South-eastern Asia']}
whereas the Income Groups are \emph{[`Low Income', `Lower Middle Income', `Upper Middle Income', `High Income']}. Using the country predictions of the \syscomma framework, we select equal number of images from each region and income group (or all images in case the number of images is less for a certain region and income group combination). For proper evaluation, our test set has both in-distribution (ID) data (i.e., images from countries already seen in the training set), and out-of-distribution (OOD) data (i.e., images from countries not seen in the training set).
The following set of countries make up the OOD test set: \emph{[`China', `Chile', `New Zealand', `Philippines', `Spain']}.

\subsubsection{Survey Details}
\label{subsec:survey}
For each entity $e$, we host a survey with the selected images, and appoint $3$ annotators from the Prolific platform~\citep{prolific} to mark the images containing the entity. To each annotator, we ask the following question:
``\texttt{In this survey, you will be provided with a survey website. Click on the Generate Survey button. When you click on the button, a set of 25 images will show up. Some of them may not show up, we suggest you ignore those images. These images are possible \textit{\{e\}} images. For the valid images, select the images where a \textit{\{e\}} is visible in the image. After you have made the selection, click on the Update Annotations button which will submit your current results and show you the next set of images. Please wait some time after clicking the Update Annotations button for old images to be replaced with new set of images. At the end, you will get a survey code, please provide us the survey code for our reference.}''
We provide the screenshots of the instructions and the images used in the survey for houses in Figures ~\ref{fig:survey-1} \& ~\ref{fig:survey-2}.

\begin{table*}[h!]
\centering
\caption{
\textbf{Fleiss' Kappa and agreement scores across annotators}. After conducting the surveys, we find the mean agreement (over both classes) among all pairs of crowdworkers to be $83.6\%$.}
\label{tab:annotators}
\begin{tabular}{l|c|c|c|c}
\toprule
entity & $\kappa$ & \begin{tabular}[c]{@{}c@{}}Avg agreement for \\ class $0$ (\%) \end{tabular} & \begin{tabular}[c]{@{}c@{}}Avg agreement for \\ class $1$ (\%) \end{tabular} & Overall Agreement (\%) \\ \midrule
house & $0.7$ & $85.2$ & $86$ & $85.6$ \\
car & $0.8$ & $87.1$ & $90.2$ & $88.6$ \\
road & $0.7$ & $84.6$ & $81.6$ & $83.1$ \\
beach & $0.6$ & $86.5$ & $79.8$ & $83.2$ \\
flag & $0.6$ & $68.1$ & $91.3$ & $79.7$ \\
hotel & $0.6$ & $84$ & $74.8$ & $79.4$ \\
toilet & $0.8$ & $93.8$ & $85.7$ & $89.8$ \\
bedroom & $0.8$ & $97.2$ & $85.8$ & $91.5$ \\
kitchen & $0.8$ & $92.5$ & $84$ & $88.2$ \\
apartment & $0.5$ & $62.9$ & $93.4$ & $78.2$ \\
mountain & $0.7$ & $84.1$ & $87.6$ & $85.8$ \\
hill & $0.7$ & $85.0$ & $86.4$ & $85.7$ \\
valley & $0.6$ & $84.8$ & $73.7$ & $79.2$ \\
wedding dress & $0.6$ & $69.1$ & $92.4$ & $80.7$ \\
shop & $0.6$ & $85.1$ & $75.0$ & $80.0$ \\
office & $0.7$ & $85.0$ & $82.9$ & $83.9$ \\
lake & $0.6$ & $69.4$ & $91.6$ & $80.5$ \\
living room & $0.8$ & $86.9$ & $89.7$ & $88.3$\\
jewelry & $0.6$ & $70.0$ & $89.6$ & $79.8$ \\
island & $0.6$ & $78.5$ & $82.5$ & $80.5$ \\
\bottomrule
\end{tabular}
\end{table*}

\begin{table*}[h!]
\centering
\small
\caption{\textbf{Entity-wise details on the number of images annotated}. Alongside the number of images annotated, we also report the number of images marked with and without the entity.}
\label{tab:annotation-image}
\begin{tabular}{@{}l|l|l|l|l|l|l|l|l|l|l@{}}
\toprule
 & \begin{tabular}[c]{@{}l@{}}Apart-\\ ment\end{tabular} & \begin{tabular}[c]{@{}l@{}}Kit-\\ chen\end{tabular} & \begin{tabular}[c]{@{}l@{}}Ho-\\ use\end{tabular} & \begin{tabular}[c]{@{}l@{}}Ho-\\ tel\end{tabular} & \begin{tabular}[c]{@{}l@{}}Toi-\\ let\end{tabular} & \begin{tabular}[c]{@{}l@{}}Bed-\\ room\end{tabular} & \begin{tabular}[c]{@{}l@{}}Ro-\\ ad\end{tabular} & \begin{tabular}[c]{@{}l@{}}Be-\\ ach\end{tabular} & Car & \begin{tabular}[c]{@{}l@{}}Fl-\\ ag\end{tabular} \\ \midrule
Total Num & 577 & 526 & 651 & 660 & 554 & 585 & 640 & 590 & 634 & 660 \\
Present & 507 & 172 & 334 & 293 & 161 & 152 & 300 & 244 & 370 & 535 \\
Absent & 70 & 354 & 317 & 367 & 393 & 433 & 340 & 346 & 264 & 125 \\ \bottomrule
\toprule
 & Lake & \begin{tabular}[c]{@{}l@{}}Moun-\\ tain\end{tabular} & Hill & Shop & \begin{tabular}[c]{@{}l@{}}Isl-\\ and\end{tabular} & \begin{tabular}[c]{@{}l@{}}Living\\ room\end{tabular} & \begin{tabular}[c]{@{}l@{}}Off-\\ ice\end{tabular} & \begin{tabular}[c]{@{}l@{}}Jewe-\\ lry\end{tabular} & Valley & \begin{tabular}[c]{@{}l@{}}Wedding\\ Dress\end{tabular} \\ \midrule
Total Num & 680 & 691 & 682 & 689 & 628 & 572 & 676 & 676 & 684 & 560 \\
Present & 437 & 425 & 308 & 266 & 190 & 424 & 289 & 512 & 272 & 502 \\
Absent & 243 & 266 & 374 & 423 & 438 & 148 & 387 & 121 & 412 & 58 \\ \bottomrule
\end{tabular}
\end{table*}

\subsubsection{Inter-Annotator Agreements}
\label{subsec:inter}
For each entity $e$, we show the crowdworkers $550-650$ images.
To ensure that the annotators are trustworthy, we initially conduct a survey on a smaller dataset (which we annotate by ourselves), and verify the honesty of the crowdworkers by matching their annotations with ours. 
Finally, we filter out the annotators which have an absolute agreement  $<70\%$. 
For the real survey, we employ $3$ annotators for each entity, and select the final annotation by majority voting among the received votes.
Each such survey is conducted for a duration of around 15 minutes, and each crowdworker is paid at the rate of $\$9.85$ per hour.
Here, we discuss the inter-annotator agreement among the crowdworkers for each entity, using Fleiss' Kappa ~\citep{fleiss1971measuring}.
Fleiss' Kappa $\kappa$ measures the reliability of agreement between multiple raters for categorical data and is ideal for our use case as it handles fixed numbers of raters with complete ratings effectively. 
According to Fleiss's interpretation, $\kappa$ between $0.41-0.60$ is considered moderate,  a $\kappa$ between $0.61-0.80$ as substantial and a $\kappa$ between $0.81-1$ as almost perfect agreement. 
We further measure average agreement between every pair of annotators for both positive and negative classes. To calculate the agreement values for a class, we find the percentage of images for which the pair of annotators agree. Specifically, we define the agreement between a pair of annotators $(i, j)$ as $(\frac{\text{A}_{(i,j)}^{c}}{\text{N}_{j}^{c}})*100$ where $\text{A}_{(i, j)}^{c}$ is the number of images where both annotators $i$ and $j$ agree on the class $c$ and ${\text{N}_{j}^{c}}$ refers to the total images where annotator $j$ marked it as class $c$. We find the average over all combinations of ordered pairs of annotators for the class $c$ for each entity $n$ and report this value. We also report the overall agreement which is calculated as the average of the class-wise values.
We present the $\kappa$ values, average agreement for classes $0$ and $1$ and also the overall agreement value, averaged across these classes for each entity in Table~\ref{tab:annotators}. Further, for each entity, the total number of images annotated and the number of images with and without the entity is presented in Table~\ref{tab:annotation-image}. It is to be remembered that to curate the annotation dataset for each entity, we uniformly sample geo-diverse images for a given entity from the country-tagged image-caption pairs from a combination of $17$ regions and $4$ income groups across the world. Since an equal number of images from all region-income group combinations may be unavailable, the total number of images varies across entities.
\noindent

\begin{table*}[h!]
\centering
\caption{\textbf{Performance of image recognition models across entities}. We evaluate the \emph{f1-scores} of positives on crowd-annotated ID and OOD test sets for each entity on: a) CLIP \citep{radford2021learning} zero-shot prompting with a negative and positive prompt, b) the BLIPv2~\citep{li2023blip} model, c) SVM model trained on the GeoDE dataset~\citep{ramaswamy2024geode}, d) SVM model trained on the crowd-annotated training set. 
For all 20 entities, the SVM classifier trained on the crowd-annotated dataset outperforms the other three methods.}
\small
\label{tab:recognition}
\begin{tabular}{c|cc|cc|cc|cccc}
\toprule
\multicolumn{1}{c|}{entity} & \multicolumn{2}{c|}{CLIP} & \multicolumn{2}{c|}{BLIP} & \multicolumn{2}{c|}{SVM (GeoDE)} & \multicolumn{2}{c}{SVM (Ours)} \\
\multicolumn{1}{c|}{} & \multicolumn{1}{c}{ID} & \multicolumn{1}{c|}{OOD} & \multicolumn{1}{c}{ID} & \multicolumn{1}{c|}{OOD} & \multicolumn{1}{c}{ID} & \multicolumn{1}{c|}{OOD} & \multicolumn{1}{c}{ID} & \multicolumn{1}{c}{OOD} \\
\midrule
House & $0.75$ & $0.78$ & $0.79$ & \multicolumn{1}{c|}{$0.89$} & $0.77$ & $0.88$ & \textbf{$\mathbf{0.85}$} & \textbf{$\mathbf{0.92}$} & \\
Flag & $0.66$ & $0.72$ & $0.84$ & $0.82$ & \textbf{$\mathbf{0.95}$} & $0.94$ & $0.92$ & \textbf{$\mathbf{0.94}$} & \\
Car & $0.82$ & $0.76$ & $0.88$ & $0.74$ & $0.83$ & $0.88$ & \textbf{$\mathbf{0.89}$} & \textbf{$\mathbf{0.98}$} & \\
Kitchen & $0.70$ & $0.76$ & $0.78$ & $0.80$ & NA & NA & \textbf{$\mathbf{0.93}$} & \textbf{$\mathbf{0.88}$} & \\
Beach & $0.80$ & $0.61$ & $0.80$ & $0.67$ & NA & NA & \textbf{$\mathbf{0.85}$} & \textbf{$\mathbf{0.81}$} & \\
Road & $0.63$ & $0.59$ & $0.75$ & \textbf{$\mathbf{0.91}$} & NA & NA & \textbf{$\mathbf{0.80}$} & $0.88$ & \\
Hotel & $0.75$ & $0.76$ & $0.85$ & $0.78$ & NA & NA & \textbf{$\mathbf{0.89}$} & \textbf{$\mathbf{0.90}$} & \\
Bedroom & $0.37$ & $0.62$ & $0.67$ & $0.80$ & NA & NA & \textbf{$\mathbf{0.82}$} & \textbf{$\mathbf{0.90}$} \\
Toilet & $0.65$ & $0.67$ & $0.73$ & $0.71$ & NA & NA & \textbf{$\mathbf{0.93}$} & \textbf{$\mathbf{0.75}$} & \\
Apartment & $0.91$ & $0.87$ & $0.95$ & $0.88$ & NA & NA & \textbf{$\mathbf{0.95}$} & \textbf{$\mathbf{0.98}$} & \\
Mountain & $0.78$ & $0.70$ & $0.90$ & $0.78$ & NA & NA & \textbf{$\mathbf{0.93}$} & \textbf{$\mathbf{0.82}$} & \\
Lake & $0.74$ & $0.72$ & $0.89$ & $0.82$ & NA & NA & \textbf{$\mathbf{0.95}$} & \textbf{$\mathbf{0.84}$} & \\
Hill & $0.76$ & $0.73$ & $0.80$ & $0.83$ & NA & NA & \textbf{$\mathbf{0.88}$} & \textbf{$\mathbf{0.95}$} & \\
Shop & $0.64$ & $0.67$ & $0.71$ & $0.69$ & NA & NA & \textbf{$\mathbf{0.88}$} & \textbf{$\mathbf{0.89}$} & \\
Wedding Dress & $0.81$ & $0.87$ & $0.55$ & $0.61$ & NA & NA & \textbf{$\mathbf{0.95}$} & \textbf{$\mathbf{0.92}$} & \\
Island & $0.67$ & $0.61$ & $0.71$ & $0.69$ & NA & NA & \textbf{$\mathbf{0.80}$} & \textbf{$\mathbf{0.84}$} & \\
Living Room & $0.54$ & $0.47$ & $0.82$ & $0.88$ & NA & NA & \textbf{$\mathbf{0.86}$} & \textbf{$\mathbf{0.96}$} & \\
Valley & $0.65$ & $0.65$ & $0.71$ & $0.76$ & NA & NA & \textbf{$\mathbf{0.81}$} & \textbf{$\mathbf{0.85}$} & \\
Jewelry & $0.53$ & $0.60$ & $0.61$ & $0.68$ & NA & NA & \textbf{$\mathbf{0.92}$} & \textbf{$\mathbf{0.90}$} & \\
Office & $0.56$ & $0.54$ & $0.67$ & $0.66$ & NA & NA & \textbf{$\mathbf{0.81}$} & \textbf{$\mathbf{0.73}$} & \\
\bottomrule
\end{tabular}
\end{table*}

\subsubsection{Benchmarking the Entity-Presence Classifier}
\label{subsec:benchmark-svm}
In section~\ref{subsec: entity-presence}, we mention that we train an SVM model using the CLIP features of the crowd-annotated images to predict if the entity is present or absent in them. 
Here, we compare its performance with other methods: a) the zero-shot CLIP~\citep{radford2021learning} model, b) the zero-shot BLIPv2~\citep{li2023blip} model, c) an SVM model trained on the geo-diverse \textit{GeoDE}~\citep{ramaswamy2024geode} dataset.

The $f1$-score for each model is presented in Table~\ref{tab:recognition}. Recall that the test set consists of both ID and OOD subsets. For the CLIP model, we compare each image of a certain entity by the following text prompts: ``\texttt{Photo of a \{entity\}}'', and ``\texttt{Not a photo of a \{entity\}}'', and based on the similarity of the image with these two prompts, we assign label $0$ to it if it is more similar to the latter prompt, else we assign $1$.
We also evaluate the VQA model of BLIP, and for each image of an entity, we ask the following question: ``\texttt{Is this a photo of any \{entity\}}?''. We again assign $0$ to the image if the answer is `no', otherwise $1$ is assigned. 
Additionally, we evaluate an SVM model trained on the GeoDE dataset, which has images from different parts of the world with respect to multiple entities. The only entities that are common with our paper, are house, flag and car. For all the compared methods involving CLIP, BLIP and the GeoDE dataset respectively, we notice that the classifier trained on the crowd-annotated dataset surpasses them in terms of $f1$-score for both the ID and OOD subsets. This demonstrates the necessity of the crowd-annotation step in order to train the entity presence classifier.

\subsubsection{Examples of Irrelevant Images}
We show examples of relevant and irrelevant images as identified by our entity-presence classifier for each entity in Table~\ref{tab:images-qualitative}, further demonstrating the requirement of the entity-presence classifier in the proposed tool.
\begin{longtable}{|>{\centering\arraybackslash}m{1.5cm}|>{\centering\arraybackslash}m{5cm}|>{\centering\arraybackslash}m{5cm}|}

\caption{
\textbf{Irrelevant \& Relevant Images}
}
\label{tab:images-qualitative} \\
\toprule
& Irrelevant & Relevant \\ \midrule
\endfirsthead

\multicolumn{3}{c}{{\tablename\ \thetable{} -- continued from previous page}} \\ \midrule
& Irrelevant & Relevant \\ \midrule
\endhead

\midrule \multicolumn{3}{r}{{Continued on next page}} \\ \bottomrule
\endfoot

\bottomrule
\endlastfoot

House & \includegraphics[width=0.8\linewidth, height=3.5cm]
{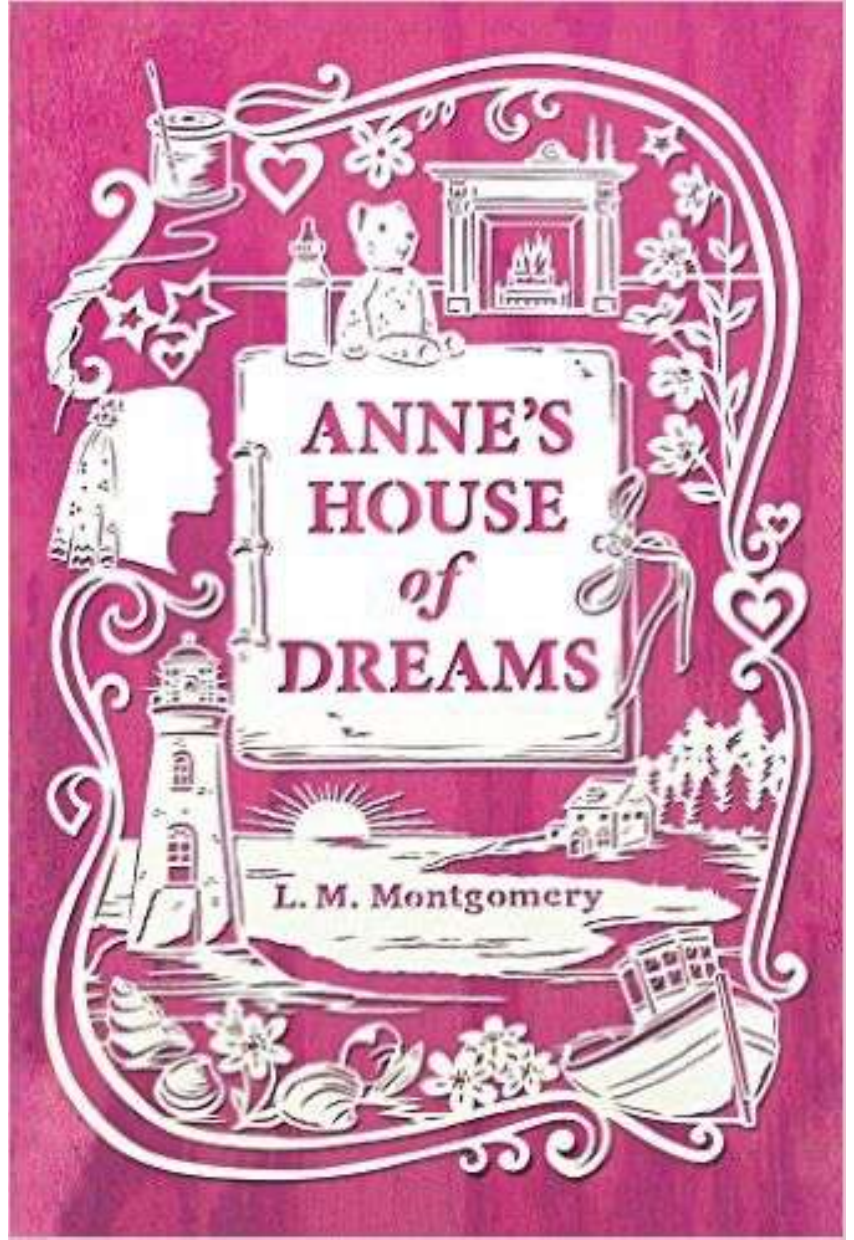} & \includegraphics[width=0.8\linewidth, height=3.5cm]{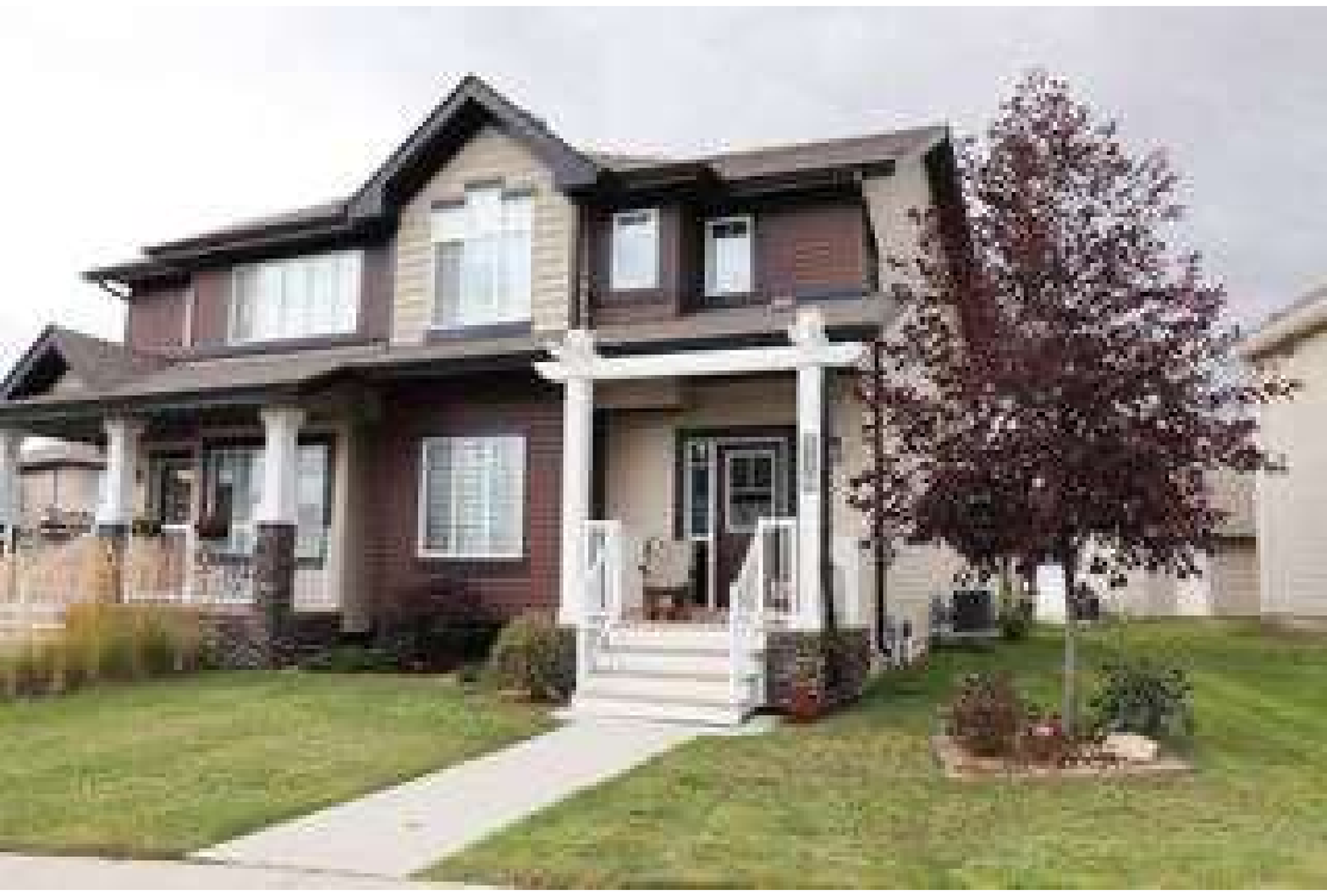} \\ \midrule
Flag & \includegraphics[width=0.8\linewidth, height=4cm]
{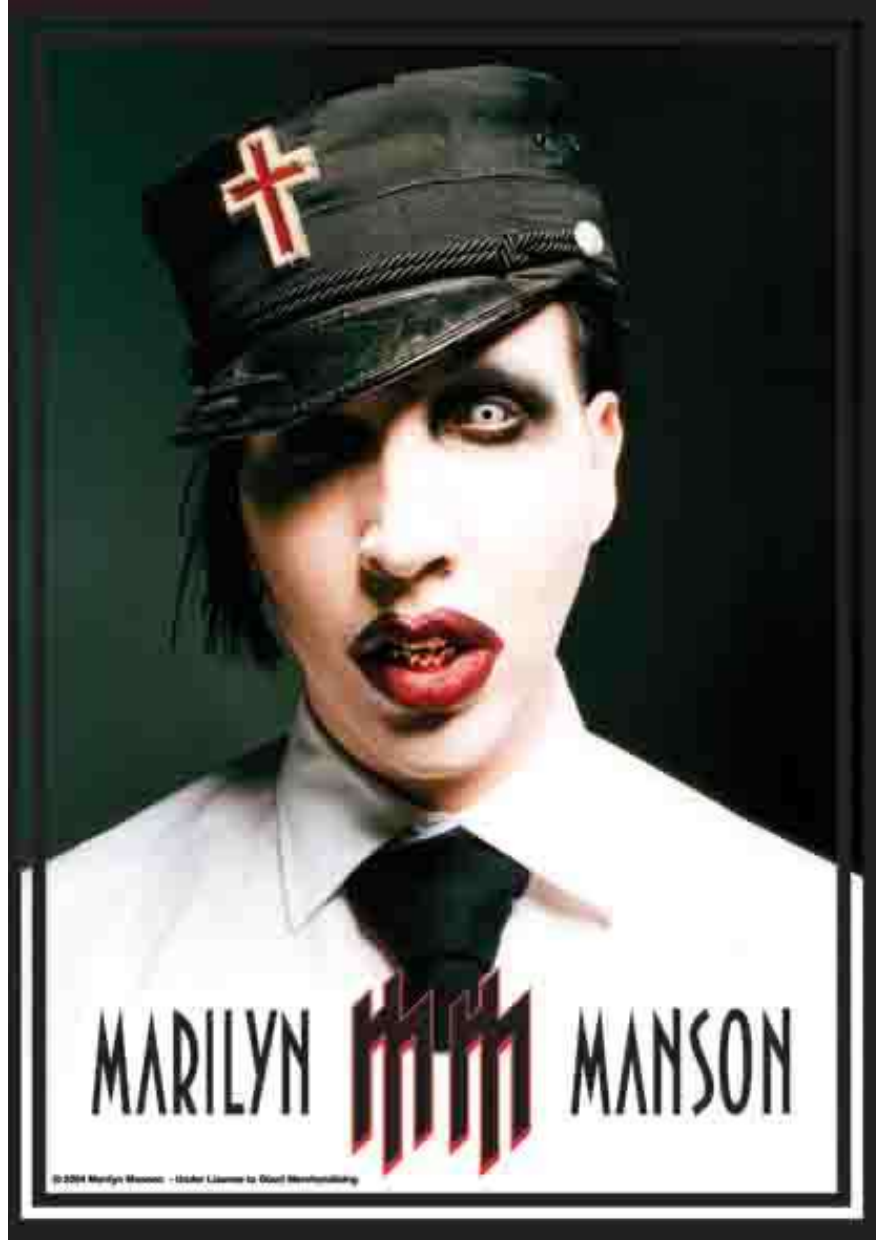} & \includegraphics[width=0.8\linewidth, height=3.5cm]{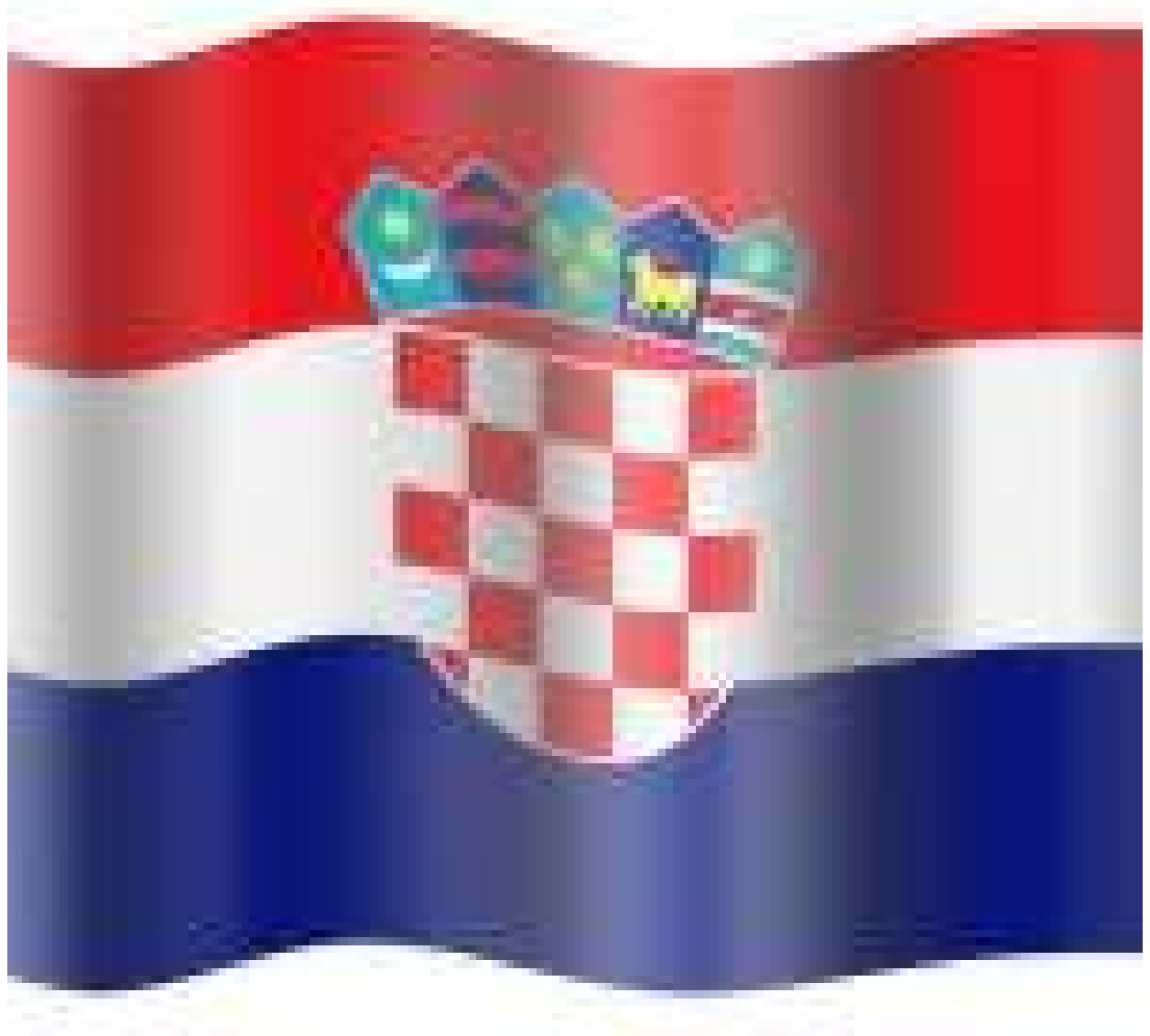} \\ \midrule
Car & \includegraphics[width=0.8\linewidth, height=3.5cm]
{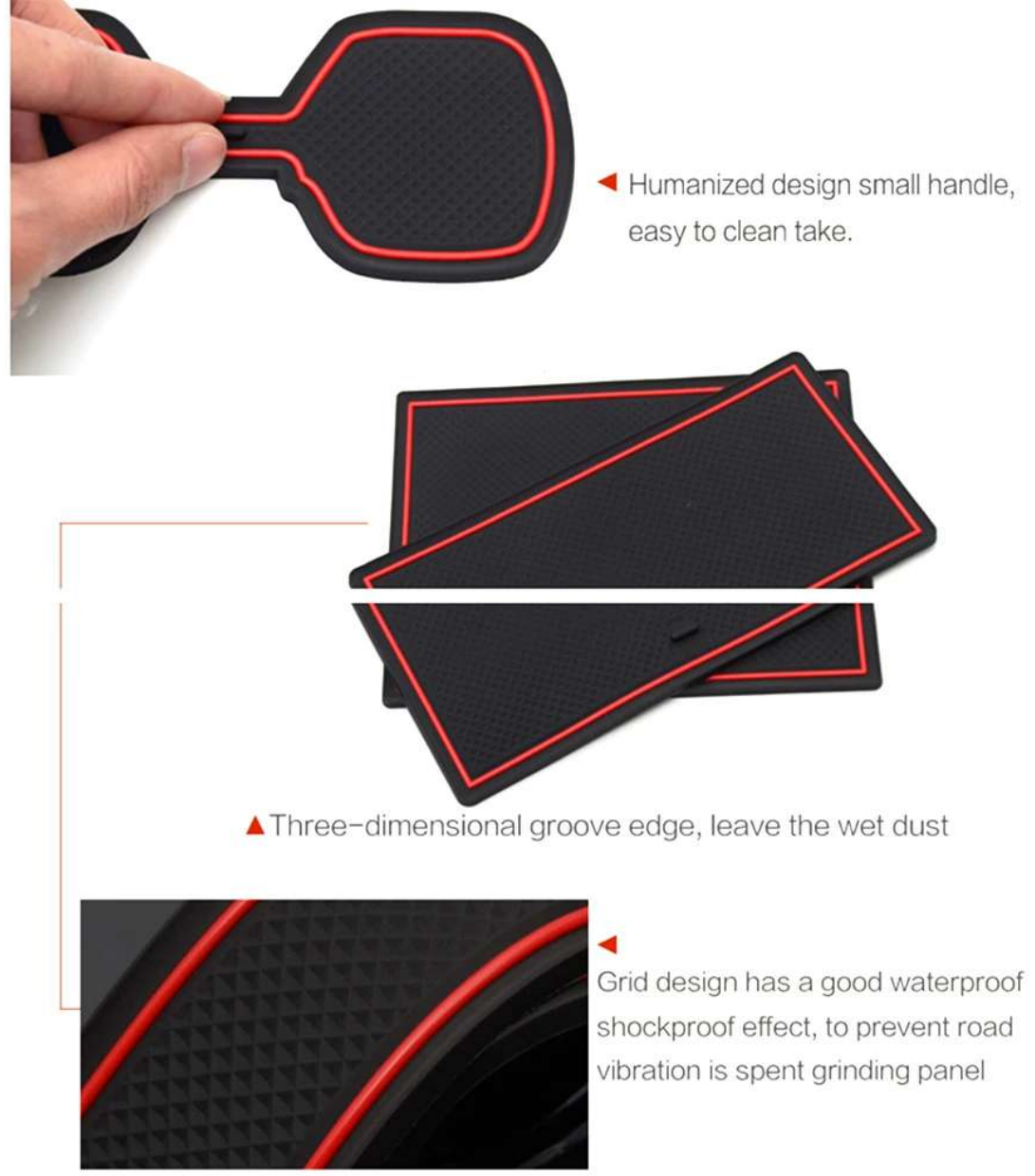} & \includegraphics[width=0.8\linewidth, height=3.5cm]{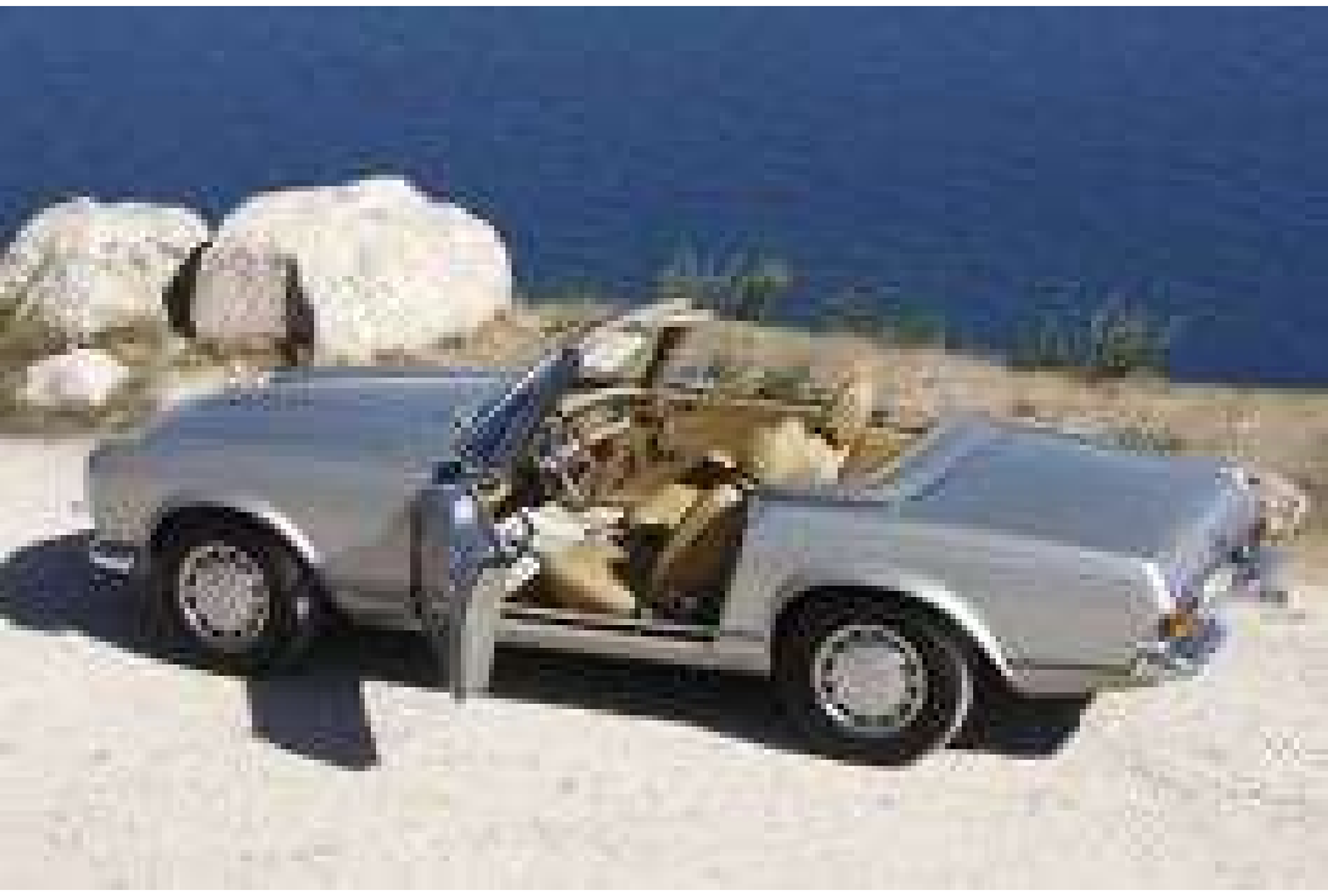} \\ \midrule
Kitchen & \includegraphics[width=0.8\linewidth, height=3.5cm]
{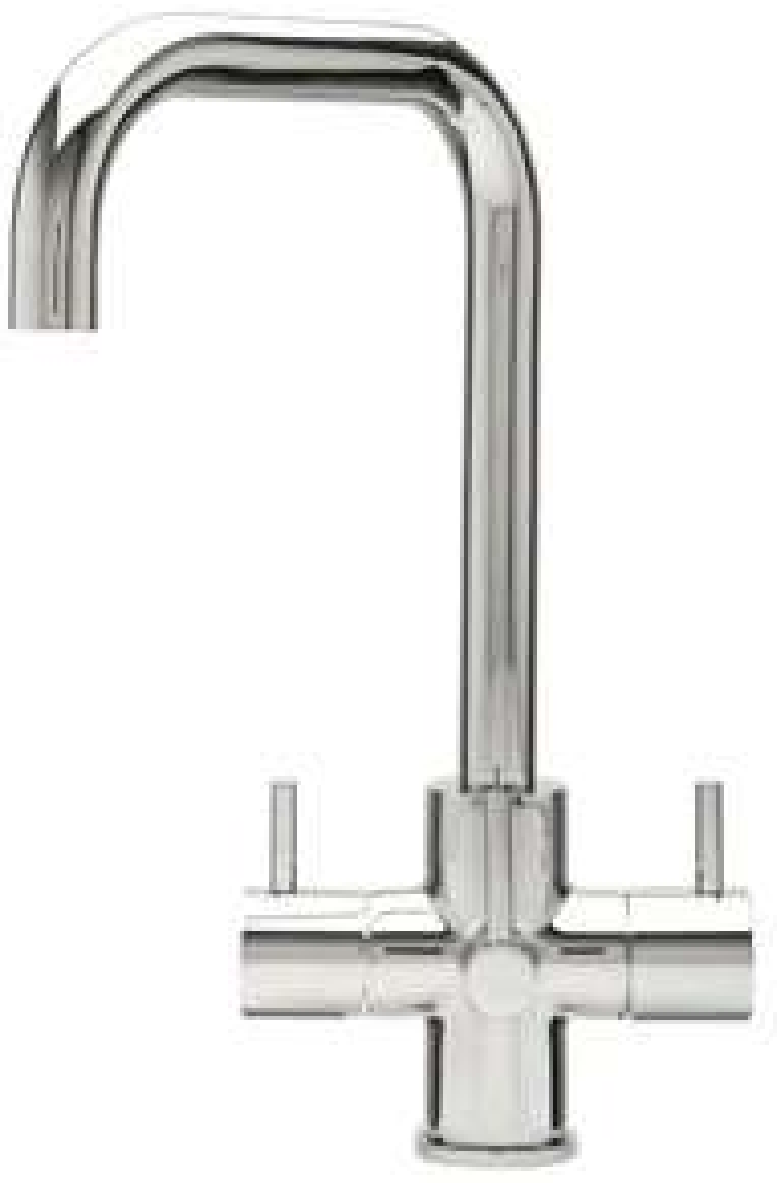} & \includegraphics[width=0.8\linewidth, height=3.5cm]{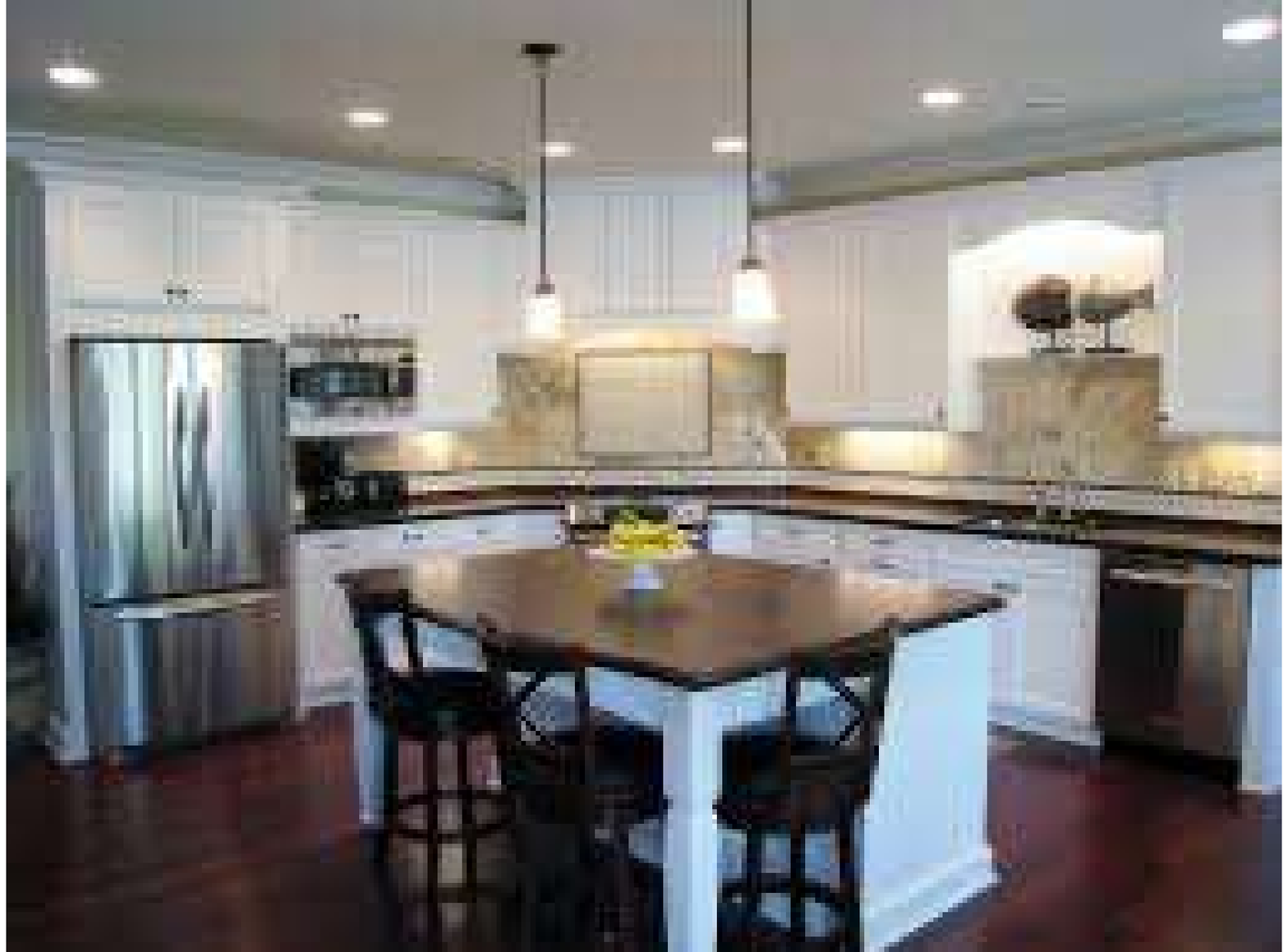} \\ \midrule
Road & \includegraphics[width=0.8\linewidth, height=3.5cm]
{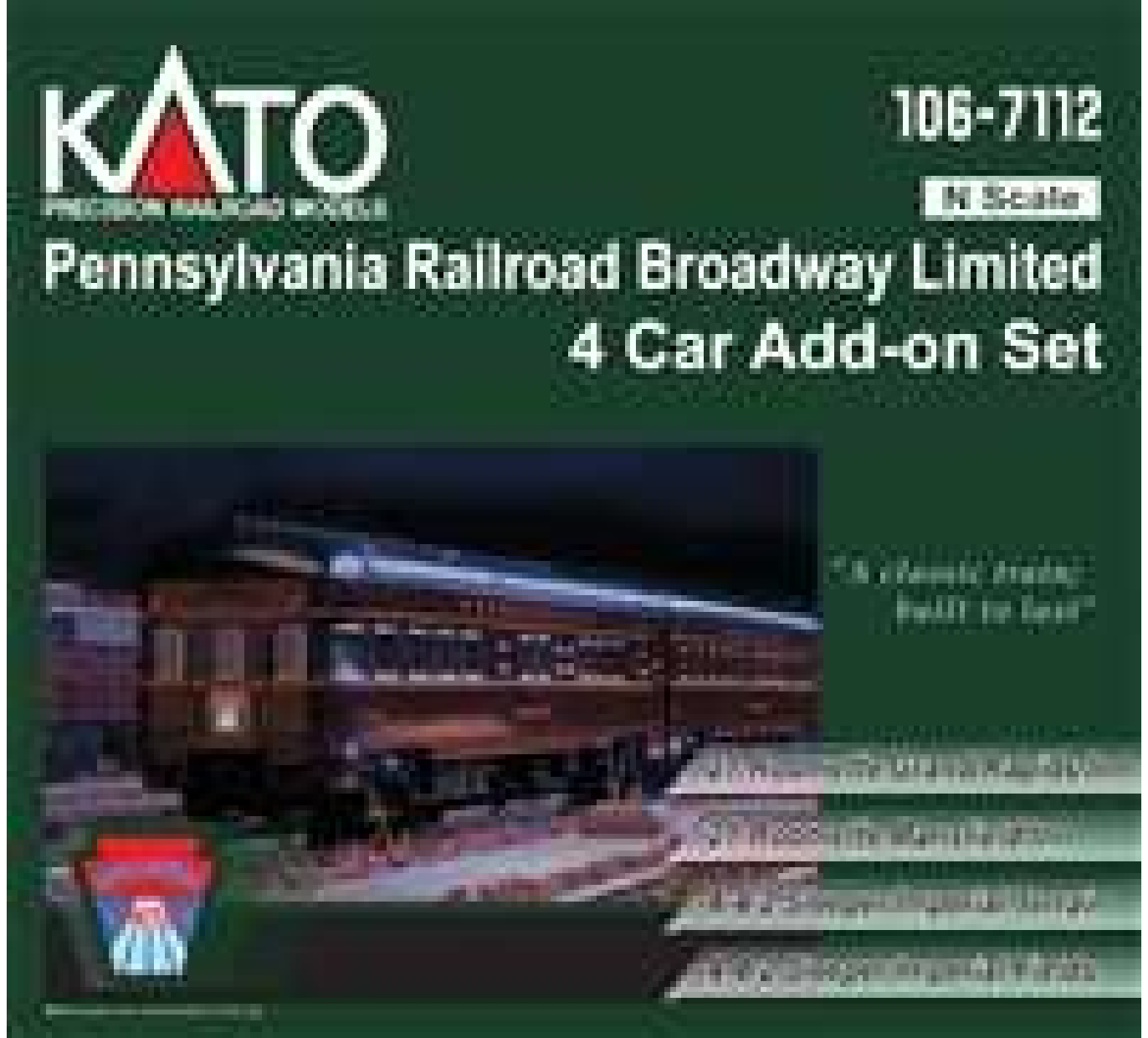} & \includegraphics[width=0.8\linewidth, height=3.5cm]{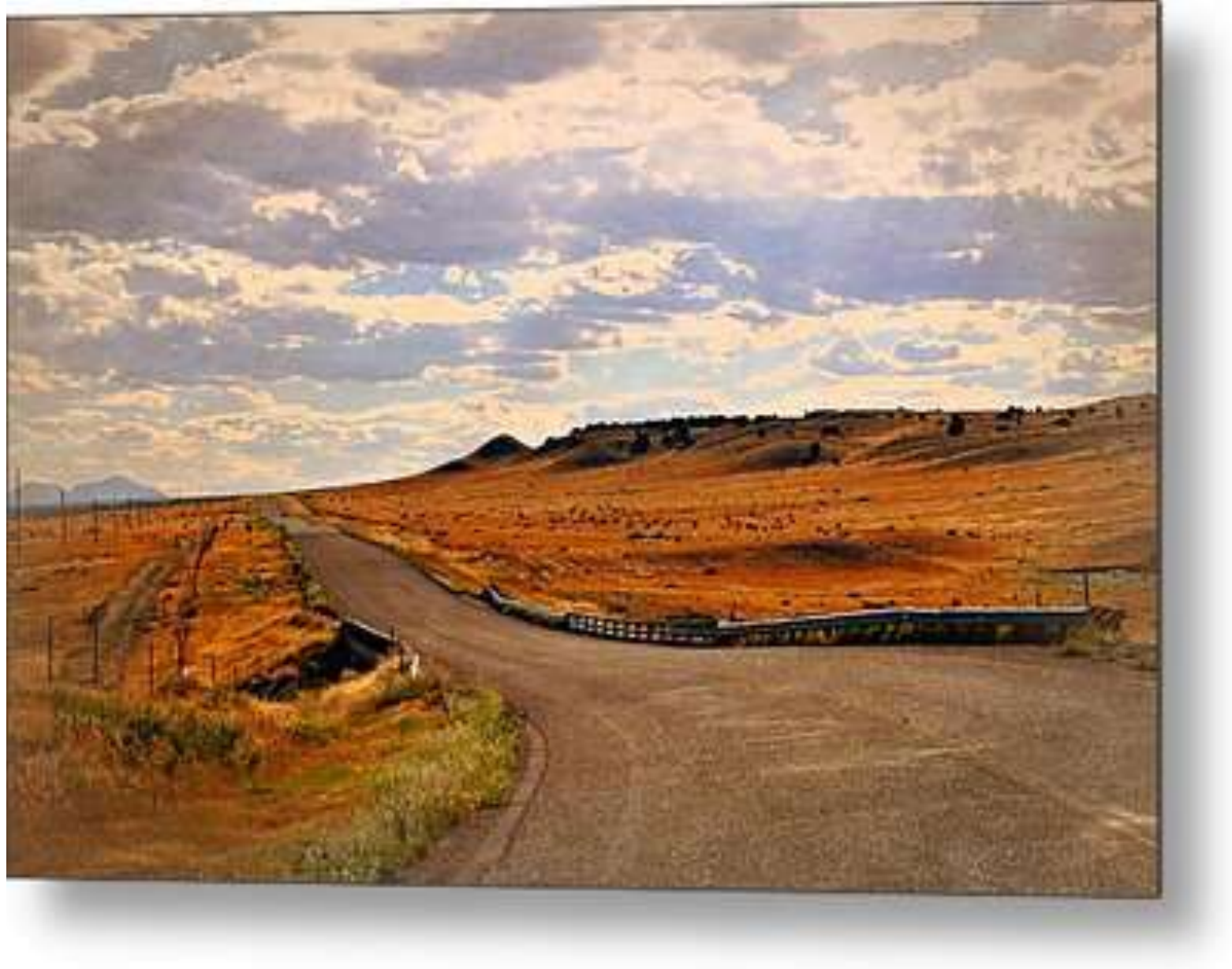} \\ \midrule
Beach & \includegraphics[width=0.8\linewidth, height=3.5cm]
{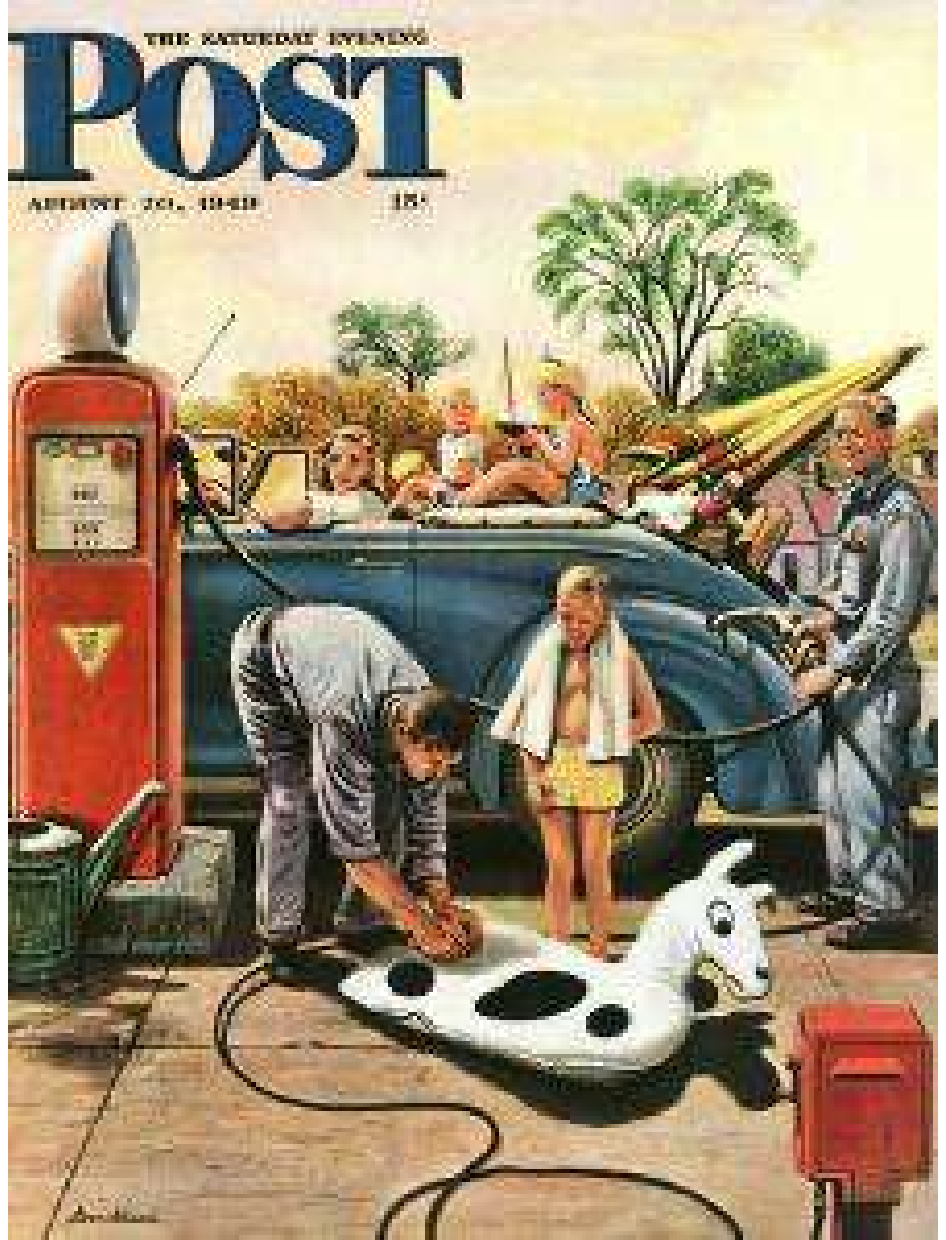} & \includegraphics[width=0.8\linewidth, height=3.5cm]{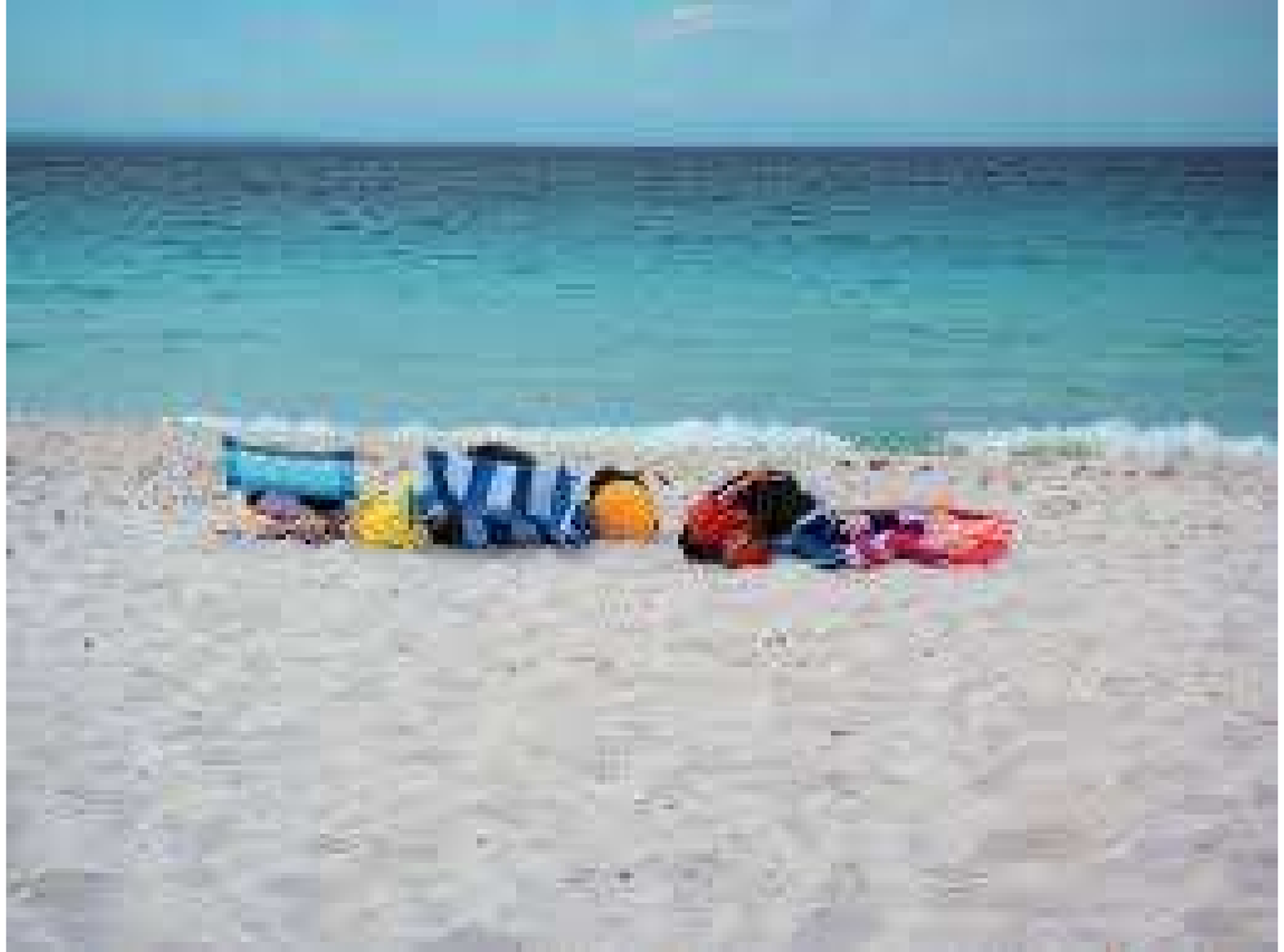} \\ \midrule
Hotel & \includegraphics[width=0.8\linewidth, height=3.5cm]
{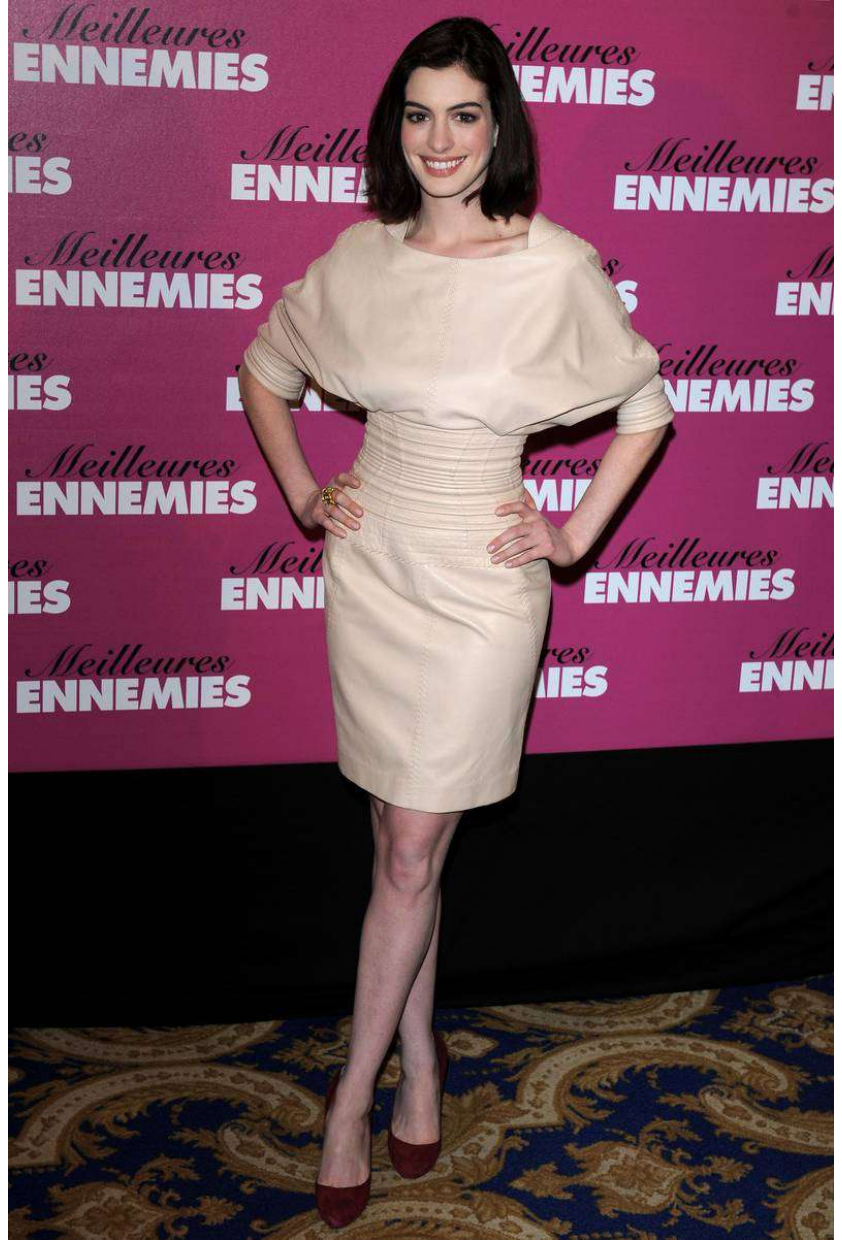} & \includegraphics[width=0.8\linewidth, height=3.5cm]{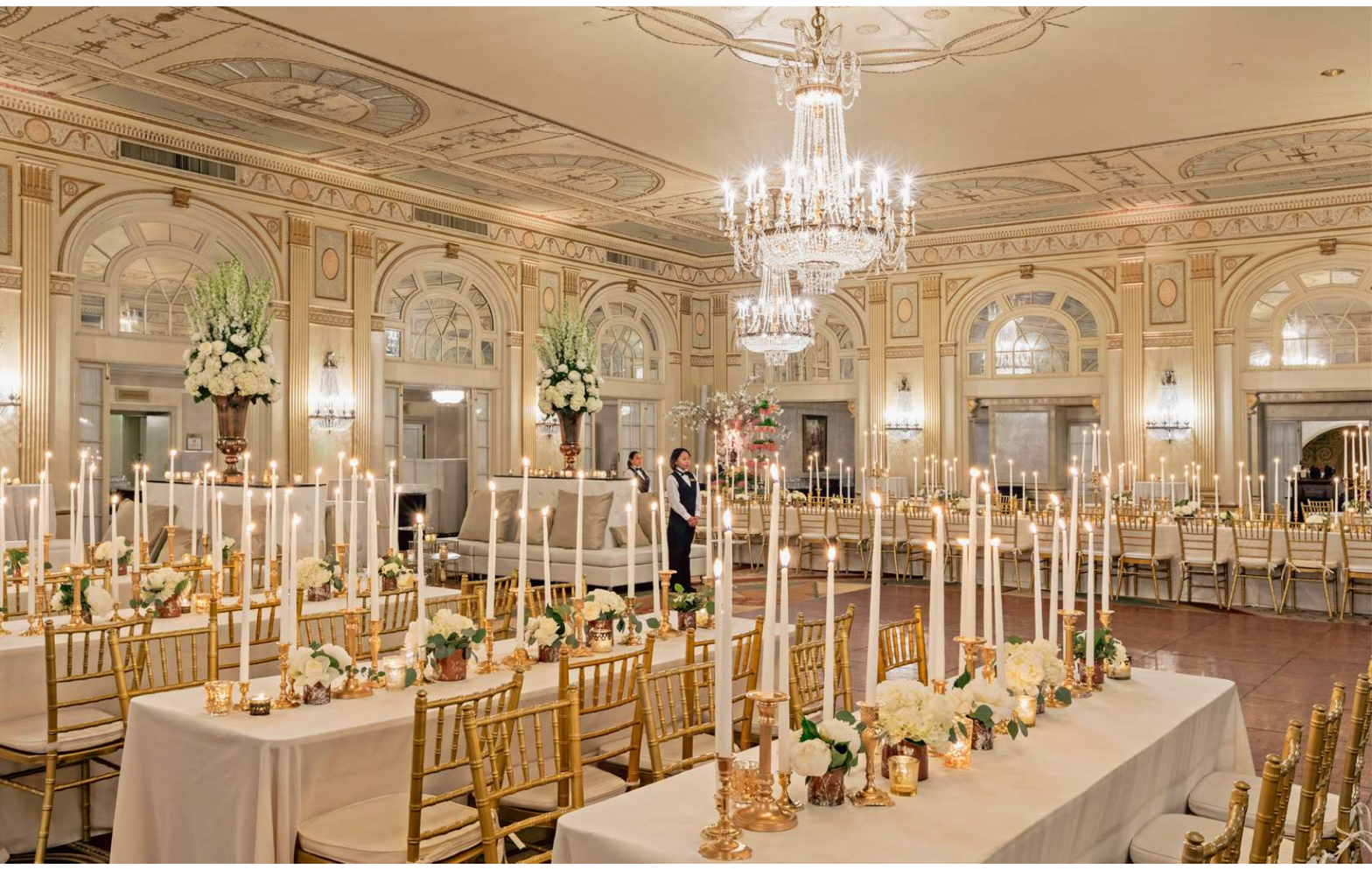} \\ \midrule
Toilet & \includegraphics[width=0.8\linewidth, height=3.5cm]
{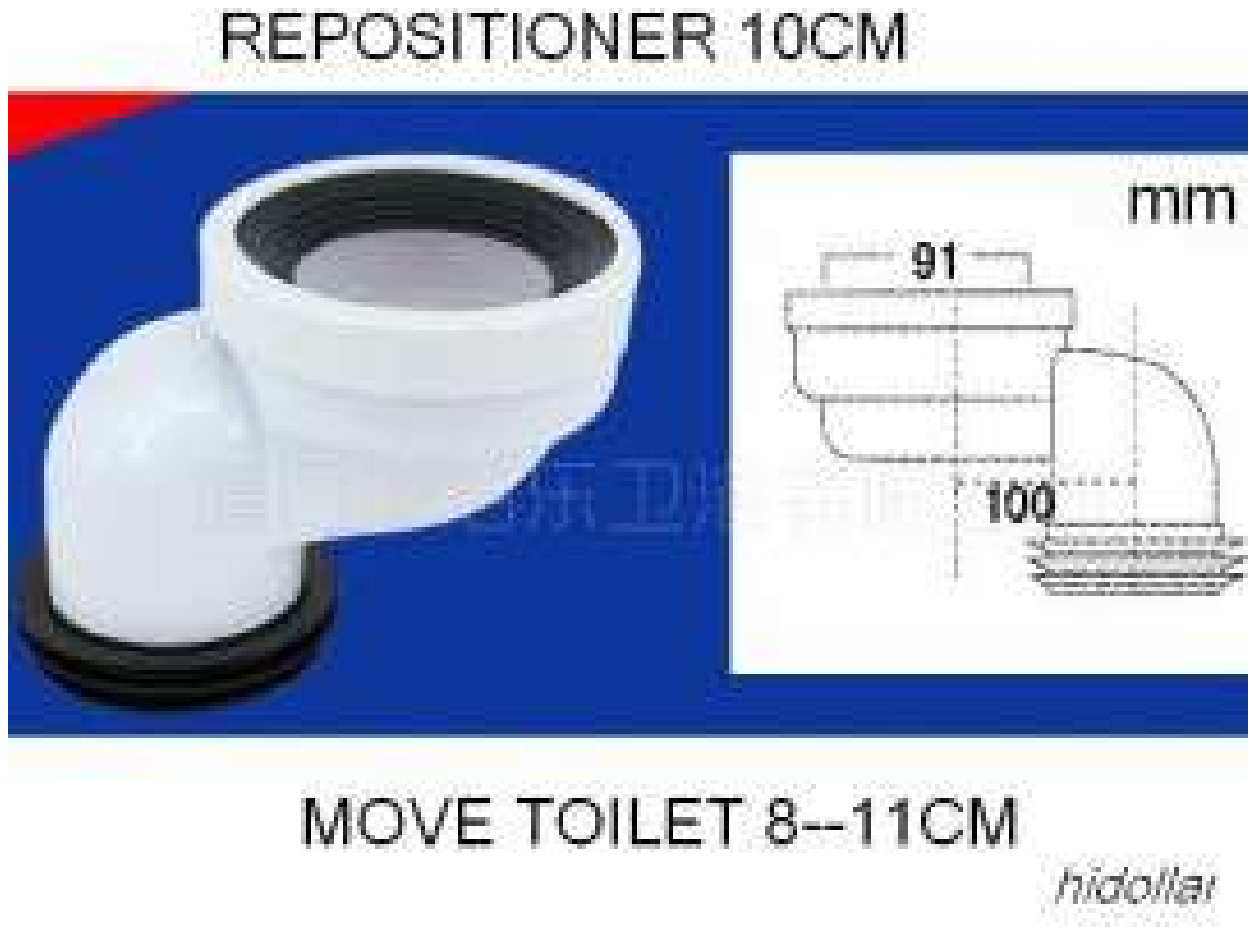} & \includegraphics[width=0.8\linewidth, height=3.5cm]{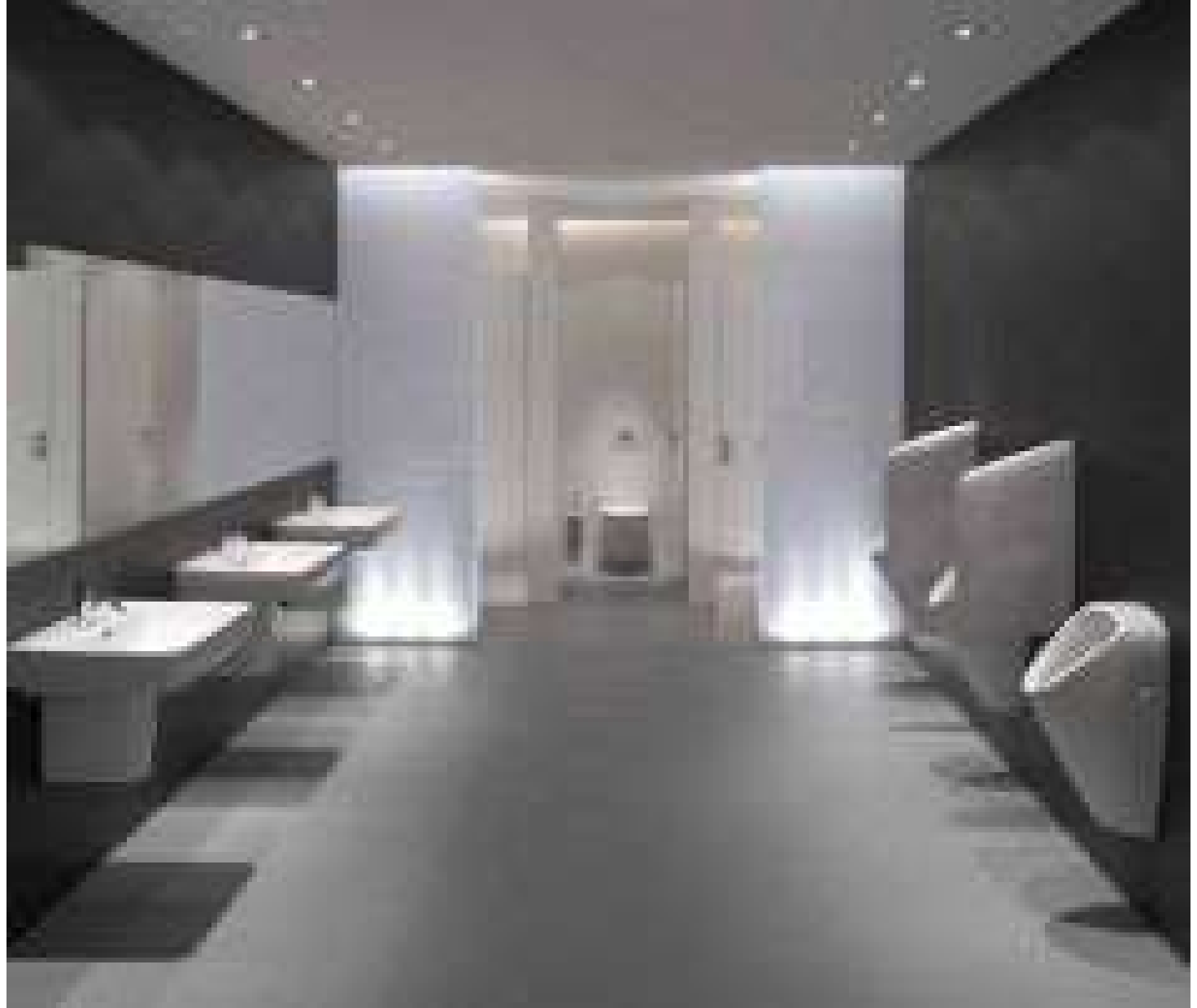} \\ \midrule
Bedroom & \includegraphics[width=0.8\linewidth, height=3.5cm]
{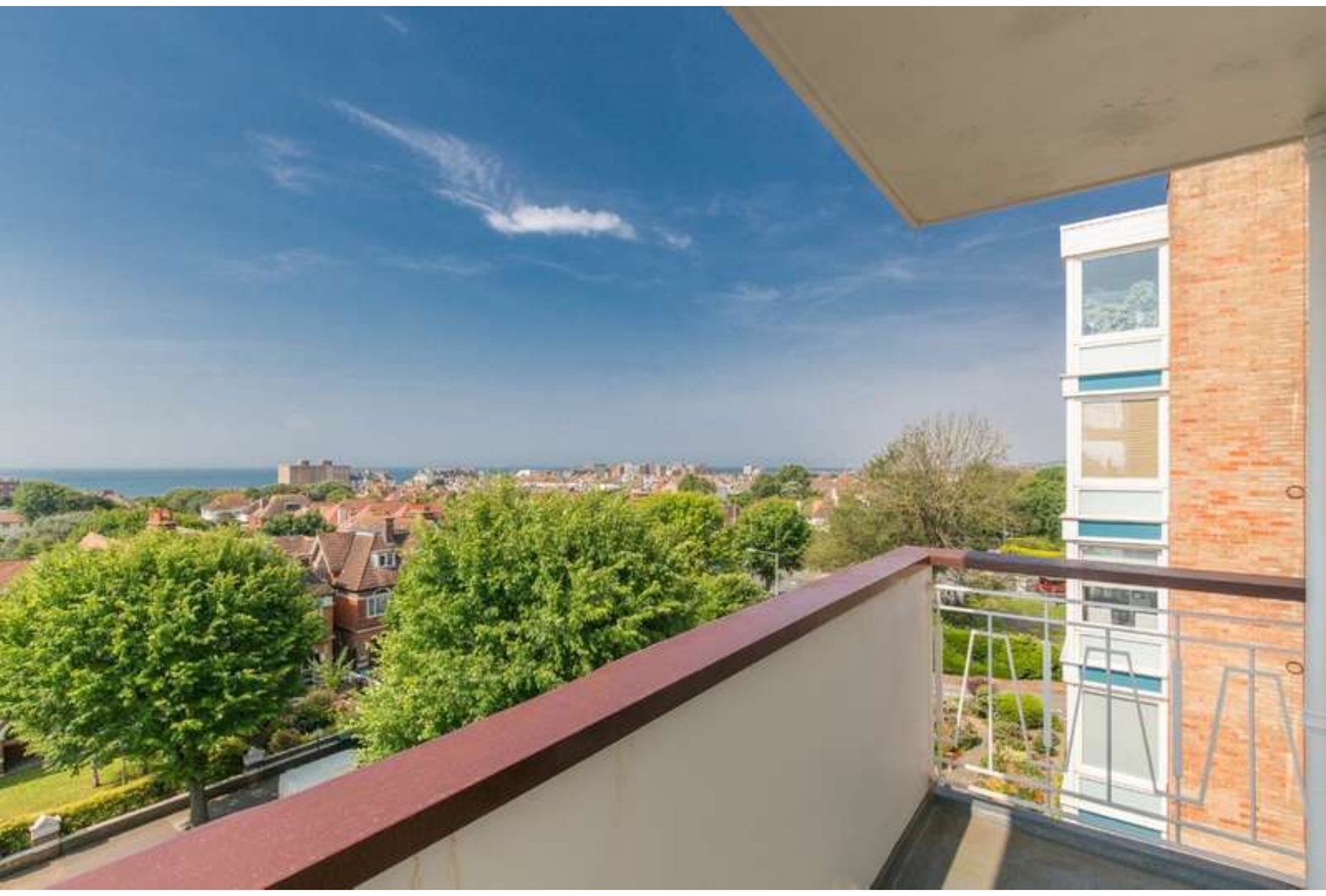} & \includegraphics[width=0.8\linewidth, height=3.5cm]{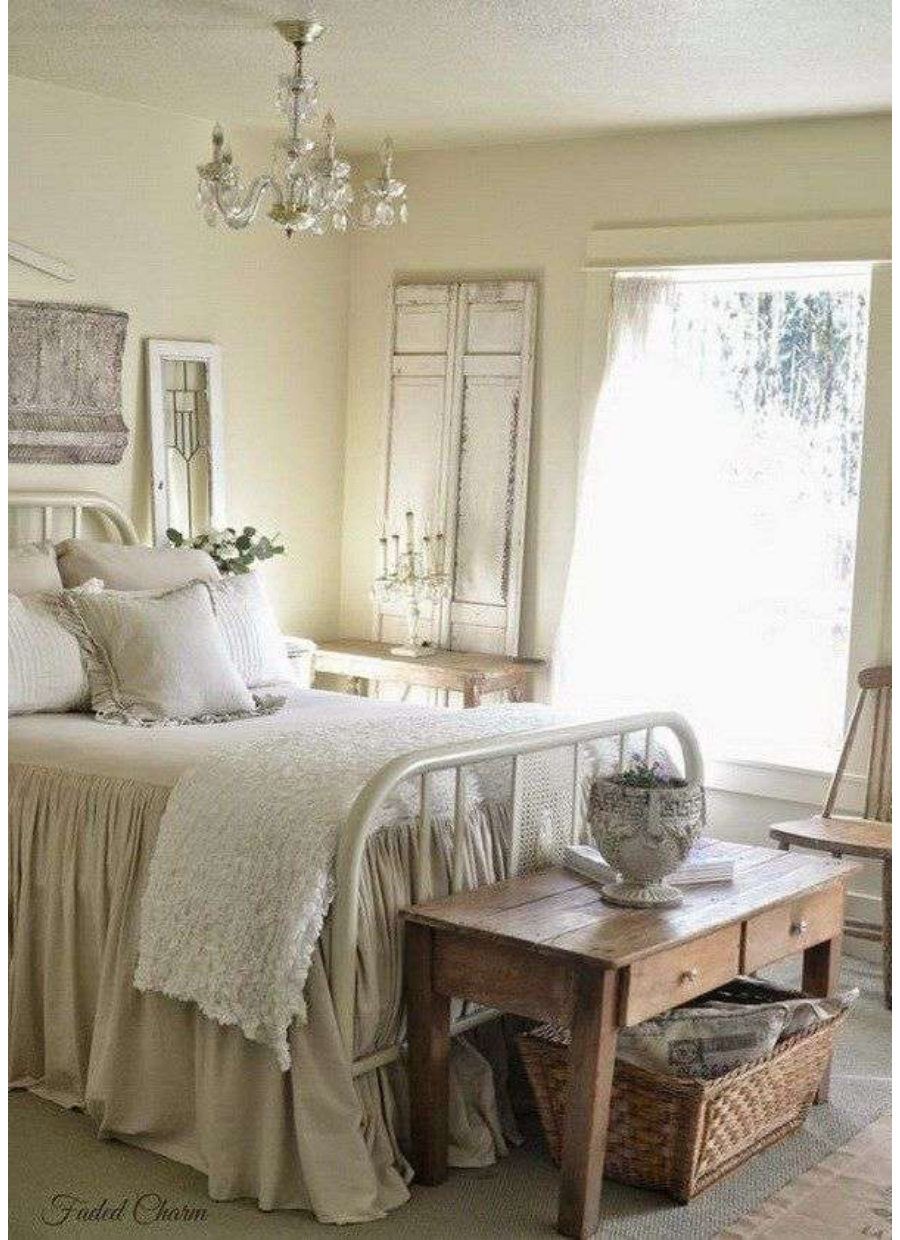} \\ \midrule
Apartment & \includegraphics[width=0.8\linewidth, height=3.5cm]
{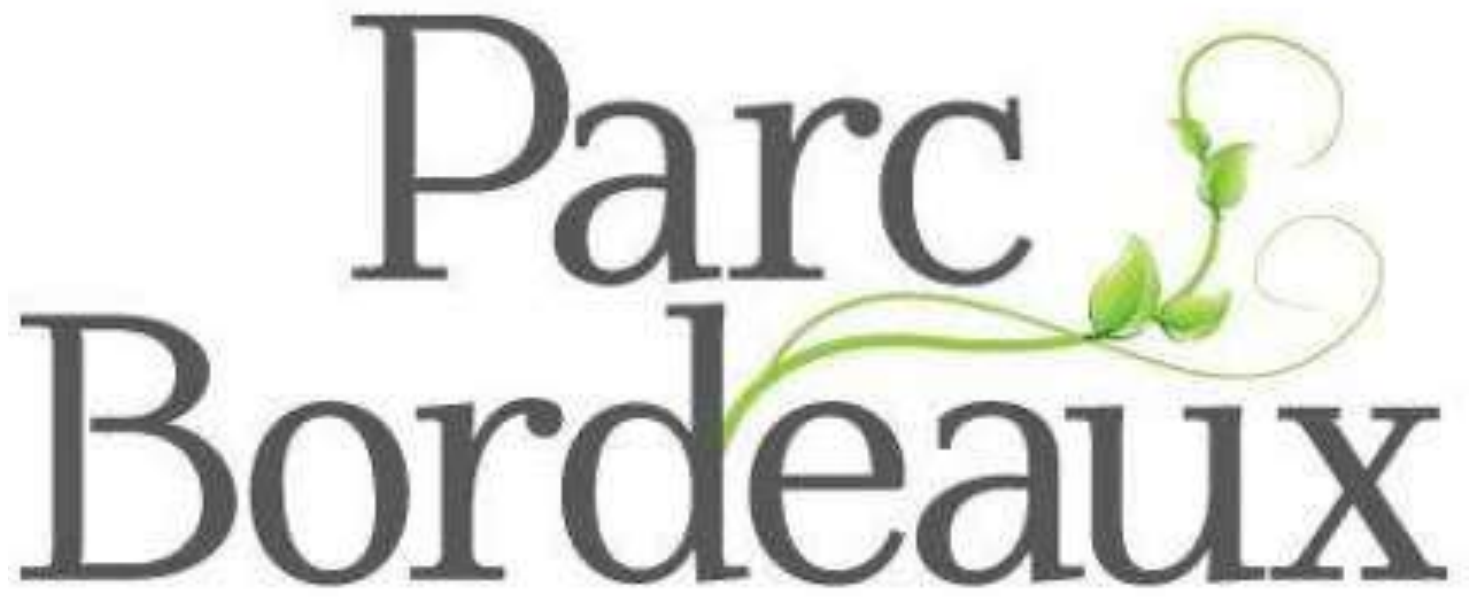} & \includegraphics[width=0.8\linewidth, height=3.5cm]{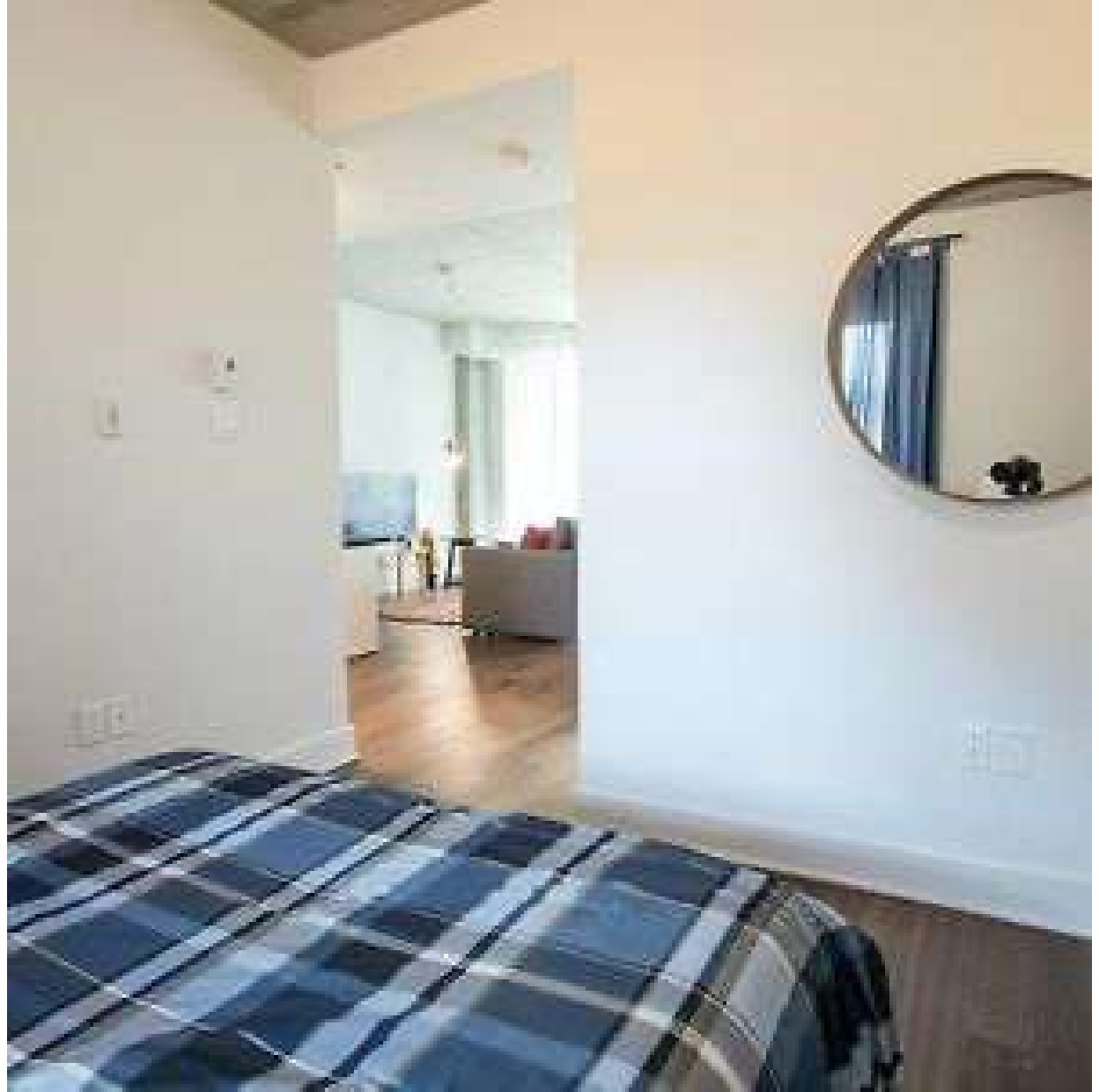} \\ \bottomrule
\end{longtable}
\subsection{Further Details on Geographically Profiling the Captions}
\label{subsec: geoprofile-2}

\textbf{Comparison of different geo-localizing methods}.
Subsection~\ref{subsec:geoprofile} in the main paper discusses the challenges faced in predicting countries for image-text pairs. We explore a number of alternatives for geographical profiling, which we describe below.
\mycolortext{
Using the three evaluation datasets described in subsection~\ref{subsec:geoprofile}, we evaluate several methods. We also analyse the effectiveness of other existing LLMs and VLMs for the task: Qwen2.5-7B Instruct~\footnote{\url{https://huggingface.co/Qwen/Qwen2.5-7B-Instruct}}, llama3.1-8B Instruct~\footnote{\url{https://huggingface.co/meta-llama/Llama-3.1-8B-Instruct}}, mistral-7B-Instruct-v0.3~\footnote{\url{https://huggingface.co/mistralai/Mistral-7B-Instruct-v0.3}}, gemma2-9b-it~\footnote{\url{https://huggingface.co/google/gemma-2-9b-it}}, gpt 5.1~\footnote{\url{https://openai.com/index/gpt-5-1/}}, Qwen2.5-VL-7B-Instruct~\footnote{\url{https://huggingface.co/Qwen/Qwen2.5-VL-7B-Instruct}}, and Llama-3.2-11B-Vision-Instruct~\footnote{https://huggingface.co/meta-llama/Llama-3.2-11B-Vision-Instruct}}. The LLM zero-shot prompts are shown in Figure~\ref{box:zero-shot}.

\textbf{Caption-only}:
\begin{itemize}
    \item \textbf{String Matching}: It searches for substrings in the captions that can be potential places as listed in the geodatabase we describe in subsection~\ref{subsec:geoprofile} in the main paper. While it is a fast algorithm, it can lead to a lot of false positives on account of ignoring the context.
    \item \textbf{NER taggers}: Instead of searching for substrings in the captions blindly, we use NER taggers to detect locations. Specifically, we choose the `GPE' and `LOC' tags returned by the spacy NER taggers~\citep{spacy} as place names. Spacy provides four models: small, medium, large and transformers. To take full advantage of all models, we pass the caption through each of them iteratively, until a place name is captured by one of them. Finally, the country is identified by searching for the country associated with the detected location with the help of the geodatabase. This method is more precise than string matching methods, as it can more accurately identify locations. But its recall is less, as very often NER taggers (including all their models) tend to miss place mentions in a given text.
    \item \textbf{Gemini-based extract-retrieve-estimate}: This is the final method we use to identify country names from captions. The method is described in subsection~\ref{subsec:geoprofile}. We use this method to first extract the location from a caption, if present, then retrieve the top 10 nearest matching locations from the GeoNames database, and augment them to the prompt along with their country names for the final output. We refer the reader to Figures~\ref{box:extract} and ~\ref{box:estimate} for the specific prompts. We also try an in-context-learning method, described in subsection~\ref{subsec:geoprofile}, which is rendered ineffective due to its high context length. The prompt for this method can be seen in Figure~\ref{box:icl}.
    \item \mycolortext{\textbf{Effectiveness of other LLMs}: While the open-source models are free of cost, their performance suffers compared to the closed-source ones on all datasets. Gemini and gpt, the proprietary models, are comparable in their zero-shot performance. We choose Gemini for all experiments in the rest of the work.} 
    \item \textcolor{black}{\textbf{Geoparsepy}: We additionally employ the Geoparsepy~\citep{middleton2018location} to extract country names from captions. As explained in Section~\ref{sec:related}, it is a geoparser that utilizes the OSM~\citep{osm} database to predict the geospatial characteristics of a given text. When applied on our annotated dataset, we find it to successfully extract location mentions in captions in multiple cases, but fail to map those locations to their countries. Hence, we query the Geonames~\citep{geonames} database on the location mentions for which the system fails to output a country name. Overall, we find that the method is inferior, compared to the LLM-based methods.}

    \item \textbf{Geograpy3}: Geograpy3~\footnote{\url{https://pypi.org/project/geograpy3/}} is a Python library that extracts place names from text. It internally uses NLTK~\footnote{\url{https://www.nltk.org//}}, Wikidata~\footnote{\url{https://www.wikidata.org/wiki/Wikidata:Main_Page}} and other resources to recognise entities and disambiguate regions based on population. Although it offers user-friendly APIs for easy access to such information, its evaluation on our test datasets shows inadequate results.
\end{itemize}

\begin{figure}[t!]
\begin{tcolorbox}[colback=cyan!5, colframe=cyan!40!black]

\texttt{You are a geotagging agent who tags each given text to a country, if a reference to a location is present in the text. The only output you give is either the coutry name or `NO' in case the text cannot be tagged to a country. Do NOT use abbreviations for country names. }

\texttt{Here are a few instructions.} 

\texttt{
\begin{itemize}
    \item Read the text carefully and understand the context.
    \item Find any location specified in the text, and then try to map it to a country name. If no location is specified, output `no'.
    \item Do NOT use abbreviations for any country name. 
    \item Do NOT just mention the inferred location name. Only mention the mapped country name, and nothing else.
    \item You cannot mention the US state name if present. You should predict United States for them.
    \item All countries and places under United Kingdom (e.g., england, london, northern ireland, scotland, wales, etc) should be marked as United Kingdom.
    \item All states, cities or places under United States should be marked as United States.
    \item Do NOT use terms like America, USA, United States of America, for United States.
    \item Do NOT make any assumptions about a caption for which the country name cannot be inferred directly from the caption.
    \item Do NOT make any assumptions about a caption for which the country name cannot be inferred directly from the caption.
    \item Do NOT make any assumptions about a caption for which no location is directly specified in the caption.
    \item Output should be no if only the continent name is specified, or other vague regions are mentioned in the caption like `mediterranean', `caribbean', etc.
    \item Ensure that the output only contains the country name, or no, if no country can be inferred from the caption.
\end{itemize}}

\end{tcolorbox}

\begin{center}
\refstepcounter{figure}
\label{box:zero-shot}
\textbf{Figure~\thefigure:} Prompt for \textbf{Zero-shot geolocalization of captions}.
\end{center}
\end{figure}

\begin{figure}
\begin{tcolorbox}[colback=cyan!5, colframe=cyan!40!black]

\texttt{You are a geoparsing agent who extracts the primary location mentioned in a text, if a reference to a location is present in the text. The only output you give is either the location name or `NO' in case the text cannot be tagged to a country. Do NOT use abbreviations for country names.}\\\\

\texttt{Here are a few instructions.}

\texttt{\begin{itemize}
\item Read the text carefully and understand the context.
\item Only output the location name, as present in the text. Do not convert it to a country name.
\item Only in case the country name is directly mentioned in the text, output the country name.
\item Find any location specified in the text. If no location is specified, output `no'.
\item Do NOT use abbreviations for any country name or location name. 
\item If the location is United Kingdom, the output should be United Kingdom, not any other abbreviation like UK, U.K., GB, Great Britain, etc.
\item If the location is United States, the output should be United States, not any other abbreviation like USA, America, etc.
\item Do NOT use terms like America, USA, United States of America, for United States.
\item Do NOT make any assumptions about a text for which the location name cannot be inferred directly from the text.
\item Do NOT make any assumptions about a text for which no location is directly specified in the text.
\item Output should be no if only the continent name is specified, or other vague regions are mentioned in the text like `mediterranean', `caribbean', etc.
\item Ensure that the output only contains the location name, or no, if no location can be inferred from the text.
\end{itemize}}

\end{tcolorbox}
\begin{center}
\refstepcounter{figure}
\label{box:extract}
\textbf{Figure~\thefigure:} \textbf{Prompt for extracting locations from captions}: the first step in the extract-retrieve-estimate method.
\end{center}
\end{figure}

\begin{figure}[t!]
\begin{tcolorbox}[colback=cyan!5, colframe=cyan!40!black]

\texttt{You are a geotagging agent who tags each given text to a country, if a reference to a location is present in the text. The only output you give is either the coutry name or `NO' in case the text cannot be tagged to a country. Do NOT use abbreviations for country names.}\\\\ 
                
\texttt{Here are a few examples of locations and their corresponding geotags. You can use this to identify the location and the country name. If the detected location is not present in the examples, you can use your own knowledge to identify the country name.\{examples\}}\\\\

\texttt{Here are a few instructions. }
\texttt{
\begin{itemize}
\item Read the text carefully and understand the context.
\item Find any location specified in the text, and then try to map it to a country name. If no location is specified, output `no'.
\item Do NOT use abbreviations for any country name. 
\item Do NOT just mention the inferred location name. Only mention the mapped country name, and nothing else.
\item You cannot mention the US state name if present. You should predict United States for them.
\item All countries and places under United Kingdom (e.g., england, london, northern ireland, scotland, wales, etc) should be marked as United Kingdom.
\item All states, cities or places under United States should be marked as United States.
\item Do NOT use terms like America, USA, United States of America, for United States.
\item Do NOT make any assumptions about a caption for which the country name cannot be inferred directly from the caption.
\item Do NOT make any assumptions about a caption for which no location is directly specified in the caption.
\item Output should be no if only the continent name is specified, or other vague regions are mentioned in the caption like `mediterranean', `caribbean', etc.
\item Ensure that the output only contains the country name, or no, if no country can be inferred from the caption.
\end{itemize}}

\end{tcolorbox}
\begin{center}
\refstepcounter{figure}
\label{box:estimate}
\textbf{Figure~\thefigure:} Prompt for \textbf{predicting locations from captions} using locations retrieved from the GeoNames database as examples: the last step in the extract-retrieve-estimate method.
\end{center}
\end{figure}

\begin{figure}[t!]
\begin{tcolorbox}[colback=cyan!5, colframe=cyan!40!black]

\texttt{You are a geotagging agent who tags each given text to a country, if a reference to a location is present in the text. The only output you give is either the coutry name or `NO' in case the text cannot be tagged to a country. Do NOT use abbreviations for country names. }\\\\

\texttt{Here are a few example texts with their corresponding geotags as examples.
\{examples\}}\\\\

\texttt{Here are a few instructions. }

\texttt{
\begin{itemize}
\item Read the text carefully and understand the context.
\item Find any location specified in the text, and then try to map it to a country name. If no location is specified, output `no'.
\item Do NOT use abbreviations for any country name. 
\item Do NOT just mention the inferred location name. Only mention the mapped country name, and nothing else.
\item You cannot mention the US state name if present. You should predict United States for them.
\item All countries and places under United Kingdom (e.g., england, london, northern ireland, scotland, wales, etc) should be marked as United Kingdom.
\item All states, cities or places under United States should be marked as United States.
\item Do NOT use terms like America, USA, United States of America, for United States.
\item Do NOT make any assumptions about a caption for which the country name cannot be inferred directly from the caption.
\item Do NOT make any assumptions about a caption for which no location is directly specified in the caption.
\item Output should be no if only the continent name is specified, or other vague regions are mentioned in the caption like `mediterranean', `caribbean', etc.
\item Ensure that the output only contains the country name, or no, if no country can be inferred from the caption.
\end{itemize}}

\end{tcolorbox}

\begin{center}
\refstepcounter{figure}
\label{box:icl}
\textbf{Figure~\thefigure:} Prompt for geolocalizing the captions using \textbf{in-context-learning}, via examples provided explicitly.
\end{center}
\end{figure}

\begin{table*}[t!]
\centering
\caption{\textbf{Performance of different image geolocalization methods}. Based on evaluations on the \textit{Location-Specified Subset} of $\mathcal{D}_{\text{self}}$, we observe that captions offer more reliable and grounded cues for geolocalizing than images. Hence, we opt to geolocalize the datasets with the help of the captions.}
\label{tab:geotagging-images}
\begin{tabular}{c|l|cc}
\hline
\multirow{2}{*}{\textbf{Focus}} & \multirow{2}{*}{\textbf{Method}}  & \multicolumn{2}{c}{\textbf{Location-Specified Subset}} \\ 
 &  & \textbf{Precision} & \textbf{Recall} \\ \hline
\multirow{5}{*}{Image} 
 & LLama-Vision & $0.56$ & $0.52$ \\
 & Qwen2.5 VL & $0.77$ & $0.61$ \\
 & GeoCLIP & $0.66$ & $0.64$ \\
 & Osv-5m & $0.44$ & $0.40$ \\
 & GeoReasoner & $0.64$ & $0.41$ \\
\midrule
\multirow{2}{*}{Image+Caption} 
 & LLama-Vision & $0.87$ & $0.73$ \\
 & Qwen2.5 VL & $0.97$ & $0.85$ \\
\bottomrule
\end{tabular}
\end{table*}

\textbf{Image-only}:
We explore the possibility of utilizing the image modality for geolocalizing image-caption pairs. Note that as the web-scraped vision-language datasets are rarely accompanied by image metadata, annotating locations in which the images are clicked can be prohibitively costly. Hence, we use the locations annotated in the captions of $\mathcal{D}_{\text{self}}$ as the ground truth for evaluating existing image-geolocalizers. Note that we filter $\mathcal{D}_{\text{self}}$ to only contain the location-specified captions to ensure accurate evaluation, which we denote as the `Location-Specified Subset'. The results are demonstrated in Table~\ref{tab:geotagging-images}. We discuss our observations below:
\begin{itemize}
    \item \textbf{Vision-Language Models (VLMs)}. We assess two VLMs in a zero-shot manner: Qwen 2.5 VL and Llama Vision. For both, we use the following prompt: ``\texttt{You are a geoparsing agent that maps the given image to a country name. You can look at the image, and output the country name if you find that there are definite indications, else output 'no'.  
                Here are a few instructions. 
                Do NOT use abbreviations for any country name. 
                Do NOT just mention the inferred location name. Only mention the mapped country name, and nothing else.
                All countries and places under United Kingdom (e.g., england, london, ireland, scotland, wales, etc) should be marked as United Kingdom.
                All states, cities or places under United States should be marked as United States.
                Do NOT use terms like America, USA, United States of America, for United States.
                Do NOT make any assumptions about a image for which the country name cannot be inferred directly.}''. While QwenVL achieves a precision of $0.77$ on the \textit{Location-Specified Subset}, its recall is low, highlighting the unreliability of the zero-shot open-source VLMs for geolocalizing in-the-wild images. The performance of Llama Vision is worse than that of QwenVL. 
    \item \textbf{Image Geo-Localizers}. We explore the performance of three popular Image Geo-Localizers in this paper: GeoCLIP~\cite{geoclip}, OSV-5M~\cite{osm} and GeoReasoner~\cite{li2024georeasoner}. Their performance in Table~\ref{tab:geotagging-images} shows that such methods are inadequate, as they specialize in geo-localizing street-view images. GeoCLIP is the best performing among these methods. \mycolortext{To explore the feasibility of geo-profiling the images using such geo-localizers, we create a small dataset for each entity, having images from a diverse set of countries like United States, United Kingdom, India, Australia, Nigeria, Brazil, Uganda, Thailand, and Argentina, as tagged by the Gemini model. For each country, we randomly sample upto $100$ images (we consider all images for a country with frequency $<100$). We evaluate GeoCLIP and OSV-5M using these datasets for every entity. While both these methods return GPS coordinates for any given image, we convert them to countries and continents with the help of the Nominatim API. The results are presented in Table~\ref{tab:geoclip}. We find that GeoCLIP performs better overall than OSV-5M, but its country accuracy is still low. Also, the continent accuracies of GeoCLIP suffer for indoor entities like toilet and kitchen. Overall, we feel that geoprofiling images is highly challenging, hence, we infer countries based on captions as mentioned in the main paper. Note that while recent papers~\citep{10.5555/3692070.3693246, wang2024llmgeo} demonstrate the effectiveness of using VLMs for the task of geo-localization, the best of these large vision-language models are closed-source, which would be infeasible to use given the scale of our study. Hence, we refrain from using them due to financial considerations.}
\end{itemize}
\begin{table*}[h!]
\centering
\small
\caption{\textbf{Geolocalizing Images Directly}. We evaluate the performance of two models and report the country/continent accuracies for each entity in this table.}
\label{tab:geoclip}
\begin{tabular}{@{}c|c|cc|cc@{}}
\toprule
\multirow{2}{*}{Entity} & \multirow{2}{*}{\begin{tabular}[c]{@{}c@{}}\# of test\\ images\end{tabular}} &  \multicolumn{2}{c|}{GeoCLIP} & \multicolumn{2}{c}{OSV-5M} \\ \cmidrule(l){3-6} 
 &  & Country & Continent & Country & Continent \\ \midrule
House & 1340 &  0.40 & 0.74 & 0.30 & 0.66 \\
Island & 659  & 0.57 & 0.76 & 0.32 & 0.50 \\
Flag & 3780 &  0.45 & 0.76 & 0.12 & 0.43 \\
Road & 1120 &  0.55 & 0.83 & 0.32 & 0.65 \\
Shop & 772  & 0.54 & 0.72 & 0.31 & 0.55 \\
Lake & 846  & 0.51 & 0.72 & 0.28 & 0.49 \\
Kitchen & 580 &  0.20 & 0.50 & 0.18 & 0.50 \\
Office & 738  & 0.41 & 0.62 & 0.30 & 0.49 \\
Car & 1140 &  0.32 & 0.66 & 0.16 & 0.51 \\
Beach & 1060 &  0.48 & 0.72 & 0.23 & 0.52 \\
Apartment & 1740  & 0.31 & 0.75 & 0.25 & 0.62 \\
Hill & 733  & 0.60 & 0.78 & 0.33 & 0.56 \\
Bedroom & 660  & 0.29 & 0.66 & 0.26 & 0.62 \\
Valley & 779  & 0.11 & 0.20 & 0.07 & 0.18 \\
Toilet & 500 & 0.19 & 0.54 & 0.12 & 0.49 \\
Hotel & 1800  & 0.36 & 0.74 & 0.23 & 0.61 \\
Living Room & 778  & 0.16 & 0.26 & 0.12 & 0.27 \\
Mountain & 820  & 0.51 & 0.75 & 0.29 & 0.48 \\
Wedding Dress & 750  & 0.30 & 0.43 & 0.15 & 0.33 \\
Jewelry & 794  & 0.10 & 0.20 & 0.04 & 0.19 \\
\bottomrule
\end{tabular}
\end{table*}

\textbf{Image+Caption}. We provide both the caption and the image from the `Location-specified Subset' of $\mathcal{D}_{\text{self}}$ to a VLM model to enquire about the possible country associated with a data point. We explore QwenVL and LLama Vision, and use the following prompt: 

``\texttt{You are a geoparsing agent that maps the given image-caption pair to a country name. You can look at the image, and output the country name if you find that there are certain indications, else first find any location specified in the given caption, and then map it to a country name.
                Given that the caption associated with the image is: {caption}, specify the country name associated with the image. Answer should only be country name, or no in case the country cannot be inferred. 
                Here are a few instructions. 
                Do NOT use abbreviations for any country name. 
                Do NOT just mention the inferred location name. Only mention the mapped country name, and nothing else.
                All countries and places under United Kingdom (e.g., england, london, ireland, scotland, wales, etc) should be marked as United Kingdom.
                All states, cities or places under United States should be marked as United States.
                Do NOT use terms like America, USA, United States of America, for United States.
                Do NOT make any assumptions about a image-caption pair for which the country name cannot be inferred directly from either the image or the caption.
                Do NOT make any assumptions about a image-caption pair for which no location is directly specified in the caption.
                Output should be no if only the continent name is specified, or other vague regions are mentioned in the caption like `mediterranean', `caribbean', etc.}''. We find a significant performance boost when the caption is provided to the VLM (compared to the image-only case), highlighting the reliance of such models on the captions to predict country name. However, as the LLMs perform similarly, we choose the LLama LLM and geo-profile only captions as they are time and resource efficient.
The performance of each of these methods is shown in Table~\ref{tab:geotagging}.

\begin{table*}[t]
\centering
\caption{
\textbf{Underspecified and location-specific captions} for each entity
}
\label{tab:captions-qualitative}
\begin{tabular}{l|p{5.5cm}|p{5.5cm}}
\toprule
& Specified & Unspecified \\ \midrule
\multicolumn{1}{c|}{House} & "Thumbnail 4 bed detached house for sale in Southfields, Rochester" & "Exterior house colors with brown roof 04" \\ \midrule
\multicolumn{1}{c|}{Flag} & "Wooden Framed HOME American Flag" & "Medieval knight on horse carrying a flag - Vector..." \\ \midrule
\multicolumn{1}{c|}{Car} & "Car on Rent in Vadodara with Driver" & "Under a Car Stock Photography" \\ \midrule
\multicolumn{1}{c|}{Kitchen} & "Kitchen Countertop Suppliers Calgary" & "Kitchen and dining space" \\ \midrule
\multicolumn{1}{c|}{Road} & "How Calabar-Odukpani Road Dualization Caused Five Accidents Within A Week" & "nature, road, and forest image" \\ \midrule
\multicolumn{1}{c|}{Beach} & "Happy boys at Copacabana Beach" & "Abstract background of sand at the beach" \\ \midrule
\multicolumn{1}{c|}{Hotel} & "Jade Court Motor Lodge, hotel in Hokitika" & "Reception of a hotel with a bell, 3d illustration" \\ \midrule
\multicolumn{1}{c|}{Toilet} & "Rules for using the toilet in Sochi" & "A toilet used as exhibition space" \\ \midrule
\multicolumn{1}{c|}{Bedroom} & "Four Bedroom House In Ntinda For Rent" & "Master bedroom with King Bed" \\ \midrule
\multicolumn{1}{c|}{Apartment} & "Rental apartment Toulouse 758€ CC - Picture 7" & "\$3285 Two bedroom Apartment for rent" \\ \bottomrule
\end{tabular}
\end{table*}

\textbf{Examples of the different Captions}. We show examples of captions with and without location mentions in Table~\ref{tab:captions-qualitative}. It shows how some captions are underspecified, and while others may contain place mentions, it may not be trivial to extract them and find the associated country names, as evident in Table~\ref{tab:geotagging}.

\subsection{Distribution-based Analysis -- Further Details}
\label{subsec: datacomp-en}

\subsubsection{Percentage of Underspecified Captions}
\label{subsubsec:underspec}
We present the percentages of the underspecified captions per entity across Re-LAION2B-en, DataComp1B, CC12M and the multilingual subsets in Table~\ref{tab:underspecifications}. $55.1\%$ of Re-LAION2B-en, $57.6\%$ of DataComp1B and $58.0\%$ of CC12M captions are found to be location-underspecified. Of the multilingual captions, $60.3\%$ are estimated to be underspecified, averaged across all studied languages.

\begin{table}[]
\caption{\textbf{Percentage of location-underspecified captions per entity}. Across datasets, we find $55.1\%$ of Re-LAION2B-en, $57.6\%$ of DataComp1B and $58.0\%$ of CC12M captions are found to be location-underspecified. Of the multilingual captions, $60.3\%$ are predicted to be underspecified, averaged across all studied languages.}
\label{tab:underspecifications}
\small
\begin{tabular}{@{}l|lllllll@{}}
\toprule
Entity        & Re-LAION2B-en & DataComp1B & CC12M  & Japanese & Spanish & Greek & Hindi \\ \midrule
house         & $37.1$        & $49.3$     & $61.3$ & $75.4$           & $40.5$           & $68.9$           & $68.3$           \\
flag          & $19.8$        & $24.5$     & $16.1$ & $41.6$           & $27.3$           & $18.9$           & -                \\
hotel         & $16.8$        & $30.8$     & $27.6$ & $40.6$           & $52.8$           & $48.8$           & $63.6$           \\
car           & $66.8$        & $73.5$     & $74.4$ & $61.0$           & $84.8$           & $83.2$           & -                \\
toilet        & $87.9$        & $88.2$     & $83.9$ & $88.8$           & $89.9$           & -                & -                \\
bedroom       & $80.3$        & $73.6$     & $80.6$ & $80.1$           & $87.1$           & $98.5$           & $39.9$           \\
beach         & $41.8$        & $38.2$     & $47.9$ & $52.6$           & $53.2$           & $59.0$           & -                \\
apartment     & $27.9$        & $32.8$     & $55.0$ & $23.2$           & $32.5$           & $52.6$           & $44.1$           \\
kitchen       & $83.0$        & $88.3$     & $82.2$ & $90.9$           & $86.6$           & $97.4$           & -                \\
road          & $41.3$        & $47.5$     & $50.8$ & $52.0$           & $56.3$           & $65.2$           & $45.2$           \\
lake          & $25.4$        & $27.9$     & $37.1$ & $48.1$           & $46.6$           & $53.5$           & -                \\
living room   & $84.8$        & $87.5$     & $83.6$ & $89.4$           & $93.2$           & $95.9$           & -                \\
mountain      & $38.8$        & $56.7$     & $48.5$ & $44.7$           & $60.1$           & $53.8$           & -                \\
shop          & $48.6$        & $77.5$     & $51.0$ & $50.3$           & $54.7$           & -                & -                \\
valley        & $18.2$        & $23.9$     & $21.5$ & $40.2$           & $32.5$           & $28.9$           & -                \\
island        & $12.8$        & $36.5$     & $15.7$ & $27.2$           & $26.5$           & $27.5$           & -                \\
wedding dress & $85.9$        & $84.2$     & $79.0$ & $91.6$           & $87.3$           & -                & -                \\
office        & $79.1$        & $78.4$     & $80.1$ & $62.6$           & $56.1$           & $94.7$           & -                \\
jewelry       & $85.4$        & $87.9$     & $75.7$ & $74.8$           & $87.2$           & $88.8$           & -                \\
hill          & $27.1$        & $39.1$     & $41.1$ & $57.4$           & $56.4$           & -                & -                \\ \bottomrule
\end{tabular}
\end{table}

\subsubsection{Percentage of Occurrence of Top 15 Countries Across Entities}
\label{subsubsec:top15}
\begin{figure*}[t!]
    \centering
    \includegraphics[trim=0cm 8cm 0cm 2cm, clip, width=\linewidth]{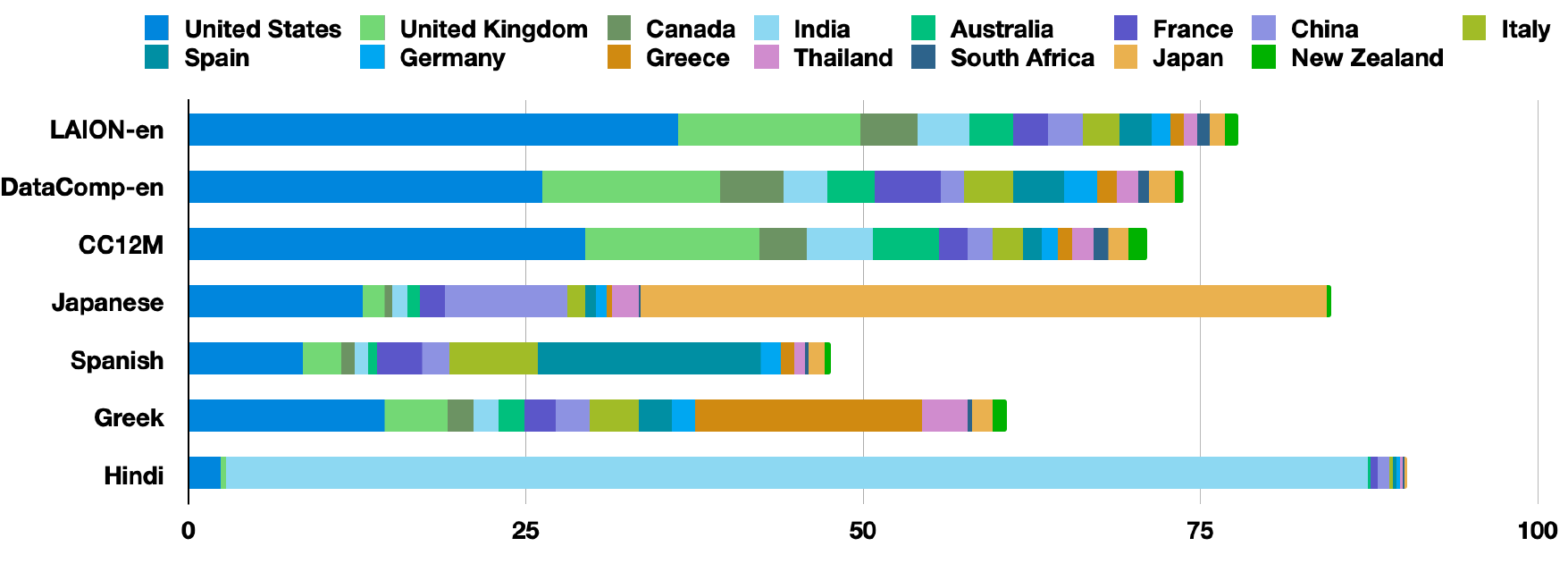} %
    \caption{\textbf{Top 15 countries across English and multilingual caption datasets}. While $12$ of the top $15$ most frequent countries are common across Re-LAION2B-en, DataComp1B and CC12M, the most frequently occurring countries in the multilingual captions are those where the respective languages are predominently spoken.
    }
    \label{fig:top15appn}
\end{figure*}

Our findings suggest that in Re-LAION2B-en, $77.2\%$ of all location-specified captions belong to only $15$ countries. In DataComp1B and CC12M, the percentages are $74.0$ and $70.8$ respectively. Similar patterns are seen in case of the multilingual captions, where $82.5\%$ of the Japanese captions, $73.0\%$ of the Spanish captions, $63.5\%$ of the Greek captions and $94.5$ of the Hindi captions belong to the top 15 countries. These findings indicate the high skewness in the country distributions in vision-language datasets, irrespective of languages, datasets and entities. Entity-wise details are shown in Table~\ref{tab:top15}. The names and percentages of the top $15$ countries per dataset are shown in Fig.~\ref{fig:top15appn}.

\begin{table}[]
\caption{\textbf{Percentage of Occurrence of Top 15 Countries per entity}. As seen in the table, for most entities, $>70\%$ of the corresponding captions belong to only $15$ countries of the world, across datasets. Averaged across entities, $77.2\%$ of the location-specified captions of Re-LAION2B-en belong to $15$ countries, whereas in DataComp1B and CC12M, the percentages are $74.0$ and $70.8$ respectively. $82.3\%$ of the Japanese captions, $72.5\%$ of the Spanish captions, $63.0\%$ of the Greek captions and $94.5\%$ of the Hindi captions belong to the corresponding top 15 countries, indicating high skewness in country representations, irrespective of language and entities.}
\label{tab:top15}
\small
\begin{tabular}{@{}l|lllllll@{}}
\toprule
Entity        & Re-LAION2B-en & DataComp1B & CC12M  & Japanese & Spanish & Greek & Hindi \\ \midrule
house         & $89.8$        & $84.3$     & $80.0$ & $92.6$           & $92.9$           & $71.9$           & $97.3$           \\
flag          & $65.8$        & $63.9$     & $51.3$ & $65.5$           & $64.1$           & $58.2$           & -                \\
hotel         & $72.6$        & $64.7$     & $76.0$ & $87.5$           & $65.7$           & $77.2$           & $100.0$          \\
car           & $86.8$        & $84.5$     & $84.1$ & $97.6$           & $83.3$           & $80.4$           & -                \\
toilet        & $88.4$        & $86.0$     & $83.7$ & $93.1$           & $89.5$           & -                & -                \\
bedroom       & $85.9$        & $82.5$     & $83.2$ & $83.6$           & $91.0$           & $100.0$          & $99.4$           \\
beach         & $76.8$        & $74.9$     & $70.1$ & $85.1$           & $79.1$           & $73.4$           & -                \\
apartment     & $85.7$        & $74.9$     & $79.8$ & $98.1$           & $97.9$           & $94.8$           & $98.6$           \\
kitchen       & $92.8$        & $90.6$     & $89.4$ & $94.1$           & $93.0$           & $90.9$           & -                \\
road          & $77.6$        & $81.7$     & $70.8$ & $88.3$           & $75.8$           & $67.7$           & $97.8$           \\
lake          & $86.1$        & $90.4$     & $84.2$ & $84.7$           & $74.3$           & $68.4$           & -                \\
living room   & $89.4$        & $85.4$     & $85.2$ & $95.4$           & $81.9$           & $91.7$           & -                \\
mountain      & $76.0$        & $82.7$     & $74.7$ & $89.3$           & $67.1$           & $67.7$           & -                \\
shop          & $82.2$        & $80.3$     & $76.5$ & $94.9$           & $82.4$           & -                & -                \\
valley        & $89.9$        & $91.4$     & $79.9$ & $85.5$           & $82.4$           & $68.4$           & -                \\
island        & $74.7$        & $79.1$     & $63.2$ & $88.3$           & $66.1$           & $80.1$           & -                \\
wedding dress & $86.6$        & $83.3$     & $85.3$ & $94.0$           & $88.0$           & -                & -                \\
office        & $85.5$        & $82.5$     & $87.3$ & $97.3$           & $93.9$           & $85.9$           & -                \\
jewelry       & $80.9$        & $76.8$     & $77.4$ & $97.0$           & $81.4$           & $95.3$           & -                \\
hill          & $80.5$        & $92.6$     & $77.8$ & $80.0$           & $59.6$           & -                & -                \\ \bottomrule
\end{tabular}
\end{table}

\subsubsection{Frequency vs Socio-Economic Factors for DataComp1B}
\label{subsec: datacomp-socio}
\begin{figure}[t!]
    \centering
    \includegraphics[trim=1.5cm 0.7cm 1.6cm 1.6cm, clip, width = 0.5\columnwidth]{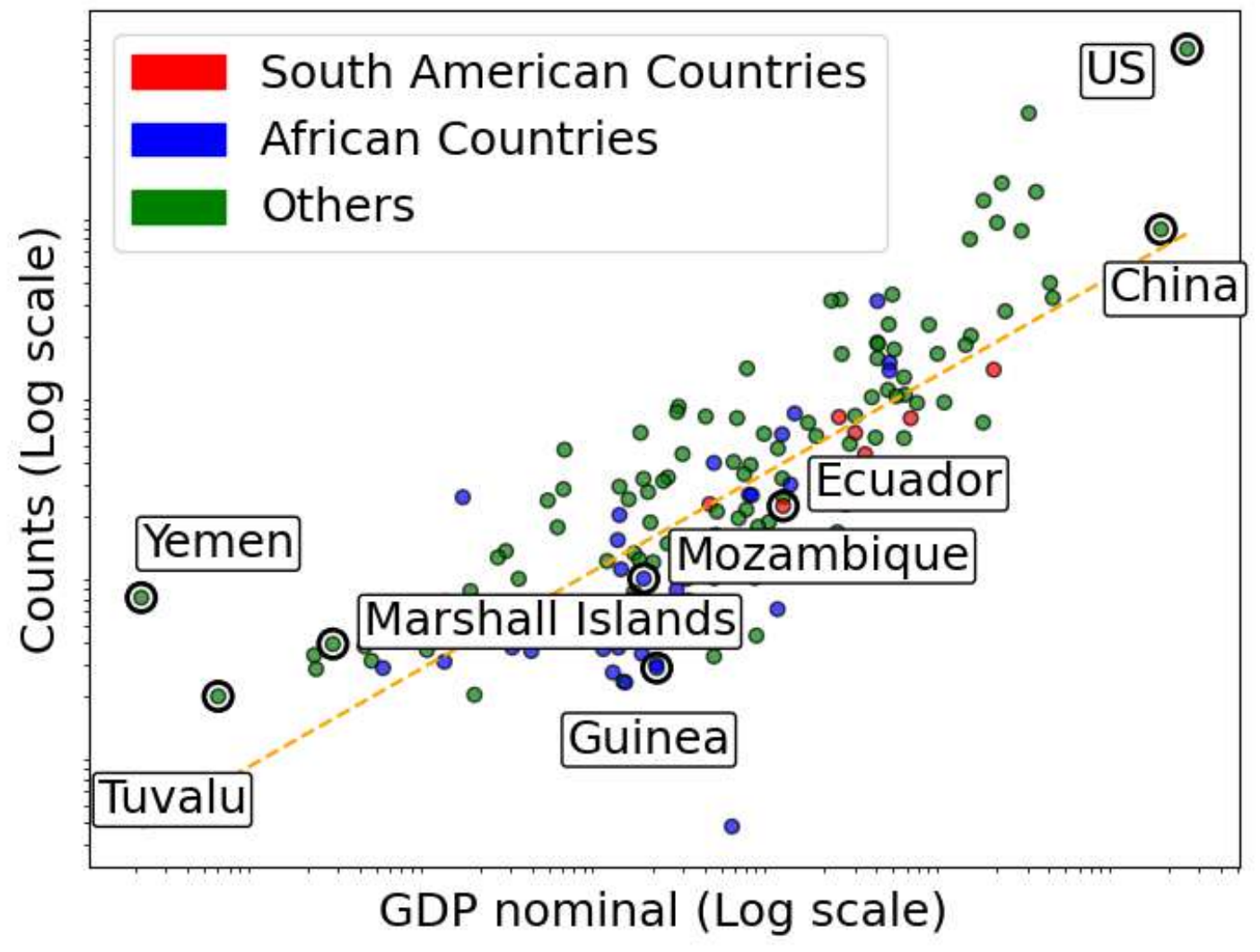}
    \caption{\textbf{Correlations of counts across all entities and GDP (nominal) for Re-LAION2B-en}. We find a high correlation~$(\rho = 0.84)$, which shows that wealthier countries are more represented across entities. 
    }
    \label{fig:correlation-gdp}
\end{figure}

Combined all across entities, we find a strong positive correlation between frequency of countries and nominal GDP ($\rho=0.84$) for Re-LAION2B-en, as shown in Figure~\ref{fig:correlation-gdp}. For DataComp1B-en, a similar pattern is seen  ($\rho=0.81$). A weak positive correlation can be seen between frequency and both population, GDP per capita and number of internet users ($\rho=0.32$, $0.27$ and $0.35$ respectively), whereas a moderate positive correlation is observed in case of total area ($0.43$) of the countries. These observations are highly similar to those of Re-LAION2B-en.

\subsubsection{Continent-wise Analysis of Geographical Distribution}
\label{subsec: continent}

\begin{table}[]
\caption{\textbf{Percentage of occurrence for the continents (averaged across entities)} for the $5$ studied languages in Re-LAION along with DataComp1B and CC12M. For each language, the most represented continents are those where it is predominantly spoken. NA: North America, SA: South America, EU: Europe, AS: Asia, AF: Africa, OC: Oceania.}
\small
\label{tab:continents_all}
\begin{tabular}{@{}l|ccccccc@{}}
\toprule
Continents & \begin{tabular}[c]{@{}l@{}}Re-\\ LAION2B\\ -en\end{tabular} & \begin{tabular}[c]{@{}l@{}}Data\\ Comp\\ 1B\end{tabular} & CC12M & \begin{tabular}[c]{@{}l@{}}Re-\\ LAION2B\\ (Jap)\end{tabular} & \begin{tabular}[c]{@{}l@{}}Re-\\ LAION2B\\ (Spa)\end{tabular} & \begin{tabular}[c]{@{}l@{}}Re-\\ LAION2B\\ (Gre)\end{tabular} & \begin{tabular}[c]{@{}l@{}}Re-\\ LAION2B\\ (Hin)\end{tabular} \\ \midrule
NA & $\mathbf{42.8}$ & $33.6$ & $\mathbf{36.9}$ & $14.7$ & $18.0$ & $20.2$ & $2.8$ \\
SA & $1.6$ & $1.9$ & $1.8$ & $1.4$ & $26.4$ & $3.6$ & $0.2$ \\
EU & $30.8$ & $\mathbf{39.2}$ & $28.6$ & $10.3$ & $\mathbf{40.9}$ & $\mathbf{47.4}$ & $3.0$ \\
AS & $16.7$ & $17.4$ & $22.0$ & $\mathbf{70.4}$ & $11.0$ & $21.6$ & $\mathbf{92.6}$ \\
AF & $3.7$ & $3.7$ & $3.9$ & $1.7$ & $2.3$ & $3.8$ & $1.0$ \\
OC & $4.4$ & $4.3$ & $6.8$ & $1.5$ & $1.3$ & $3.4$ & $0.3$ \\ \bottomrule
\end{tabular}
\end{table}

\begin{table*}[]
\centering
\small
\caption{\textbf{Distribution of continents in captions across entities in the Re-LAION2B-en dataset}. We tabulate the representation of the continents across the globe for all $20$ entities. As expected, we observe that Europe (EU) and North America (NA) are the most dominant continents, followed by Asia (AS), while Africa (AF), Oceania (OC) and South America (SA) have comparatively lower representations.}
\label{tab:continent}
\begin{tabular}{@{}c|cccccc@{}}
\toprule
 & \multicolumn{1}{c}{NA} & \multicolumn{1}{c}{SA} & \multicolumn{1}{c}{EU} & \multicolumn{1}{c}{AS} & \multicolumn{1}{c}{Af} & \multicolumn{1}{c}{OC} \\ \midrule
House & $22.0$ & $0.5$ & $\mathbf{60.8}$ & $9.9$ & $2.8$ & $3.9$ \\
Flag & $\mathbf{47.5}$ & $4.1$ & $25.5$ & $13.8$ & $6.4$ & $2.7$ \\
Car & $\mathbf{47.1}$ & $0.5$ & $30.7$ & $15.8$ & $2.6$ & $3.4$ \\
Kitchen & $\mathbf{57.2}$ & $0.2$ & $25.8$ & $9.9$ & $1.7$ & $5.3$ \\
Beach & $\mathbf{41.9}$ & $2.1$ & $26.9$ & $15.0$ & $4.7$ & $9.4$ \\
Road & $27.4$ & $1.5$ & $\mathbf{42.0}$ & $17.1$ & $4.3$ & $7.7$ \\
Hotel & $32.1$ & $1.3$ & $\mathbf{35.0}$ & $23.8$ & $3.4$ & $4.4$ \\
Bedroom & $\mathbf{40.0}$ & $0.4$ & $36.1$ & $16.6$ & $3.5$ & $3.4$ \\
Toilet & $\mathbf{35.5}$ & $0.4$ & $30.5$ & $26.2$ & $1.8$ & $5.5$ \\
Apartment & $\mathbf{47.4}$ & $0.5$ & $27.5$ & $18.6$ & $3.5$ & $2.6$ \\ 
Lake & $\mathbf{59.1}$ & $1.7$ & $22.7$ & $10.3$ & $1.3$ & $4.9$ \\
Hill & $32.5$ & $1.8$ & $\mathbf{42.4}$ & $14.1$ & $2.0$ & $7.2$ \\
Shop & $\mathbf{32.7}$ & $0.8$ & $32.2$ & $18.8$ & $2.7$ & $12.8$ \\
Valley & $\mathbf{68.8}$ & $1.9$ & $13.8$ & $7.5$ & $2.6$ & $5.4$ \\
Jewelry & $32.1$ & $1.1$ & $19.9$ & $\mathbf{39.6}$ & $5.0$ & $2.3$ \\
Wedding Dress & $28.3$ & $0.3$ & $\mathbf{30.7}$ & $30.4$ & $4.2$ & $6.1$ \\
Office & $\mathbf{44.6}$ & $0.3$ & $25.0$ & $21.8$ & $2.0$ & $6.2$ \\
Island & $22.8$ & $1.6$ & $\mathbf{33.6}$ & $22.5$ & $4.0$ & $15.5$ \\
Living Room & $\mathbf{48.9}$ & $0.4$ & $25.9$ & $19.5$ & $2.4$ & $2.9$ \\
Mountain & $\mathbf{46.3}$ & $3.5$ & $27.2$ & $14.5$ & $5.7$ & $2.9$ \\
\bottomrule
\end{tabular}
\end{table*}

In this section, we analyze the continental distribution of images across the studied datasets. Datasets with English captions disproportionately represent North America and Europe, whereas multilingual datasets predominantly reflect the continents in which the corresponding languages are primarily spoken. Notably, South America, severely underrepresented in English-captioned datasets (with less than $2\%$ representation)—emerges as the second most represented continent in Spanish captions, accounting for $26.4\%$ of the data. Dataset-wise continental representation statistics are summarized in Table~\ref{tab:continents_all}.

We further examine the continental distribution at the entity level within Re-LAION2B-en. For $13$ out of the 
$20$ entities, North America (NA) is the most frequently represented continent, followed by Europe (EU). In contrast, Europe dominates the distribution for $6$
entities, while Asia is the most represented continent for the entity jewelry. Overall, Asia (AS) ranks as the third most represented continent, followed by Oceania (OC), Africa (AF), and South America (SA). Detailed continent-wise frequencies for each entity in Re-LAION2B-en are reported in Table~\ref{tab:continent}.

\subsection{Qualitative Examples}
\label{subsec: q}
\begin{figure*}[t] %
    \centering
    \includegraphics[width=\linewidth]{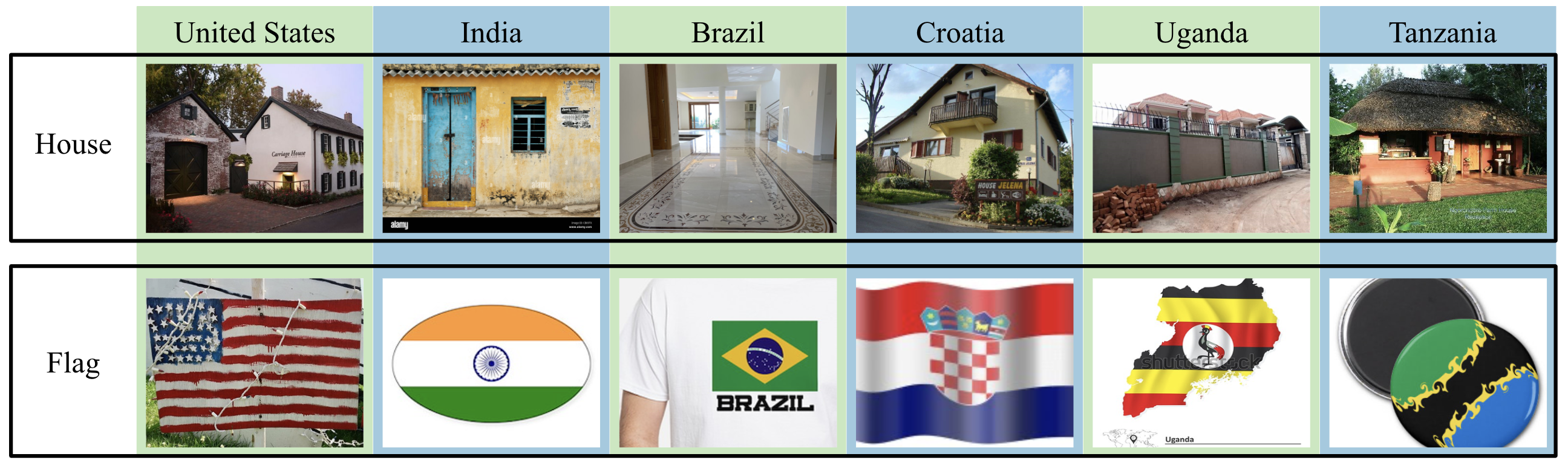} %
    \caption{\textbf{Images of Houses and Flags} belonging to different countries as predicted by \sysstop For this visual analysis, we pick two high-frequency countries (United States and India), two mid-frequency countries (Brazil, Croatia), and two low-frequency countries (Uganda, Tanzania).}
    \label{fig:qual}
\end{figure*}

We visualize the image-caption pairs of Re-LAION2B-en after being assigned by the \systemname to their corresponding countries in Figure~\ref{fig:qual}. Specifically, we show images from two high-frequency countries (United States and India), two mid-frequency countries (Brazil and Croatia), and two low-frequency countries (Uganda and Tanzania).

\subsection{Further Details on Comparisons with Real-world Distributions}
\label{subsec: gr2}
\textbf{Ground Truth Distribution For Each Entity}.
In the main paper, we motivate the need for comparing the geographical distributions obtained from the dataset with a ground truth reference distribution (subsection~\ref{subsec: gr}). Given an entity, we obtain the ground truth distribution from real-world data. For house, we use the available data on number of households in the world~\citep{house}. For car, we use country-wise data on motor vehicles per capita\citep{car}, similarly for road, we use details on road network size~\citep{road}. For hotel we use the information provided by UN World Tourism~\citep{hotel}. For mountain, we use the average elevation of a country as the reference distribution~\citep{mountain}. For jewelry, we use the data on annual country-wise jewelry exports~\cite{jewelry}. We approximate the number of beaches in a country through its length of coastline available~\citep{beach}. For island, we use available data on number of islands per country~\cite{island}. Note that we only select those entities for which we could find reliable ground truth distributions.

\noindent \textbf{Misalignment of Entities in DataComp1B}. We show the geographical misalignment scores with respect to the studied $8$ entities for DataComp1B in Table~\ref{tab:representation-datacomp}, and find that for mountain, more than $50\%$ of countries underrepresented, whereas $71\%$ countries are overrepresented in case of jewelry. On average, when combined across entities, $31.6\%$ countries are underrepresented, while $35.1\%$ are overrepresented.

\begin{table*}
\centering
\small
\caption{\textbf{Comparison of the geographical distributions for specific entities in the DataComp1B dataset}. We show the percentage of countries that are underrepresented (Under) and overrepresentated (Over) for different entities. On average, $31.6\%$ countries are underrepresented, whereas $35.1\%$ are overrepresented across entities.}
\label{tab:representation-datacomp}
\begin{tabular}{c|c|c|c|c|c|c|c|c|c}
\hline
 & House & Car & Road & \begin{tabular}[c]{@{}c@{}}Jewe-\\ lry\end{tabular} & Beach & Island & Hotel & \begin{tabular}[c]{@{}c@{}}Moun-\\ tain\end{tabular} & Avg \\ \hline
Under & $38.8$ & $39.9$ & $30.9$ & $13.6$ & $33.1$ & $5.5$ & $31.7$ & $\mathbf{59.3}$  & $31.6$ \\
Over & $19.9$ & $26.0$ & $32.6$ & $\mathbf{71.4}$ & $40.8$ & $48.4$ & $28.4$ & $13.1$  & $35.1$ \\ \hline
\end{tabular}
\end{table*}

\begin{table}[]
\caption{\textbf{Frequency and diversity rank correlations for DataComp1B.} For each entity, we compute the correlation between the frequency of countries and diversity scores of the images ($\rho_{fi}$). We notice that the majority of entities exhibit strong correlation (average Spearman's $\rho=0.67$).}
\label{tab:div_corr_datacomp}
\small
\begin{tabular}{c|cccccccccc}
\toprule
 & House & Flag & Car & Kitchen & Beach & Road & Hotel & \begin{tabular}[c]{@{}c@{}}Bed-\\ room\end{tabular} & Toilet & \begin{tabular}[c]{@{}c@{}}Apart-\\ ment\end{tabular} \\ \midrule
$\rho_{\text{fi}}$ & $0.37$ & $0.63$ & $0.80$ & $0.22$ & $0.68$ & $0.67$ & $0.76$ & $0.93$ & $-0.29$ & $0.91$ \\ \midrule
 & \begin{tabular}[c]{@{}c@{}}Moun-\\ tain\end{tabular} & Island & Shop & Valley & \begin{tabular}[c]{@{}c@{}}Living\\ Room\end{tabular} & \begin{tabular}[c]{@{}c@{}}Wedding\\ Dress\end{tabular} & Hill & Lake & Office & Jewelry \\ \hline
$\rho_{\text{fi}}$ & $0.87$ & $0.81$ & $0.96$ & $0.89$ & $0.61$ & $0.50$ & $1.0$ & $0.95$ & $0.46$ & $0.58$ \\ \bottomrule
\end{tabular}
\end{table}

\subsection{Further Details on Diversity Analysis}
\label{subsec:div_appendix}
\subsubsection{Diversity vs Frequency for DataComp1B}
We explain the Vendi-Score~\cite{rezaei2025vendi} metric we use for measure diversity in subsection~\ref{subsec: diversity} in the main paper, and share the Spearman's rank correlation coefficient scores we obtain between the country-wise diversity scores of the entity-images and frequency in case of Re-LAION2B-en (average $\rho_{\text{fi}}=0.54$). A similar analysis for DataComp1B is presented in Table~\ref{tab:div_corr_datacomp}, where we find frequency and diversity to be strongly correlated (average $\rho_{\text{fi}}=0.67$). 

\begin{table*}[h!]
\centering
\small
\caption{\textbf{Frequency and diversity rank correlation coefficient scores for captions in Re-LAION2B-en.} For each entity, we compute the correlation between the frequency of countries and diversity scores of the captions ($\rho_{fc}$). While house, office and jewelry show $\rho_{fc} \leq 0.40$, overall a moderate correlation is observed (average $\rho_{fc} = 0.49$).}
\label{tab:div_corr_text_appendix_laion}
\begin{tabular}{c|cccccccccc}
\toprule
 & House & Flag & Car & Kitchen & Beach & Road & Hotel & \begin{tabular}[c]{@{}c@{}}Bed-\\ room\end{tabular} & Toilet & \begin{tabular}[c]{@{}c@{}}Apart-\\ ment\end{tabular} \\ \midrule
$\rho_{\text{fi}}$ & $0.08$ & $0.83$ & $0.69$ & $0.40$ & $0.71$ & $0.61$ & $0.67$ & $0.45$ & $0.47$ & $0.14$ \\ \midrule
 & \begin{tabular}[c]{@{}c@{}}Moun-\\ tain\end{tabular} & Island & Shop & Valley & \begin{tabular}[c]{@{}c@{}}Living\\ Room\end{tabular} & \begin{tabular}[c]{@{}c@{}}Wedding\\ Dress\end{tabular} & Hill & Lake & Office & Jewelry \\ \hline
$\rho_{\text{fi}}$ & $0.80$ & $0.59$ & $0.58$ & $0.57$ & $0.31$ & $0.44$ & $0.58$ & $0.61$ & $0.00$ & $0.31$ \\ \bottomrule
\end{tabular}
\end{table*}

\begin{table*}[h!]
\centering
\small
\caption{\textbf{Frequency and diversity correlations for captions in DataComp1B.} For each entity, we compute the correlation between the frequency of countries and diversity scores of the captions ($\rho_{fc}$). While jewelry does not exhibit any correlation and bedroom is found to have a weak correlation, overall a moderate correlation is observed for all other entities (average $\rho_{fc} = 0.51$).}
\label{tab:div_corr_text_appendix_datacomp}
\begin{tabular}{c|cccccccccc}
\toprule
 & House & Flag & Car & Kitchen & Beach & Road & Hotel & \begin{tabular}[c]{@{}c@{}}Bed-\\ room\end{tabular} & Toilet & \begin{tabular}[c]{@{}c@{}}Apart-\\ ment\end{tabular} \\ \midrule
$\rho_{\text{fi}}$ & $0.20$ & $0.83$ & $0.67$ & $0.43$ & $0.71$ & $0.55$ & $0.63$ & $0.39$ & $0.24$ & $0.30$ \\ \midrule
 & \begin{tabular}[c]{@{}c@{}}Moun-\\ tain\end{tabular} & Island & Shop & Valley & \begin{tabular}[c]{@{}c@{}}Living\\ Room\end{tabular} & \begin{tabular}[c]{@{}c@{}}Wedding\\ Dress\end{tabular} & Hill & Lake & Office & Jewelry \\ \hline
$\rho_{\text{fi}}$ & $0.52$ & $0.70$ & $0.32$ & $0.14$ & $0.31$ & $1.0$ & $0.80$ & $0.86$ & $0.38$ & $0.28$ \\ \bottomrule
\end{tabular}
\end{table*}

\subsubsection{Captions Diversity}
\label{appn:subsubsec:captions}
To compute the score for the captions, we first encode them using the same \texttt{CLIP ViTB/32} text model. Similar to the images, we compute diversity scores for captions belonging to a specific country and entity, but only consider those that have more than $100$ image-caption pairs for a given
entity. Overall, we find that car, road, valley, hill, lake, office and shop have the most variations in their captions in case of Re-LAION2B-en, while the captions mentioning mountain, island and jewelry additionally exhibit high diversity in case of DataComp1B. The captions of bedroom, office, and apartment are the least diverse for both datasets, since they are mostly
descriptions of the entity, whereas captions for other entities often describe surroundings and related
context. The detailed scores for each entity are shown in Table~\ref{tab:div_entity_merged} for both Re-LAION2B-en and DataComp1B.
We further define $\rho_{fc}$ as the Spearman’s
rank correlation coefficient of country-wise frequency values with the diversity scores of these captions, and find that most entities exhibit moderate positive correlation with frequency with $\rho = 0.49$ and $0.51$ respectively (see Tables~\ref{tab:div_corr_text_appendix_laion} and ~\ref{tab:div_corr_text_appendix_datacomp}) for Re-LAION2B-en and DataComp1B.

\begin{table*}[]
\centering
\small
\caption{\textbf{Entity-wise diversity scores} for the training captions and images of Re-LAION2B-en and Datacomp1B.}
\label{tab:div_entity_merged}
\begin{tabular}{@{}l|ll|ll@{}}
\toprule
\multirow{2}{*}{Entities} & \multicolumn{2}{c|}{Re-LAION2B-en} & \multicolumn{2}{c}{Datacomp1B} \\ 
\cmidrule(lr){2-3} \cmidrule(lr){4-5} 
 & Text & Image & Text & Image \\ \midrule
House & $7.5^{\pm 3.8}$ & $8.3^{\pm 1.8}$ & $6.8^{\pm 2.3}$ & $8.5^{\pm 1.2}$ \\
Flag & $11.0^{\pm 3.8}$ & $9.1^{\pm 2.1}$ & $10.3^{\pm 3.9}$ & $8.3^{\pm 2.3}$ \\
Car & $19.1^{\pm 9.3}$ & $15.8^{\pm 5.4}$ & $15.4^{\pm 7.0}$ & $11.6^{\pm 3.2}$ \\
Kitchen & $5.9^{\pm 2.0}$ & $4.6^{\pm 0.7}$ & $5.5^{\pm 1.6}$ & $4.4^{\pm 0.7}$ \\
Beach & $12.1^{\pm 4.8}$ & $7.9^{\pm 2.2}$ & $9.5^{\pm 3.8}$ & $7.3^{\pm 1.9}$ \\
Road & $15.0^{\pm 5.2}$ & $11.1^{\pm 2.7}$ & $13.6^{\pm 3.9}$ & $11.6^{\pm 2.6}$ \\
Hotel & $6.5^{\pm 1.6}$ & $8.1^{\pm 1.5}$ & $5.5^{\pm 1.1}$ & $8.0^{\pm 1.2}$ \\
Bedroom & $6.7^{\pm 2.3}$ & $4.6^{\pm 0.7}$ & $5.7^{\pm 2.4}$ & $5.6^{\pm 0.3}$ \\
Toilet & $7.8^{\pm 2.3}$ & $5.2^{\pm 1.2}$ & $5.9^{\pm 1.5}$ & $3.9^{\pm 0.5}$ \\
Apartment & $4.0^{\pm 1.9}$ & $7.0^{\pm 1.3}$ & $4.4^{\pm 1.2}$ & $9.9^{\pm 1.0}$ \\ 
Hill & $14.8^{\pm6.2}$ & $9.9^{\pm 2.5}$ & $17.4^{\pm6.6}$ & $15.9^{\pm 1.6}$ \\
Mountain & $13.8^{\pm 4.5}$ & $9.8^{\pm 2.3}$ & $16.4^{\pm 5.4}$ & $12.8^{\pm 1.7}$ \\
Island & $10.0^{\pm 4.4}$ & $7.2^{\pm 2.3}$ & $16.7^{\pm 6.3}$ & $10.5^{\pm 1.2}$ \\
Valley & $15.6^{\pm 6.9}$ & $13.7^{\pm 1.9}$ & $16.4^{\pm 6.5}$ & $12.3^{\pm 1.8}$ \\
Shop & $15.1^{\pm 5.8}$ & $16.3^{\pm 4.2}$ & $16.6^{\pm 8.3}$ & $18.3^{\pm 3.1}$ \\
Wedding Dress & $7.5^{\pm 2.8}$ & $7.3^{\pm 1.8}$ & $10.8^{\pm 3.9}$ & $9.4^{\pm 2.1}$ \\
Lake & $14.0^{\pm 5.8}$ & $10.1^{\pm 2.3}$ & $14.8^{\pm 4.7}$ & $14.0^{\pm 1.7}$ \\
Living Room & $9.6^{\pm 3.3}$ & $6.5^{\pm 1.2}$ & $7.7^{\pm 2.9}$ & $6.1^{\pm 1.2}$ \\
Jewelry & $12.2^{\pm 4.2}$ & $10.9^{\pm 1.6}$ & $15.6^{\pm 5.3}$ & $9.9^{\pm 2.2}$ \\
Office & $8.8^{\pm 3.8}$ & $10.0^{\pm 2.2}$ & $7.6^{\pm 3.1}$ & $9.3^{\pm 2.1}$ \\
\bottomrule
\end{tabular}
\end{table*}

\begin{figure*}[t] %
    \centering
    \includegraphics[trim=6cm 0cm 6cm 0cm, clip, width=\linewidth]{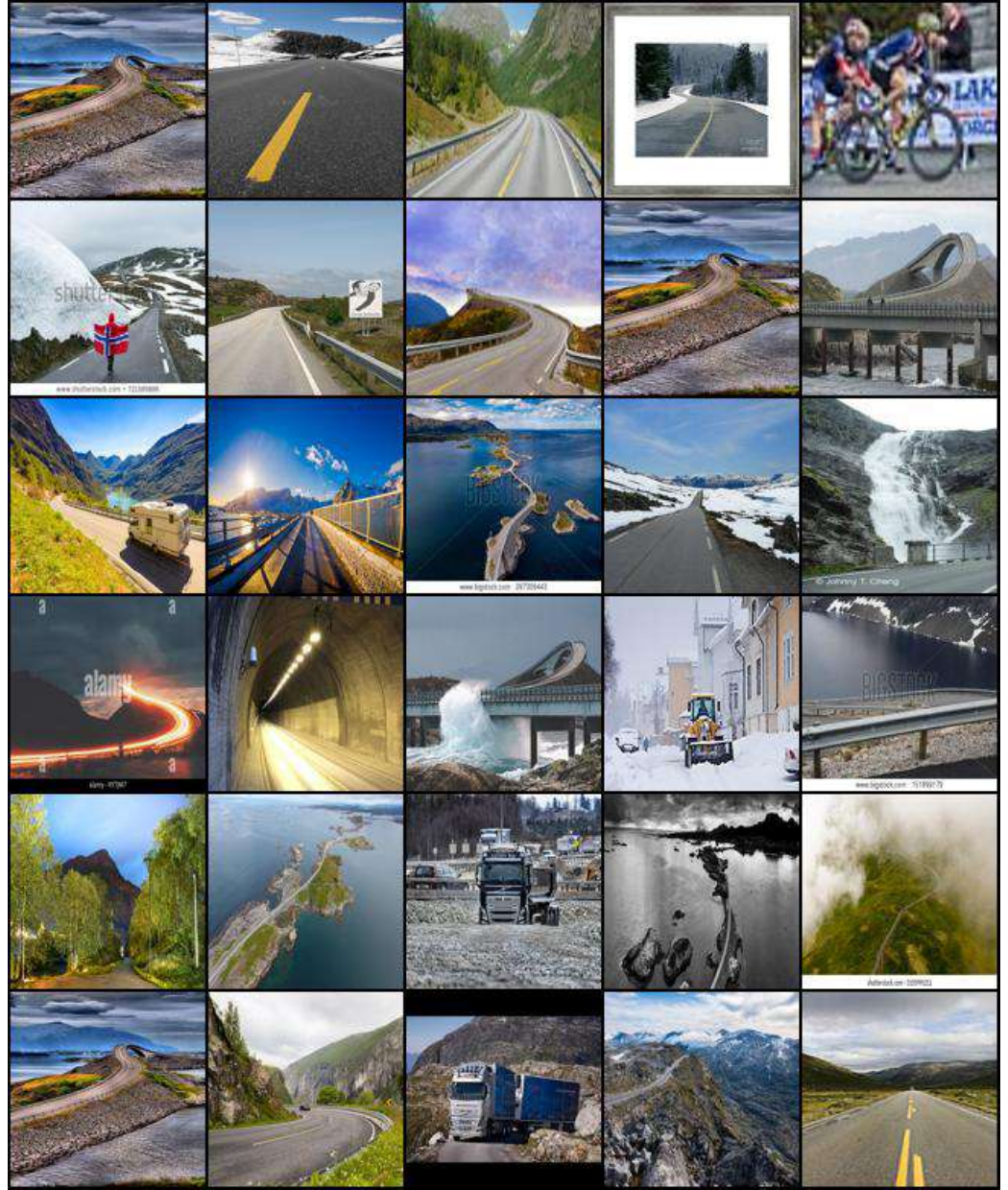} %
    \caption{\textbf{Training images of Road for Norway.} There are $715$ images of roads from Norway. The diversity score for the images is $8.82$.}
    \label{fig:norway}
\end{figure*}

\begin{figure*}[t] %
    \centering
    \includegraphics[trim=6cm 0cm 6cm 0cm, clip, width=\linewidth]{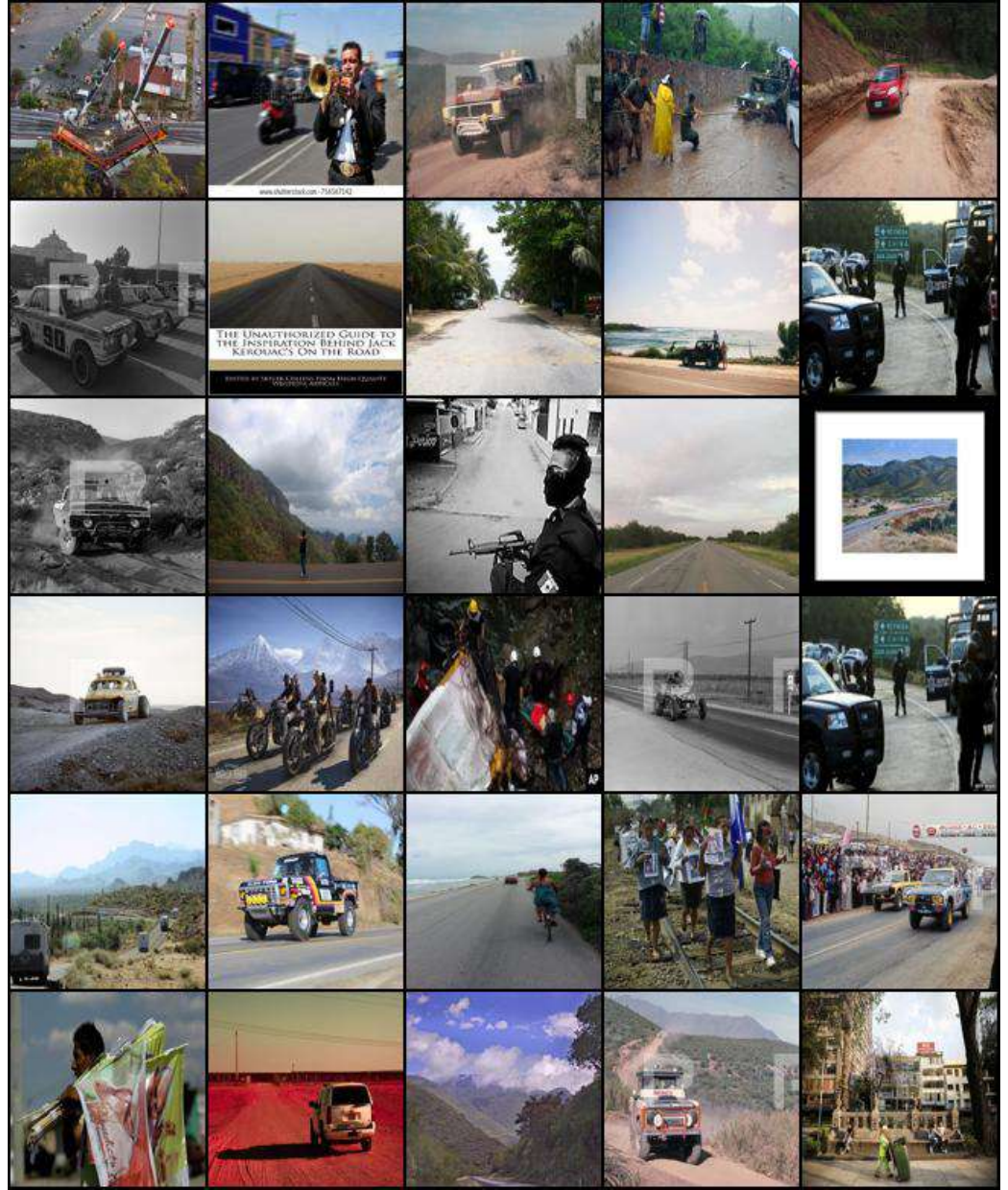} %
    \caption{\textbf{Training images of Road for Mexico.} There are $274$ images of roads from Mexico. The diversity score for the images is $13.98$.}
    \label{fig:mexico}
\end{figure*}

\begin{figure*}[t] %
    \centering
    \includegraphics[trim=0cm 6cm 0cm 6cm, clip, width=\linewidth]{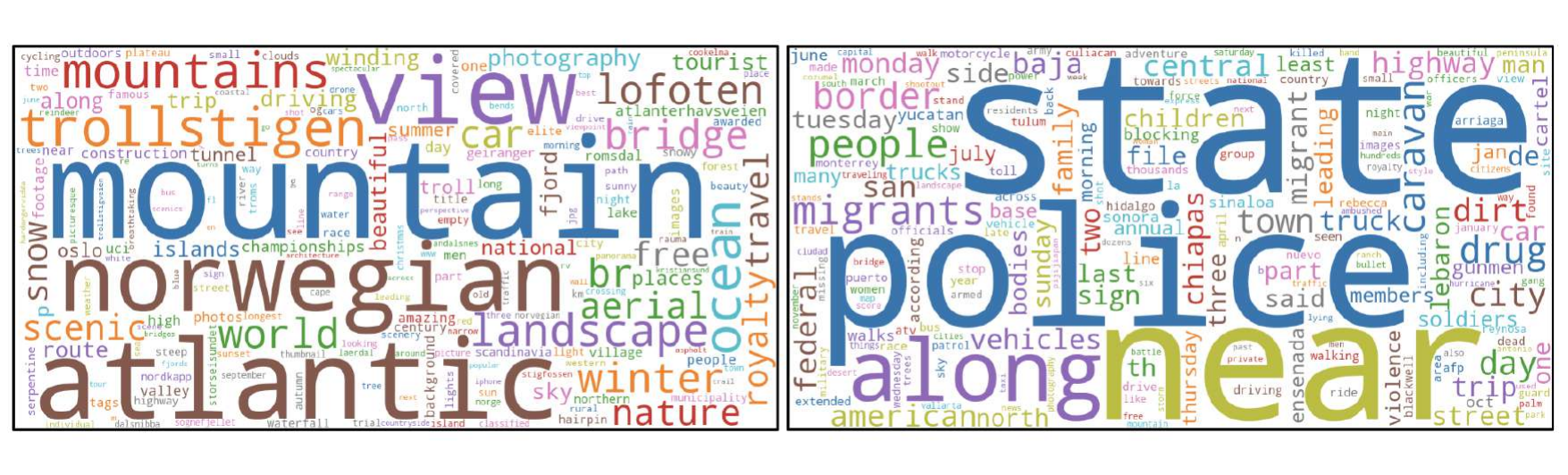} %
    \caption{\textbf{Caption word clouds for road images} from Norway (left) and Mexico (right) in the Re-LAION2B-en dataset.}
    \label{fig:word_cloud}
\end{figure*}

\textbf{Qualitative Visualizations}. Further, we show the images of roads from Norway ($div(\text{Norway})=8.82$) and Mexico ($div(\text{Mexico})=13.98$) as present in the Re-LAION2B-en dataset in Figures~\ref{fig:norway} and \ref{fig:mexico}. We notice that while the images from Norway mostly show images of roads in various landscapes, the images from Mexico depict other noises like humans and cars, leading to Mexican roads having higher diversity scores than those of Norway. 
To support this observation, we additionally present word cloud visualizations constructed from the captions associated with these images for Norway and Mexico (Figure~\ref{fig:word_cloud}). They provide an intuitive summary of the dominant semantic concepts present in the images.

\begin{figure*}[t] %
    \centering
    \includegraphics[width=\linewidth]{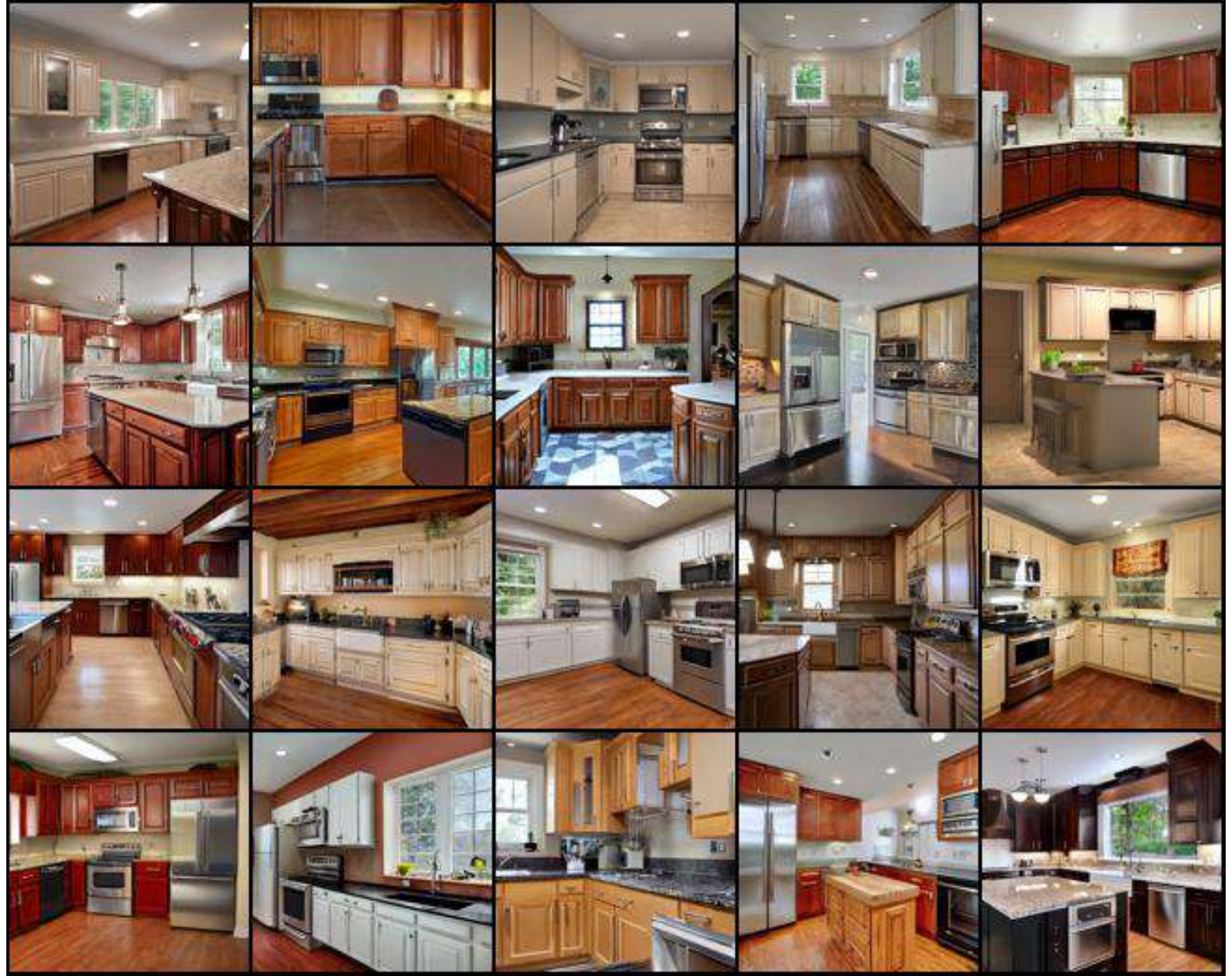} %
    \caption{\textbf{Kitchen images generated for United States.} While we found $30818$ American kitchens in the corresponding subset of Re-LAION2B-en, the diversity score for generated images from the same country is $1.67$.}
    \label{fig:us}
\end{figure*}

\begin{figure*}[t] %
    \centering
    \includegraphics[width=\linewidth]{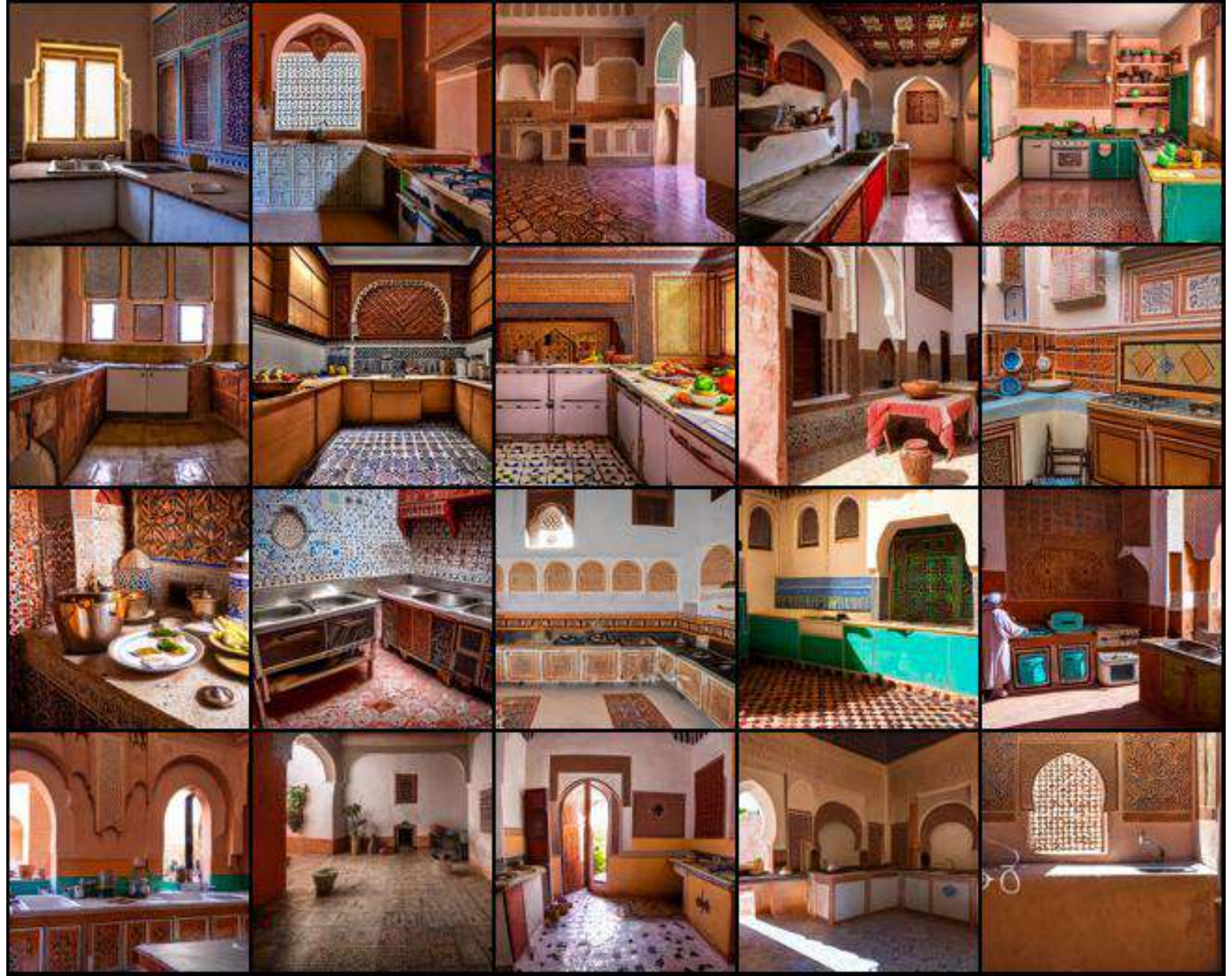} %
    \caption{\textbf{Kitchen images generated for Morocco.} While we found $144$ Moroccan kitchens in the corresponding subset of Re-LAION2B-en, the diversity score for generated images from the same country is $2.52$.}
    \label{fig:morocco}
\end{figure*}

\subsection{Examining Country-wise Model Generations - Extended Analysis}
\label{appn:subsec:generated}

In case of generated images, we see that kitchens of United States, having frequency of $30818$ in the examined subset of Re-LAION2B-en, have a diversity score of $1.67$, whereas Morocco, with a frequency of $144$ has a diversity score of $2.52$. Example images can be seen in Figures~\ref{fig:us} and \ref{fig:morocco} for United States and Morocco respectively. While the images of United States seem visually alike, those of Morocco show more variations. For example, some images are inside kitchens, some are outside. 
In fact, some images of Morocco do not have kitchens at all, thereby contributing towards a notion of increased diversity at the cost of inaccurate depictions of the entity. Note that the prompt we use for generating the images for any entity $e$ and country $c$ is: ``\texttt{High definition image of a \{e\} in \{c\}}'' . Such a simplistic instruction prompts the model to generate the exact entity in question, based on its understanding of the same. Hence, we notice much lesser variations in the generated images than those of the real images.

\end{document}